%% file: ms.tex
\pdfoutput=1
\documentclass{article}

\usepackage{microtype}
\usepackage{graphicx}
\usepackage{subfigure}
\usepackage{booktabs} 

\usepackage{hyperref}

\usepackage[accepted]{icml2021}

\usepackage{amssymb}
\usepackage{dsfont}
\usepackage{amsmath}
\DeclareMathOperator*{\argmax}{arg\,max}
\newcommand{\quan}{\textcolor{red}{QN: }}
\newcommand{\acro}[1]{\textsc{\MakeLowercase{#1}}}
\usepackage{multirow}
\usepackage{xcolor}
\usepackage{pgfplots}
\pgfplotsset{compat=1.17, every non boxed x axis/.append style={x axis line style=-},
     every non boxed y axis/.append style={y axis line style=-}}
\usepackage{setspace}
\usepackage{breakurl}

\usepackage{algorithm}
\usepackage{algpseudocode}
\algrenewcommand\algorithmicrequire{\textbf{inputs}}
\algrenewcommand\algorithmicensure{\textbf{returns}}

\makeatletter
\newcommand{\pushright}[1]{\ifmeasuring@#1\else\omit\hfill$\displaystyle#1$\fi\ignorespaces}
\newcommand{\pushleft}[1]{\ifmeasuring@#1\else\omit$\displaystyle#1$\hfill\fi\ignorespaces}
\makeatother

\icmltitlerunning{Nonmyopic Multifidelity Active Search}

\begin{document}

\twocolumn[
\icmltitle{Nonmyopic Multifidelity Active Search}




\begin{icmlauthorlist}
\icmlauthor{Quan Nguyen}{wustl}
\icmlauthor{Arghavan Modiri}{uot}
\icmlauthor{Roman Garnett}{wustl}
\end{icmlauthorlist}

\icmlaffiliation{wustl}{Washington University in St. Louis, MO, USA}
\icmlaffiliation{uot}{University of Toronto, Toronto, Canada}

\icmlcorrespondingauthor{Quan Nguyen}{quan@wustl.edu}

\icmlkeywords{Multifidelity, Active Search, Nonmyopic, ICML}

\vskip 0.3in
]



\printAffiliationsAndNotice{}  

\begin{abstract}
Active search is a learning paradigm where we seek to identify as many members of a rare, valuable class as possible given a labeling budget.
Previous work on active search has assumed access to a faithful (and expensive) oracle reporting experimental results.
However, some settings offer access to cheaper surrogates such as computational simulation that may aid in the search.
We propose a model of \emph{multifidelity} active search, as well as a novel, computationally efficient policy for this setting that is motivated by state-of-the-art classical policies.
Our policy is nonmyopic and budget aware, allowing for a dynamic tradeoff between exploration and exploitation.
We evaluate the performance of our solution on real-world datasets and demonstrate significantly better performance than natural benchmarks.
\end{abstract}

\input{main_01_intro}
\input{main_02_def}
\input{main_03_approx}
\input{main_04_related}
\input{main_05_exp}

\input{main_06_conclusion}

\section*{Acknowledgements}

We would like to thank the anonymous reviewers for their feedback.
\acro{QN} and \acro{RG} were supported by the National Science Foundation (\acro{NSF}) under award numbers \acro{OAC}–1940224 and \acro{IIS}–1845434.


\bibliography{ref}
\bibliographystyle{icml2021}

\newpage
\appendix

\section{Pseudocode}

We give the pseudocode for \acro{MF--ENS} in Algorithm \ref{alg:1} (for $H$ queries) and Algorithm \ref{alg:2} (for $L$ queries).
The two versions are nearly identical; however, in Algorithm \ref{alg:2}, we marginalize both the putative label $y_L$ and the pending label $y_H$ for $L$ queries, and the $\Pr(y_H = 1 \mid x, \mathcal{D})$ term is dropped from the final line according to (1) in the main paper.

\section{The One-Stage Approximation Policy}

We name the one-stage approximation policy described in \S 3.1 of the main paper $H$--\acro{ENS}.
Recall $H$--\acro{ENS} applies the \acro{ENS} heuristic directly by assuming in its lookahead that immediately after the putative query, all budget on $H$ is spent simultaneously on a batch of queries, disregarding any $L$ queries that may be made before that batch.

The time complexity of $H$--\acro{ENS} is the same as that of \acro{ENS}, $\mathcal{O} \bigl( n \, ( \log n + m \log m ) \bigr)$, where $n$ is the size of the candidate pool and $m$ is the maximum number of points whose probablities are affected by a newly revealed label \cite{jiang2017efficient}.
To avoid having to consider a large number of unpruned candidates on a single iteration, we set $u = s = 1000$, where $u$ is the upper limit of the number of candidates are to be considered before a random subset of size $s$ of the remaining candidates is selected.
This strategy is described in detail in \S 3.3 of the main paper.

\begin{figure*}
\begin{minipage}{1\textwidth}
\begin{algorithm}[H]
\caption{\acro{MF--ENS} for $H$ queries} \label{alg:1}
\begin{algorithmic}[1]
\Require $x$, $\mathcal{D}$
\Ensure approximate expected utility for querying $x$ on $H$, $f(x)$ \Comment{design query by maximizing}
\vspace*{0.1ex}
\For{$y_H \in \{ -, + \}$} \Comment{probabilities: $\Pr(y_H \mid x, \mathcal{D})$}
  \State determine $X_L$ by finding top-$k$ $v$ scores among unlabeled $L$ points\strut
  \For{$Y_L \in \{ -, + \}^k$} \Comment{probabilities: $\Pr(Y_L \mid x, y_H, X_L, \mathcal{D})$}
    \State $s \left( x, y_H, X_L, Y_L \right) \leftarrow \sum\nolimits_{\ell_H}' \Pr \left( y'_H = 1 \mid x', x, y_H, X_L, Y_L \right)$ \Comment{(4)}
  \EndFor
\State $f \left( x \mid y_H \right) \leftarrow \mathbb{E}_{Y_L} \big[ s \left( x, y_H, X_L, Y_L \right) \mid x, y_H, X_L \big]$ \Comment{(4)}
\EndFor{}\\
\Return $f(x) \leftarrow \Pr \left( y_H = 1 \mid x, \mathcal{D} \right) + \mathbb{E}_{y_H} \big[ f \left( x \mid y_H \right) \mid x, \mathcal{D} \big]$ \Comment{(2)}
\end{algorithmic}
\end{algorithm}

\begin{algorithm}[H]
\caption{\acro{MF--ENS} for $L$ queries} \label{alg:2}
\begin{algorithmic}[1]
\Require $x$, $\mathcal{D}$
\Ensure approximate expected utility for querying $x$ on $L$, $f(x)$ \Comment{design query by maximizing}
\vspace*{0.1ex}
\For{$y_L, y_H \in \{ -, + \}^2$} \Comment{probabilities: $\Pr(y \mid x, \mathcal{D})$}
  \State determine $X_L$ by finding top-$\overline{k}$ $v$ scores among unlabeled $L$ points\strut
  \For{$Y_L \in \{ -, + \}^{\overline{k}}$} \Comment{probabilities: $\Pr(Y_L \mid x, y_L, X_L, \mathcal{D})$}
    \State $s \left( x, y_L, X_L, Y_L \right) \leftarrow \sum\nolimits_{\ell_H}' \Pr \left( y'_H = 1 \mid x', x, y_L, X_L, Y_L \right)$ \Comment{(4)}
  \EndFor
\State $f \left( x \mid y_L \right) \leftarrow \mathbb{E}_{Y_L} \big[ s \left( x, y_L, X_L, Y_L \right) \mid x, y_L, X_L \big]$ \Comment{(4)}
\EndFor{}\\
\Return $f(x) \leftarrow \mathbb{E}_{y_L} \big[ f \left( x \mid y_L \right) \mid x, \mathcal{D} \big]$ \Comment{(1)}
\end{algorithmic}
\end{algorithm}
\end{minipage}
\end{figure*}

\section{Further Experiment Results}

\begin{table*}
\caption{
Experiment results with an $H$ budget of $t = 300$, averaged across all repeated experiments for each setting.
$\theta$ is the simulated false positive rate of the low-fidelity oracle;
$k$ is the number of $L$ queries made between two $H$ queries (i.e., the speed ratio between the two fidelities).
Each entry denotes the average number of targets found across the repeated experiments and the corresponding standard error in parentheses.
The best performance in each column is highlighted bold;
those that are not significantly worse than the best (using a one-sided paired $t$-test with significance level $\alpha = 0.05$) are in blue italic.
}
\vskip 0.15in
\centering
\resizebox{\textwidth}{!}{%
\begin{tabular}{lrrrrrrrrrrrr}
\toprule
\multicolumn{1}{c}{} & \multicolumn{4}{c}{\acro{ECFP}4} & \multicolumn{4}{c}{\acro{G}pi\acro{DAPH}3} & \multicolumn{4}{c}{\acro{BMG}} \\
\cmidrule(lr){2-5} \cmidrule(lr){6-9} \cmidrule(lr){10-13}
\multicolumn{1}{c}{}
& \multicolumn{2}{c}{$\theta = 0.1$} & \multicolumn{2}{c}{$\theta = 0.3$}
& \multicolumn{2}{c}{$\theta = 0.1$} & \multicolumn{2}{c}{$\theta = 0.3$}
& \multicolumn{2}{c}{$\theta = 0.1$} & \multicolumn{2}{c}{$\theta = 0.3$} \\
\cmidrule(lr){2-3} \cmidrule(lr){4-5} \cmidrule(lr){6-7} \cmidrule(lr){8-9} \cmidrule(lr){10-11} \cmidrule(lr){12-13}
\multicolumn{1}{c}{} & \multicolumn{1}{c}{$k = 2$} & \multicolumn{1}{c}{$k = 5$} & \multicolumn{1}{c}{$k = 2$} & \multicolumn{1}{c}{$k = 5$}
& \multicolumn{1}{c}{$k = 2$} & \multicolumn{1}{c}{$k = 5$} & \multicolumn{1}{c}{$k = 2$} & \multicolumn{1}{c}{$k = 5$}
& \multicolumn{1}{c}{$k = 2$} & \multicolumn{1}{c}{$k = 5$} & \multicolumn{1}{c}{$k = 2$} & \multicolumn{1}{c}{$k = 5$} \\
\midrule
\acro{ENS}
& 218 (4.5) & 218 (4.5)
& 213 (4.2) & 213 (4.2)
& 197 (4.7) & 197 (4.7)
& 186 (5.3) & 186 (5.3)
& 279 (1.6) & 279 (1.6)
& 278 (1.7) & 278 (1.7) \\
\acro{MF--UCB}
& 227 (4.8) & 238 (4.4)
& 200 (5.7) & 206 (5.8)
& 200 (5.7) & 212 (5.7)
& 187 (5.5) & 200 (5.3)
& 284 (1.0) & 289 (1.0)
& 274 (2.0) & 277 (1.7) \\
\acro{UG}
& 228 (4.9) & 237 (4.4)
& 207 (5.8) & 209 (5.8)
& 204 (5.6) & 211 (5.7)
& 191 (5.6) & 202 (5.6)
& 283 (1.2) & 287 (1.1)
& 278 (2.0) & 280 (1.6) \\
$H$--\acro{ENS}
& \emph{\color{blue} 241 (3.6)} & \emph{\color{blue} 249 (3.5)}
& \emph{\color{blue} 223 (4.3)} & \textbf{231 (4.1)}
& 221 (4.5) & \textbf{245 (3.0)}
& 202 (4.8) & \textbf{222 (3.8)}
& \textbf{290 (0.7)} & \textbf{294 (0.5)}
& \emph{\color{blue} 284 (1.2)} & \emph{\color{blue} 285 (1.2)} \\
\acro{MF--ENS}
& \textbf{244 (3.5)} &  \textbf{250 (3.2)}
& \textbf{226 (4.5)} & \emph{\color{blue} 230 (4.4)}
& \textbf{230 (3.8)} & \emph{\color{blue} 243 (3.4)}
& \textbf{208 (5.0)} & \emph{\color{blue} 221 (4.3)}
& \emph{\color{blue} 290 (0.7)} & \emph{\color{blue} 294 (0.6)}
& \textbf{286 (1.1)} & \textbf{286 (1.1)} \\
\bottomrule
\end{tabular}}
\label{tab:sim-results}
\vskip -0.1in
\end{table*}

Table \ref{tab:sim-results} shows the average number of targets found by all considered policies including $H$--\acro{ENS}.
$H$--\acro{ENS} performs well in comparison with myopic baselines and the single-fidelity \acro{ENS} policy.
Against $H$--\acro{ENS}, our main policy \acro{MF--ENS} offers a slight improvement.
Across all experiments, \acro{MF--ENS} finds an average of 237 targets, whereas $H$--\acro{ENS} finds 235;
a two-sided paired $t$--test rejects the hypothesis that the average difference in the number of targets found between the two policies is zero with a $p$--value of $p = 0.044$.

\begin{table*}
\caption{
$p$--values from two-paired $t$--tests, each testing for the hypothesis that the average number of targets found between \acro{MF--ENS} and another policy is zero.
}
\vskip 0.15in
\centering
\resizebox{\textwidth}{!}{%
\begin{tabular}{lllllllllllll}
\toprule
\multicolumn{1}{c}{} & \multicolumn{4}{c}{\acro{ECFP}4} & \multicolumn{4}{c}{\acro{G}pi\acro{DAPH}3} & \multicolumn{4}{c}{\acro{BMG}} \\
\cmidrule(lr){2-5} \cmidrule(lr){6-9} \cmidrule(lr){10-13}
\multicolumn{1}{c}{}
& \multicolumn{2}{c}{$\theta = 0.1$} & \multicolumn{2}{c}{$\theta = 0.3$}
& \multicolumn{2}{c}{$\theta = 0.1$} & \multicolumn{2}{c}{$\theta = 0.3$}
& \multicolumn{2}{c}{$\theta = 0.1$} & \multicolumn{2}{c}{$\theta = 0.3$} \\
\cmidrule(lr){2-3} \cmidrule(lr){4-5} \cmidrule(lr){6-7} \cmidrule(lr){8-9} \cmidrule(lr){10-11} \cmidrule(lr){12-13}
\multicolumn{1}{c}{} & \multicolumn{1}{c}{$k = 2$} & \multicolumn{1}{c}{$k = 5$} & \multicolumn{1}{c}{$k = 2$} & \multicolumn{1}{c}{$k = 5$}
& \multicolumn{1}{c}{$k = 2$} & \multicolumn{1}{c}{$k = 5$} & \multicolumn{1}{c}{$k = 2$} & \multicolumn{1}{c}{$k = 5$}
& \multicolumn{1}{c}{$k = 2$} & \multicolumn{1}{c}{$k = 5$} & \multicolumn{1}{c}{$k = 2$} & \multicolumn{1}{c}{$k = 5$} \\
\midrule
\acro{ENS}
& $4 \times 10^{-27}$ & $3 \times 10^{-30}$
& $7 \times 10^{-12}$ & $2 \times 10^{-18}$
& $1 \times 10^{-26}$ & $1 \times 10^{-40}$
& $3 \times 10^{-14}$ & $3 \times 10^{-22}$
& $4 \times 10^{-10}$ & $5 \times 10^{-14}$
& $7 \times 10^{-7}$ & $6 \times 10^{-5}$ \\
\acro{MF--UCB}
& $1 \times 10^{-10}$ & $1 \times 10^{-6}$
& $7 \times 10^{-18}$ & $6 \times 10^{-14}$
& $1 \times 10^{-18}$ & $6 \times 10^{-18}$
& $1 \times 10^{-13}$ & $7 \times 10^{-18}$
& $5 \times 10^{-8}$ & $8 \times 10^{-6}$
& $1 \times 10^{-8}$ & $8 \times 10^{-8}$ \\
\acro{UG}
& $5 \times 10^{-8}$ & $2 \times 10^{-7}$
& $2 \times 10^{-10}$ & $2 \times 10^{-10}$
& $2 \times 10^{-15}$ & $5 \times 10^{-19}$
& $6 \times 10^{-11}$ & $1 \times 10^{-14}$
& $9 \times 10^{-8}$ & $4 \times 10^{-8}$
& $1 \times 10^{-5}$ & $4 \times 10^{-5}$ \\
\bottomrule
\end{tabular}}
\label{tab:p-values}
\vskip -0.1in
\end{table*}

Table \ref{tab:p-values} shows the $p$--values from two-sided paired $t$--tests, each testing for the hypothesis that the average number of targets found between \acro{MF--ENS} and each of the baselines considered in \S 5 of the main paper is zero.
All hypotheses are rejected with overwhelming confidence.

We also show the difference in the cumulative number of targets found between \acro{MF--ENS} and \acro{UG} in Figure \ref{fig:vs_ug}.
The plot shows the same trend as that for \acro{MF--ENS} and \acro{MF-UCB} in the main paper:
\acro{MF--ENS} underperforms the myopic policy at the beginning of the search during its exploration phase, but quickly recovers and is able to find significantly more targets in the end.

\begin{figure}[H]
\vskip 0.2in
\centering
\input{media/utility_diff_UG}
\caption{The difference in cumulative targets found between \acro{MF--ENS} and \acro{UG}, averaged across all experiments and datasets.}
\label{fig:vs_ug}
\vskip -0.2in
\end{figure}

\begin{table}[H]
\caption{
Average pruning rate and the contribution of each of the two pruning methods.
}
\vskip 0.15in
\centering
\resizebox{0.48\textwidth}{!}{%
\begin{tabular}{lccc}
\toprule
\multirow{2}{*}{policies} & \multirow{2}{*}{full coverage rate} & \multicolumn{2}{c}{given full coverage} \\
\cmidrule(lr){3-4}
&  & total prune \% & \multicolumn{1}{c}{partial prune \%} \\
\midrule
$H$--\acro{ENS}
& \multicolumn{1}{c}{45.2\%}
& \multicolumn{1}{c}{85.2\%}
& 14.0\% \\
\acro{MF--ENS}
& \multicolumn{1}{c}{27.6\%}
& \multicolumn{1}{c}{89.4\%}
& 9.2\% \\
\bottomrule
\end{tabular}}
\label{tab:pruning-results}
\vskip -0.1in
\end{table}

\subsection{Effects of Pruning}

Finally, we examine the effectiveness of our pruning strategies in helping $H$--\acro{ENS} and \acro{MF--ENS} cover the entire search space.
Each of the two policies has an upper limit on how many candidates are to have their scores fully computed.
If this limit is reached, we only consider a random subset of the remaining candidate pool.
We classify each time a policy returns while there are unpruned candidates remaining as an instance of failure to cover the entire space.

First, we compute the fraction of iterations, across all experiments, in which this limit is not exceeded, or in other words, how often each policy can exactly consider all candidates with the help of pruning.
When a point is pruned, it may be before any actual computation (total pruning) if its score upper bound is lower than the current highest score $f^*$, satisfying (5) in the main paper.
Otherwise, it may be pruned during the calculation of its scores (partial pruning) if its partial score, combined with partial upper bounds, is lower than $f^*$.
We keep track of the fraction of points pruned by each of the two methods;
these results are averaged across the iterations in which pruning helps cover the entire candidate pool.
We report these statistics in Table \ref{tab:pruning-results}.

We observe that while we do not cover the entire candidate pool in many iterations, the effect of pruning is dramatic when it is successful.
The existing pruning method (total pruning) helps eliminate most of the candidates, and our extension (partial pruning) raises the combined pruning rate to roughly 99\% on average.
In our experiments, successful pruning could reduce the time a policy takes to produce a decision from under an hour to mere seconds.

\end{document}

%% file: main_01_intro.tex
\section{Introduction}


The goal of \emph{active search} is to identify members of a rare and valuable class among a large pool of unlabeled points.
This is a simple model of many real-world discovery problems, such as drug discovery and fraud detection.
Active search proceeds by successively querying an oracle that returns a binary label indicating whether or not a chosen data point exhibits the desired properties.
In many applications, this oracle is expensive, limiting the number of queries that could be made.
The challenge is to design a policy to sequentially query the oracle in order to discover as many targets as possible, subject to a given labeling budget.

Active search has been extensively studied under various settings
\cite{garnett2012bayesian,jiang2017efficient,jiang2018efficient,jiang2019cost}.
Notably, \citet{jiang2017efficient} proved a hardness of approximation result, showing that no polynomial-time policy can approximate the performance of the Bayesian optimal policy within any constant factor. Thus active search is surprisingly hard. However, the authors also proposed an efficient approximation to the optimal policy that delivers impressive empirical performance.
The key feature of their policy is its ability to account for the remaining budget and dynamically trade off exploration and exploitation.

Most previous work on active search has assumed access to only a single expensive oracle providing labels.
In practice, however, we may have several methods of probing the search space, including cheap, low-fidelity surrogates.
For example, a computational simulation may serve as a noisy approximation to an experiment done in a laboratory, and we may reasonably seek to use a cheaper computational search to help design the expensive experiments.
This motivates the problem of \emph{multifidelity active search,} where oracles of different fidelities may be simultaneously accessed to accelerate the search process.
The central question in this task is how to effectively leverage these oracles in order to maximize the rate of discovery.

We present a model for multifidelity active search and study the problem under the framework of Bayesian decision theory.
We propose a novel policy for this setting inspired by the state-of-the-art single-fidelity policy mentioned above.
The result is an efficient approximation to the optimal multifidelity policy that is specifically tailored to take advantage of low-fidelity oracles.
We also create aggressive branch-and-bound pruning strategies to increase the efficiency of our proposed algorithm, enabling scaling to large datasets.
In a series of experiments, we investigate the performance of our policy on several real-world datasets for scientific discovery.
Our solution outperforms various baselines from the literature by a large margin.

%% file: main_02_def.tex
\section{Problem Definition}


We first introduce the general active search model and notations.
Suppose we are given a finite set of points $\mathcal{X} \triangleq \{ x_i \}$, which includes a rare, valuable subset $\mathcal{R} \subset \mathcal{X}$.
The members of $\mathcal{R}$, which we call \emph{targets} or \emph{positives}, are not known \emph{a priori}, but whether a given point $x \in \mathcal{X}$ is a target can be determined by making a query to an oracle that returns the binary label $y \triangleq \mathds{1} \{ x \in \mathcal{R} \}$.
We assume that querying the oracle is expensive and that we only have a limited budget of $t$ queries to do so. 
We denote a given dataset of queried points and their corresponding labels as $\mathcal{D} = \{ (x_i, y_i) \}$.
At times, we will use $\mathcal{D}_i$ to denote the dataset collected after $i$ queries.

\subsection{Multifidelity Active Search}

\begin{figure}
\vskip 0.2in
\begin{center}
\resizebox{0.48\textwidth}{!}{%
\input{media/diagram}
}
\caption{
An illustration of our multifidelity active search model.
Each short vertical line indicates when a query finishes and the next query is made.
The numbers indicate the order in which queries are made across the two oracles.
The low-fidelity oracle is $k = 2$ times faster than the high-fidelity oracle.
The budget on $H$ is $t = 10$;
in total, $T = t + kt - k = 28$ queries are made.
}
\label{fig:diagram}
\end{center}
\vskip -0.2in
\end{figure}

In addition to the expensive oracle that returns the exact label of a query, we have access to other cheaper but more noisy oracles that are low-fidelity approximations to the exact oracle.
We limit our setting to two levels of fidelity---one exact oracle and one noisy oracle, denoted as $H$ and $L$, respectively---but note that our proposed algorithm can be extended to more than two fidelities with minor modifications, as we will discuss later.

For a given point $x \in \mathcal{X}$, we denote its exact label on $H$ as $y_H$ and its noisy label on $L$ as $y_L$.
Under this setting, a dataset $\mathcal{D}$ can be partitioned into $\mathcal{D}_L$, the observations made on $L$, and $\mathcal{D}_H$, those that are made on $H$.
Recall that our objective is to query as many targets as possible;
as such, to express our preference over different datasets, we use the natural utility function
\[
u \left( \mathcal{D} = \mathcal{D}_L \cup \mathcal{D}_H \right) \triangleq\! \sum_{y_i \in \mathcal{D}_H} \!y_i,
\]
the number of targets discovered on the $H$ fidelity.

We assume that queries to different oracles are run \emph{in parallel}, but a high-fidelity query takes longer to complete than a low-fidelity one.
In particular, we will assume that each query on $H$ is $k$ times slower than a query on $L$.\footnote{This assumption is for simplicity; \emph{asynchronous} queries could be addressed with a slight modification of our proposed algorithm.}
That is, each time we make a query on $H$, we may make $k$ sequential queries on $L$ while waiting for the result.
The process will then repeat until the budget is depleted.
This model aims to emulate real-life scientific discovery procedures, the main motivation for active search, where multiple fidelities (e.g., computational and experimental campaigns) are often run in parallel but vary in response time.

If we are given a budget of $t$ queries on $H$, we can make $k t - k$ queries on $L$;
in total, $T = t + k t - k$ queries are made.
(Although a total of $kt$ queries can be made on $L$, after the final query on $H$ is made at iteration $T$, further queries on $L$ do not affect the final utility. We thus terminate after $T$ queries.)
This query schedule illustrated in Figure \ref{fig:diagram} for $t = 10$ and $k = 2$.

\subsection{The Bayesian Optimal Policy}

We now derive the optimal (expected-case) policy using Bayesian decision theory.
First, we assume access to a probabilistic classification model that computes the posterior probability that a point $x \in \mathcal{X}$ is a target given a dataset $\mathcal{D}$, $\Pr \left( y_H = 1 \mid x, \mathcal{D} \right)$, as well as the posterior probability that the same $x$ is a positive on $L$, $\Pr \left( y_L = 1 \mid x, \mathcal{D} \right)$.

Suppose we are currently at iteration $i + 1 \leq T$, having observed the dataset $\mathcal{D}_i$, and now need to make the next query, requesting the label of an unlabeled point $x_{i + 1}$.
The Bayesian optimal policy selects the point that maximizes the expected utility of the terminal dataset $\mathcal{D}_T$, assuming future queries will too be made optimally:
\[
x_{i + 1}^* = \argmax_{x_{i + 1} \in \mathcal{X} \setminus \mathcal{D}_i} \mathbb{E} \bigl[ u \left( \mathcal{D}_T \setminus \mathcal{D}_i \right) \mid x_{i + 1}, \mathcal{D}_i \bigr].
\]
If the query is on $H$, the expectation is taken over the posterior distribution of the label of the putative query $x_{i + 1}$;
if the query is on $L$, it is over the \emph{joint} distribution of the $L$ label of $x_{i + 1}$ and the label of the pending $H$ query (recall that we sequentially query $k$ points on $L$ while waiting for an $H$ query that runs in parallel to finish).

To compute this expected utility, we follow the backward induction procedure described in 
\citet{bellman1957dynamic}.
In the base case, we are at iteration $T$ and need to make the last $H$ query and
\begin{align*}
& \mathbb{E} \bigl[ u \left( \mathcal{D}_T \setminus \mathcal{D}_{T - 1} \mid x_T, \mathcal{D}_{T - 1} \right) \bigr] \hfill \\
& \qquad = \sum_{y_H} u \left( \mathcal{D}_T \setminus \mathcal{D}_{T - 1} \right) \Pr \left( y_H \mid x_T, \mathcal{D}_{T - 1} \right) \nonumber \\
& \qquad = \Pr \left( y_H = 1 \mid x_T, \mathcal{D}_{T - 1} \right).
\end{align*}
The optimal decision at this final step is therefore to greedily query the point most likely to be a target, maximizing $\Pr \left( y_H = 1 \mid x_T, \mathcal{D}_{T - 1} \right)$.
For the second-to-last query, which is on $L$, the expected utility is
\begin{multline}
\mathbb{E} \bigl[ u \left( \mathcal{D}_T \setminus \mathcal{D}_{T - 2} \mid x_{T - 1}, \mathcal{D}_{T - 2} \right) \bigr] \\
= \mathbb{E} \left[ \max_{x_T} \Pr \left( y_H = 1 \mid x_T, \mathcal{D}_{T - 1} \right) \right]. \nonumber
\end{multline}
This is the expected future reward to be collected at the next and final step, which we have shown to be optimally the greedy query.
Again, at this iteration, the second-to-last $H$ query is still pending, so the expectation is taken over the joint distribution of the label of that query and that of the putative $L$ query.
In general, we compute this expected utility at iteration $i + 1 \leq T$ for an $L$ query recursively as
\begin{multline} \label{eq:ep_l}
\mathbb{E} \bigl[ u \left( \mathcal{D}_T \setminus \mathcal{D}_i \mid x_{i + 1}, \mathcal{D}_i \right) \bigr] \\
= \mathbb{E} \left[ \max_{x_{i + 2}} \mathbb{E} \bigl[ u \left( \mathcal{D}_T \setminus \mathcal{D}_{i + 1} \right) \mid x_{i + 2}, \mathcal{D}_{i + 1} \bigr] \right].
\end{multline}
For an $H$ query, this expected utility may still be recursively computed in the same manner but has a different expansion:
\begin{align} \label{eq:ep_h}
& \mathbb{E} \bigl[ u \left( \mathcal{D}_T \setminus \mathcal{D}_i \mid x_{i + 1}, \mathcal{D}_i \right) \bigr] \nonumber \\
& \qquad = \Pr \left( y_H = 1 \mid x_{i + 1}, \mathcal{D}_i \right) + \nonumber \\
& \qquad \quad ~~ \mathbb{E} \left[ \max_{x_{i + 2}} \mathbb{E} \bigl[ u \left( \mathcal{D}_T \setminus \mathcal{D}_{i + 1} \right) \mid x_{i + 2}, \mathcal{D}_{i + 1} \bigr] \right].
\end{align}
The new term in (\ref{eq:ep_h}), $\Pr \left( y_H = 1 \mid x_{i + 1}, \mathcal{D}_i \right)$, accounts for the possibility that our running reward increases if $x_{i + 1} \in \mathcal{R}$, as we are querying on fidelity $H$.
This is simply the probability that $x_{i + 1}$ is indeed a target.
Also different from (\ref{eq:ep_l}), the expectation is taken over the distribution of the label $y_H$ of the putative point $x_{i + 1}$ only,
as there is no pending query at this time.
An intuitive interpretation of the sum in (\ref{eq:ep_h}) is the balance between exploitation, the immediate reward $\Pr \left( y_H = 1 \mid x_{i + 1}, \mathcal{D}_i \right)$, and exploration, the expected future reward $u \left( \mathcal{D}_T \setminus \mathcal{D}_{i + 1} \right)$ conditioned on this putative query.
Again, the exploitation term is not present in (\ref{eq:ep_l}) when an $L$ query is made, as the query cannot possibly increase our utility immediately.

Perhaps unsurprisingly, computing the optimal policy is a daunting task: the time complexity of computing the expectation term in (\ref{eq:ep_l}) and (\ref{eq:ep_h}) is 
$\mathcal{O} \bigl( ( 4 n )^\ell \bigr)$,
where $\ell = T - i$ is the number of remaining queries and $n$ is the number of unlabeled points.
This optimal policy is computationally intractable, and suboptimal approximations are needed in practice.
One natural solution is to shorten the lookahead horizon, pretending there are only $\ell' < T - i$ iterations remaining.
This idea constitutes \emph{myopic} policies, the most straightforward of which is the greedy strategy that queries the point with the highest probability of being a target in the single-fidelity setting, setting $\ell' = 1$.
However, the design of a greedy policy for an $L$ query is not obvious, as no immediate reward can be obtained by making the query.

\subsection{Hardness of Approximation}

In addition to demonstrating the intractability of the analogous Bayesian optimal policy in the classical single-fidelity setting,
\citet{jiang2017efficient} proved that \emph{no} polynomial-time policy can approximate the optimal policy (in terms of the expected terminal utility) by \emph{any} constant factor.
This was done by constructing an adversarial family of active search problems featuring ``hidden treasures'' that are provably difficult to uncover without exponential work.
By modifying details of this construction, the performance of any polynomial-time policy can be made arbitrarily worse in expectation than that of the optimal policy.


This hardness result naturally extends to our multifidelity model, as even access to a \emph{perfectly faithful} low-fidelity oracle cannot aid a polynomial-time active search policy in one of these adversarial examples.
If we assume positives on $L$ also count towards our utility, one such active search problem with a faithful low-fidelity oracle reduces to a single-fidelity problem with a budget on $H$ increased by a factor of $(k + 1)$.
Unless $k$ is \emph{exponential} in the initial budget, any polynomial-time policy remains incapable of approximating the optimal policy.
Under our model, only positives on $H$ count towards our utility, so the performance of any policy is further reduced.
We therefore obtain the same hardness of approximation result.
However, we can still reasonably aim to design efficient approximations to the optimal policy that perform well in practice.

%% file: media/diagram.tex
\tikzset{every picture/.style={line width=0.75pt}} 

\begin{tikzpicture}[x=0.75pt,y=0.75pt,yscale=-1,xscale=1]

\draw [line width=1.5]    (97.5,130) -- (277.5,130) (142.46,123.25) -- (142.54,136.25)(187.46,123) -- (187.53,136)(232.46,122.75) -- (232.53,135.75)(277.46,122.5) -- (277.53,135.5) ;
\draw [shift={(97.5,130)}, rotate = 539.6800000000001] [color={rgb, 255:red, 0; green, 0; blue, 0 }  ][line width=1.5]    (0,6.71) -- (0,-6.71)   ;
\draw [line width=1.5]    (96.5,50) -- (277.5,50) (186.46,44) -- (186.53,57)(276.46,43.51) -- (276.53,56.51) ;
\draw [shift={(96.5,51)}, rotate = 539.6800000000001] [color={rgb, 255:red, 0; green, 0; blue, 0 }  ][line width=1.5]    (0,6.71) -- (0,-6.71)   ;
\draw [line width=1.5]  [dash pattern={on 5.63pt off 4.5pt}]  (277.5,50) -- (367.5,50) ;
\draw [line width=1.5]  [dash pattern={on 5.63pt off 4.5pt}]  (277.5,130) -- (367.5,130) ;
\draw [line width=1.5]    (367.5,50) -- (548.5,50) (457.46,43) -- (457.53,56)(547.46,42.51) -- (547.53,55.51) ;
\draw [shift={(367.5,50)}, rotate = 539.6800000000001] [color={rgb, 255:red, 0; green, 0; blue, 0 }  ][line width=1.5]    (0,6.71) -- (0,-6.71)   ;
\draw [line width=1.5]    (367.5,129) -- (502.5,129) (412.5,122.5) -- (412.5,135.5)(457.5,122.5) -- (457.5,135.5) ;
\draw [shift={(502.5,129)}, rotate = 180] [color={rgb, 255:red, 0; green, 0; blue, 0 }  ][line width=1.5]    (0,6.71) -- (0,-6.71)   ;
\draw [shift={(367.5,129)}, rotate = 180] [color={rgb, 255:red, 0; green, 0; blue, 0 }  ][line width=1.5]    (0,6.71) -- (0,-6.71)   ;
\draw [line width=1.5]    (97.5,199) -- (554.5,199) ;
\draw [shift={(557.5,199)}, rotate = 539.75] [color={rgb, 255:red, 0; green, 0; blue, 0 }  ][line width=1.5]    (14.21,-4.28) .. controls (9.04,-1.82) and (4.3,-0.39) .. (0,0) .. controls (4.3,0.39) and (9.04,1.82) .. (14.21,4.28)   ;

\draw (7,120) node [anchor=north west][inner sep=0.75pt]   [align=left] {{\fontfamily{ptm}\selectfont {\Large fidelity $L$}}};
\draw (6,40) node [anchor=north west][inner sep=0.75pt]   [align=left] {{\fontfamily{ptm}\selectfont {\Large fidelity $H$}}};
\draw (307,210) node [anchor=north west][inner sep=0.75pt]   [align=left] {{\fontfamily{ptm}\selectfont {\Large time}}};
\draw (93,16) node [anchor=north west][inner sep=0.75pt]   [align=left] {{\large 1}};
\draw (183,16) node [anchor=north west][inner sep=0.75pt]   [align=left] {{\large 4}};
\draw (271,16) node [anchor=north west][inner sep=0.75pt]   [align=left] {{\large 7}};
\draw (358,16) node [anchor=north west][inner sep=0.75pt]   [align=left] {{\large 22}};
\draw (449,16) node [anchor=north west][inner sep=0.75pt]   [align=left] {{\large 25}};
\draw (539,16) node [anchor=north west][inner sep=0.75pt]   [align=left] {{\large 28}};
\draw (92,96) node [anchor=north west][inner sep=0.75pt]   [align=left] {{\large 2}};
\draw (138,96) node [anchor=north west][inner sep=0.75pt]   [align=left] {{\large 3}};
\draw (182,96) node [anchor=north west][inner sep=0.75pt]   [align=left] {{\large 5}};
\draw (228,96) node [anchor=north west][inner sep=0.75pt]   [align=left] {{\large 6}};
\draw (269,96) node [anchor=north west][inner sep=0.75pt]   [align=left] {\begin{minipage}[lt]{9.530608pt}\setlength\topsep{0pt}
\begin{center}
{\large 8}
\end{center}

\end{minipage}};
\draw (359,96) node [anchor=north west][inner sep=0.75pt]   [align=left] {{\large 23}};
\draw (403,96) node [anchor=north west][inner sep=0.75pt]   [align=left] {{\large 24}};
\draw (449,96) node [anchor=north west][inner sep=0.75pt]   [align=left] {{\large 26}};
\draw (491,96) node [anchor=north west][inner sep=0.75pt]   [align=left] {{\large 27}};

\end{tikzpicture}

%% file: main_03_approx.tex
\section{Efficient Nonmyopic Approximation}

Our proposed algorithm is inspired by the \acro{ENS} policy, introduced by \citet{jiang2017efficient} for the single-fidelity active search setting.
\acro{ENS} offers an efficient and nonmyopic approximation to the optimal policy by assuming that after the putative query, all remaining budget will be spent \emph{simultaneously in one batch}.
Under this heuristic, the optimal decision following the putative query is to greedily construct the batch with points having the highest probabilities.
The expected utility of this batch is simply the sum of these highest probabilities due to linearity of expectation and therefore can be computed efficiently.
An interesting interpretation by \citet{jiang2017efficient} about \acro{ENS} is that it matches the optimal policy given that after the putative query, the labels of the remaining unlabeled points are conditionally independent.

\subsection{One-stage Approximation}

As a stepping stone to our proposed policy, consider the following adoption of \acro{ENS}.
We reapply the heuristic by assuming that after the putative query,
which can be on either $L$ or $H$,
all remaining $H$ queries will be made simultaneously in one batch.
Again, the optimal choice for this batch is the set of most promising points, which we will refer to as the greedy $H$ batch.
Using the summation-prime symbol $\sum\nolimits_{s}'$ to denote the sum of the top $s$ terms, we approximate the maximum expected utility of the remaining portion of the search in (\ref{eq:ep_l}) and (\ref{eq:ep_h}) as
\begin{multline} \label{eq:h-ens}
\max_{x_{i + 2}} \mathbb{E} \bigl[ u \left( \mathcal{D}_T \setminus \mathcal{D}_{i + 1} \right) \mid x_{i + 2}, \mathcal{D}_{i + 1} \bigr] \\
\approx \sum\nolimits_{\ell_H}' \Pr \left( y_H = 1 \mid x_{i + 2}, \mathcal{D}_{i + 1} \right),
\end{multline}
where $\ell_H = \lfloor (T - i - 1) / (k + 1) \rfloor$ is the number of remaining $H$ queries.
The resulting policy queries the point that maximizes the expected utility in (\ref{eq:ep_l}) and (\ref{eq:ep_h}), using (\ref{eq:h-ens}) as an approximation.
This policy is cognizant of the remaining budget on $H$ and performs well in our experiments (see appendix).
However, by assuming that the greedy $H$ batch will be queried immediately following the putative point, the strategy fails to consider future queries that could be made to the low-fidelity oracle in its lookahead;
this motivates the design of our proposed policy described below.

\subsection{Two-stage Approximation}

Our main contribution is a policy we call \acro{MF--ENS}, an efficient approximation to the optimal policy that additionally accounts for future $L$ queries.
As above, we assume in our lookahead that after the putative query, our $H$ budget will be spent simultaneously.
However, prior to committing to that final batch, we assume we may make $\overline{k} \leq k$ additional $L$ queries (also simultaneously) with the goal of improving the expected utility of the final $H$ batch.
To faithfully emulate our search model, we set $\overline{k}$ to be the number of $L$ queries remaining before the next $H$ query is made.\footnote{
When making an $H$ query, $\overline{k} = k$; 
when making an $L$ query, we subtract the number of $L$ queries since the pending $H$ query.}
We denote this batch of exploratory $L$ queries as $X_L \subset \mathcal{X} \setminus \mathcal{D}_L$ and the corresponding labels as $Y_L \in \{ 0, 1 \}^{\overline{k}}$.
In the language of \emph{approximate dynamic programming} \cite{bertsekas1995dynamic},
the policy described in the previous subsection
is a \emph{one-stage rollout} policy where the base policy selects the optimal batch on $H$ following the putative query.
\acro{MF--ENS} on the other hand is a \emph{two-stage rollout} policy whose base policy first queries the $L$ batch $X_L$, and upon observing $Y_L$, adaptively queries the updated optimal $H$ batch.

How should we construct $X_L$ to best improve the greedy $H$ batch at the second stage?
One option would be to appeal to nonmyopic policies for \emph{batch active search}
\citep{jiang2018efficient}, but the best-promising batch policies become prohibitively expensive in this context as we would need to construct a new batch for \emph{each} putative query and label.
In general, the conditional dependence among the labels in the set $Y_L$ poses a computational challenge in approximating the expected utility gained from querying a given $X_L$ batch.

Recall that in the single-fidelity setting, the \acro{ENS} policy is optimal if labels become conditionally independent after the chosen point.
Let us make a similar assumption to ease computation: we assume that labels become conditionally independent after the putative query, \emph{except} for the pair of labels (on $H$ and $L$) corresponding to each point.
That is, revealing the label $y_L$ of a point $x$ is allowed to affect our belief about the corresponding label $y_H$, but not the belief about any other label of any other point.
This structure allows for efficiently sharing information between the fidelities and enables our multistage lookahead approach.%
\footnote{This is only used in policy construction and not in inference!}


We now aim to quantify the value of querying an unlabeled point on $L$ in improving the final $H$ batch.
Our solution is motivated by a heuristic search for alternatives to the members of the
greedy $H$ batch that is assembled under the one-stage approximation.
Given a putative query, we still construct that same greedy $H$ batch.
Then, for each candidate $x$ not in the greedy $H$ batch,
we consider the expected marginal gain in utility of querying it on $L$ and modifying the membership of the $H$ batch in light of its newly revealed label $y_L$.

Recall that observing $y_L$ only changes the distribution of $y_H$ of the same $x$ under our assumption.
Let $p_\ast$ be the lowest success probability among the current $H$ batch and consider two cases.
If $\Pr \left( y_H = 1 \mid x, \mathcal{D} \right)$ \emph{exceeds} $p_\ast$ as $y_L$ is revealed, we swap out the corresponding least-promising member of the batch with $x$ and thus increase the final batch's expected utility.
Otherwise, we do not modify the current batch.
The value of querying $x$ on $L$ and observing $y_L$, denoted as $v \left( x \mid \mathcal{D} \right)$, is then
\[
v \left( x \mid \mathcal{D} \right)
\triangleq \mathbb{E}_{y_L} \bigl[ \max \left( \Pr \left( y_H = 1 \mid x, y_L,  \mathcal{D} \right) - p_*, 0 \right) \bigr].
\]
This score can be rapidly computed for most models under our conditional independence assumption.
With this value function in hand, we then greedily construct $X_L$ with the candidates having the highest values.
Another computational benefit of label independence is that $v \left( x \mid \mathcal{D} \right)$ automatically vanishes for points already labeled on $H$, as querying its $L$ label does not affect the belief of our  model and thus cannot improve the $H$ batch.
This reduces our search space in computing the exploratory batch $X_L$.

We now proceed to the last step of the rollout procedure in \acro{MF--ENS}:
marginalizing over $Y_L$, the labels of $X_L$.
As previously described, for each possible value of $Y_L$, we approximate the optimal sequence of queries following the putative one and $X_L$ with the updated greedy $H$ batch
given the newly revealed labels $Y_L$:
\begin{multline} \label{eq:test}
\max_{x_{i + 1}} \mathbb{E} \bigl[ u \left( \mathcal{D}_T \setminus \mathcal{D}_i \right) \mid x_{i + 1}, \mathcal{D}_i \bigr] \\
\approx \mathbb{E}_{Y_L} \left[ \sum\nolimits_{\ell_H}' \Pr \left( y_H = 1 \mid x, X_L, Y_L, \mathcal{D}_i \right) \right].
\end{multline}
At each iteration, \acro{MF--ENS} queries the candidate maximizing the expected utility in (\ref{eq:ep_l}) and (\ref{eq:ep_h}), as approximated by (\ref{eq:test}).
As an extension of \acro{ENS}, our approach is nonmyopic and aware of the remaining budget;
we will demonstrate the impact of this reasoning in our experiments.
The policy also factors in future $L$ queries, actively taking advantage of the ability to query the low-fidelity oracle.
We give the pseudocode for the policy in the appendix.

\subsection{Extension to Multiple Low Fidelities}

So far, we have assumed our model only consists of one exact oracle $H$ and one noisy oracle $L$.
To extend \acro{MF--ENS} to settings where there are multiple noisy oracles $\{ L_i \}$ that approximate $H$, potentially at different levels of fidelity, we still aim to design each query to maximize the expected utility on $H$, marginalizing future experiments on the $\{ L_i \}$ oracles.
Once again limiting the conditional dependence among labels to those of the same point, now between \emph{each} lower-fidelity $L_i$ and $H$, we identify a batch of appropriate size for each $L_i$, marginalize their labels, update the probabilities on $H$, and compute the approximate expected utility.
When there is only one low fidelity, this strategy reduces to the base version of \acro{MF--ENS} we have presented here.

\subsection{Implementation and Pruning}

Active search requires a classification model computing a given point's success probability with an oracle.
We extend the $k$-nearest neighbor introduced by \citet{garnett2012bayesian} to our multifidelity setting by allowing information observed on $L$ to propagate to $H$.
Specifically, when calculating the probability that an unlabeled point $x \in \mathcal{X}$ is a positive on $H$, $\Pr \left( y_H = 1 \mid x, \mathcal{D} \right)$, we take into account the revealed labels of its nearest neighbors on both fidelities, as well as its own $L$ label, $y_L$.
Effectively, we treat each given point $x \in \mathcal{X}$ as having two separate copies: one corresponding to its $H$ label, denoted as $x_H$, and one corresponding to its $L$ label, denoted as $x_L$.
Compared to the single-fidelity $k$-nearest neighbor, the set of nearest neighbors of $x_H$ is now doubled to include both copies of its original neighbors and its own copy on $L$, $x_L$.
Formally, denote $\acro{NN}_{\text{single}}(x)$ as the original nearest neighbor set of $x$.
The nearest neighbor set of $x_H$ under our multifidelity predictive model is
\begin{align*}
\acro{NN}(x_H) \triangleq
\{ x_L \} &\cup 
\bigl\{ x'_H: x' \in \acro{NN}_{\text{single}}(x) \bigr\} \\
&\cup
\bigl\{ x'_L: x' \in \acro{NN}_{\text{single}}(x) \bigr\}.
\end{align*}
To account for the unknown accuracy of the low-fidelity oracle, we apply a damping factor $q \in (0, 1)$ to the weights of the neighbors on $L$;
$q$ is dynamically set at each iteration via maximum likelihood estimation.

This model performs well in practice, is nonparametric, and can be efficiently updated in light of new data.
The last feature is essential in allowing for fast lookahead, the central component of our method.
The time complexity of a naive implementation of \acro{MF--ENS} is $\mathcal{O} \big( 2^k \, n^2 \log n \big)$,%
\footnote{Note that we are assuming $k$ is small enough that $2^k$ is a small constant; if this is not the case, we may approximate \eqref{eq:test} by sampling instead, replacing $2^k$ by $s$, the number of samples used.}
where $k$ is the number of $L$ queries made for each $H$ query and $n$ is the size of the candidate pool.
The corresponding time complexity of the single-fidelity \acro{ENS} is $\mathcal{O} \big( n^2 \log n \big)$ \cite{jiang2017efficient}, to which \acro{MF--ENS} adds a factor of $2^k$ from the exhaustive marginalization of the exploratory $L$ labels.
We may also take advantage of the implementation trick developed by \citet{jiang2017efficient},
which reduces the time complexity to $\mathcal{O} \big( 2^k \, n \, (n + m \log m) \big)$, where $m \ll n$ is the maximum number of unlabeled points whose probabilities are affected by a newly revealed label.
The interested reader may refer to \S 3.2 of \citet{jiang2017efficient} for more detail.

We also extend existing branch-and-bound pruning strategies to further reduce the computation time of our policy at each search iteration.
First, following previous work \cite{garnett2012bayesian,jiang2017efficient}, we establish an upper bound of the score function that is the approximate expected utility defined by (\ref{eq:test}).
This allows us to eliminate candidates whose score upper bounds are lower than the current best score we have found, as their actual scores cannot possibly be the final best score.
These upper bounds are computationally cheap to evaluate, so applying this pruning check only adds a trivial overhead to each search iteration.
Further, we develop an extension of this pruning strategy by making use of the fact that computing the score of a point involves marginalizing over its unknown label.
We thus apply similar pruning checks at every step during this marginalization, which allows us to identify and prune suboptimal candidates ``on the fly,'' avoiding any unnecessary computation.

Concretely, denote by $f(x)$ the score of an unlabeled point $x \in \mathcal{X}$, defined by (\ref{eq:test}) for \acro{MF--ENS}.
Suppose $x$ has a posterior probability of $\pi = \Pr \left( y = 1 \mid x, \mathcal{D} \right)$,
where $y$ is the label to be returned by the oracle we are currently querying.
As previously described, we compute $f(x)$ by calculating its \emph{partial values} while marginalizing over $y$:
\[
f(x) = \pi\,f \left( x \mid y = 1 \right) +
(1 - \pi)\, f \left( x \mid y = 0 \right),
\]
where $f(x \mid y)$ is the partial score of $x$ according to (\ref{eq:test}) when conditioned on a value of $y$.
Suppose before computing either $f(x \mid y = 0)$ or $f(x \mid y = 1)$, we know these partial scores are upper bounded by $\overline{u}(x)$ given a new positive label and $\underline{u}(x)$ given a new negative label:
\[
f(x \mid y = 1) < \overline{u}(x); \quad f(x \mid y = 0) < \underline{u}(x).
\]
As such, $f(x)$ need not be evaluated if
\begin{equation} \label{eq:full_prune}
\pi\, \overline{u}(x) + (1 - \pi)\,\underline{u}(x) < f^*,
\end{equation}
where $f^*$ is the current best score we have found.
This pruning strategy has been found to offer a significant speedup in previous work \cite{garnett2012bayesian,jiang2017efficient,jiang2018efficient}.

We extend this strategy by considering the case in which a given candidate $x$ is not pruned because (\ref{eq:full_prune}) is not satisfied, and we proceed with the calculation of $f(x)$.
Now, suppose we have computed only $f \left( x \mid y = 1 \right)$ and not yet $f \left( x \mid y = 0 \right)$ and observe that
\[
\pi\, f \left( x \mid y = 1 \right) + (1 - \pi)\, \underline{u}(x) < f^*,
\]
then we may also safely conclude that $f(x)$ cannot possibly exceed $f^*$ without needing to go further and compute $f(x \mid y = 0)$.
If this condition is met, we simply abort the computation of the current score $f(x)$ and move on to the next unpruned candidate.
For each $H$ query, we apply this partial pruning check once (either after conditioning on $y_H = 1$ and before on $y_H = 0$ or vice versa) for each candidate that is not eliminated by the full pruning check (\ref{eq:full_prune}).
For an $L$ query, we may do this at most three times for each candidate, as the marginalization over the joint distribution of the putative label and the pending $H$ label when computing $f(x)$ involves four different possible label combinations.
We quantify the effectiveness of these pruning strategies and show that they can significantly reduce the computation time of our methods in the appendix.

At an iteration where the current best score $f^*$ does not exceed the majority of the score upper bounds, many candidates may be left unpruned.
In order to help our policy avoid having to calculate the scores of a large number of candidates and consequently spending too much time on a single iteration, we place an upper limit on how many candidates are to be considered before we terminate the search.
Our approach is to first follow the lazy-evaluation strategy introduced by \citet{jiang2018efficient} and sort the candidates by their score upper bounds.
With this sorting, candidates with higher upper bounds will be considered first, and a candidate will never be considered if it will be pruned later on.
Now, at each iteration, if after having considered $u$ candidates and noticing that there are still unpruned points remaining, we simply consider a randomly selected subset of size at most $s$ of the unpruned set, before terminating the search and returning the current best candidate.
In our experiments, we set $u = s = 500$ for \acro{MF--ENS}.

We note that this strategy is only used when pruning fails to reduce our search space to be below $u + s$, and to allow us to collect results over a long horizon over many repeated experiments.
When applied in a real-life planning setting, \acro{MF--ENS} can still cover the entirety of a large pool of candidates if desired, even with a significant portion of the pool unpruned.
In our experiments, \acro{MF--ENS} takes approximately 30 seconds to reach its quota of 1000 candidates when pruning is unsuccessful;
the time for it to fully cover a pool of 100\,000 points (about the size of the real-world datasets used in our experiments) is thus well under one hour.
In short, our policy remains tractable in real-life settings, even without successful pruning.

%% file: main_04_related.tex
\section{Related Work}

\begin{table*}
\caption{
Experiment results with an $H$ budget of $t = 300$, averaged across all repeated experiments for each setting. 
$\theta$ is the simulated false positive rate of the low-fidelity oracle;
$k$ is the number of $L$ queries made between two $H$ queries (i.e., the speed ratio between the two fidelities).
Each entry denotes the average number of targets found across the repeated experiments and the corresponding standard error in parentheses.
The best performance in each column is highlighted bold.
}
\vskip 0.15in
\centering
\resizebox{\textwidth}{!}{%
\begin{tabular}{lrrrrrrrrrrrr}
\toprule
\multicolumn{1}{c}{} & \multicolumn{4}{c}{\acro{ECFP}4} & \multicolumn{4}{c}{\acro{G}pi\acro{DAPH}3} & \multicolumn{4}{c}{\acro{BMG}} \\
\cmidrule(lr){2-5} \cmidrule(lr){6-9} \cmidrule(lr){10-13}
\multicolumn{1}{c}{} 
& \multicolumn{2}{c}{$\theta = 0.1$} & \multicolumn{2}{c}{$\theta = 0.3$}
& \multicolumn{2}{c}{$\theta = 0.1$} & \multicolumn{2}{c}{$\theta = 0.3$}
& \multicolumn{2}{c}{$\theta = 0.1$} & \multicolumn{2}{c}{$\theta = 0.3$} \\
\cmidrule(lr){2-3} \cmidrule(lr){4-5} \cmidrule(lr){6-7} \cmidrule(lr){8-9} \cmidrule(lr){10-11} \cmidrule(lr){12-13}
\multicolumn{1}{c}{} & \multicolumn{1}{c}{$k = 2$} & \multicolumn{1}{c}{$k = 5$} & \multicolumn{1}{c}{$k = 2$} & \multicolumn{1}{c}{$k = 5$}
& \multicolumn{1}{c}{$k = 2$} & \multicolumn{1}{c}{$k = 5$} & \multicolumn{1}{c}{$k = 2$} & \multicolumn{1}{c}{$k = 5$}
& \multicolumn{1}{c}{$k = 2$} & \multicolumn{1}{c}{$k = 5$} & \multicolumn{1}{c}{$k = 2$} & \multicolumn{1}{c}{$k = 5$} \\
\midrule
\acro{ENS} 
& 218 (4.5) & 218 (4.5)
& 213 (4.2) & 213 (4.2)
& 197 (4.7) & 197 (4.7)
& 186 (5.3) & 186 (5.3)
& 279 (1.6) & 279 (1.6)
& 278 (1.7) & 278 (1.7) \\
\acro{MF--UCB}
& 227 (4.8) & 238 (4.4)
& 200 (5.7) & 206 (5.8)
& 200 (5.7) & 212 (5.7)
& 187 (5.5) & 200 (5.3)
& 284 (1.0) & 289 (1.0)
& 274 (2.0) & 277 (1.7) \\
\acro{UG}
& 228 (4.9) & 237 (4.4)
& 207 (5.8) & 209 (5.8)
& 204 (5.6) & 211 (5.7)
& 191 (5.6) & 202 (5.6)
& 283 (1.2) & 287 (1.1)
& 278 (2.0) & 280 (1.6) \\
\acro{MF--ENS}
& \textbf{244 (3.5)} & \textbf{250 (3.2)}
& \textbf{226 (4.5)} & \textbf{230 (4.4)}
& \textbf{230 (3.8)} & \textbf{243 (3.4)}
& \textbf{208 (5.0)} & \textbf{221 (4.3)}
& \textbf{290 (0.7)} & \textbf{294 (0.6)}
& \textbf{286 (1.1)} & \textbf{286 (1.1)} \\
\bottomrule
\end{tabular}}
\label{tab:sim-results}
\vskip -0.1in
\end{table*}

This work is an extension of the larger active search paradigm, first introduced by \citet{garnett2012bayesian}.
Active search is a variant of \emph{active learning} \cite{settles2009active} where the goal is not to learn an accurate model but to find and query positive labels.
Previous work has studied active search under different settings such as finding a given number of targets as fast/cheaply as possible \cite{warmuth2002active,warmuth2003active,jiang2019cost}, making queries in batches \cite{jiang2018efficient}, or when points have real-valued utility \cite{vanchinathan2015discovering}.
Our work generalizes \acro{ENS}, the policy proposed by \citet{jiang2017efficient} from the single-fidelity setting.
The authors of that work demonstrated that their policy is nonmyopic and aware of its exact budget, allowing it to automatically balance between exploration and exploitation during search and outperform various baselines by a large margin.
We will make the same observations about our policy.

Multifidelity active search was first examined by \citet{klyuchnikov2019figuring}, 
who specifically considered the search problem of a recommender system: identifying items that users of a given application are interested in.
They modeled predictions made by a trained preference model as output of a low-fidelity oracle and proposed a co-kriging predictive model \cite{alvarez2012kernels} to perform inference on the users' true preferences.
Under their setting, queries to the oracles are made sequentially, one after another.
Our multifidelity setting is different, modeling situations where oracles of different fidelities are available to run in parallel and vary in their response times, common in scientific experiments and testing.
Regardless, our proposed policy could be naturally adopted to their sequential model;
the only difference in the computation is that the marginalization over a pending $H$ label is no longer necessary.
Further, as we will show in later experiments, our algorithm outperforms the adoption of their upper confidence bound (\acro{UCB}) policy for various datasets.
To our knowledge, our work is the first to tackle multifidelity active search using Bayesian decision theory.

Active search is equivalent to \emph{Bayesian optimization} (\acro{BO}) \cite{brochu2010tutorial,snoek2012practical} with binary observations and cumulative reward.
Multifidelity \acro{BO} itself has been studied considerably, and policies corresponding to common acquisition functions have been adopted to multifidelity settings, including expected improvement \cite{huang2006sequential,picheny2013quantile}, knowledge gradient \cite{poloczek2017multi,wu2020practical}, and \acro{UCB} \cite{kandasamy2017multi}.
However, most of these policies are derived from or motivated by greedy approximations to the optimal policy under different utility functions, and to our knowledge, no \emph{nonmyopic} multifidelity \acro{BO} policies have been proposed.

Active search is related to the \emph{multi-armed bandit} (\acro{MAB}) problem \cite{lai1985asymptotically}.
In particular, querying the label of a point can be viewed as ``pulling an arm'' in \acro{MAB};
in active search, an arm cannot be pulled twice but is correlated to its neighbors.
\citet{kandasamy2016multi} studied a formulation of multifidelity \acro{MAB} in which each arm may be pulled on different fidelities at different costs, and proposed a policy that is a variant of \acro{UCB}.
By assuming the accuracy of the lower fidelities is known, the authors derived strong theoretical guarantees for their proposed policy.

%% file: main_05_exp.tex
\section{Experiments}

\begin{figure}
\vskip 0.2in
\centering
\input{media/utility_diff_ENS}
\caption{The difference in cumulative targets found between \acro{MF--ENS} and \acro{ENS}, averaged across all experiments and datasets.}
\label{fig:vs_single}
\vskip -0.2in
\end{figure}

\begin{figure*}
\vskip 0.2in
\centering
\input{media/queried_probs} \qquad
\input{media/utility_diff_UCB}
\caption{
An illustration of budget-awareness exhibited by \acro{MF--ENS}.
Left: The average progressive probabilities of points queried on $H$ by different active search policies.
Right: The difference in cumulative targets found between \acro{MF--ENS} and \acro{MF--UCB}.
The results are averaged across all experiments and datasets.}
\label{fig:nonmyopia}
\vskip -0.2in
\end{figure*}

We now compare the empirical performance of \acro{MF--ENS} against several benchmarks.%
\footnote{
Matlab implementations of our policies are available at:
\burl{https://github.com/KrisNguyen135/multifidelity-active-search}
.}
%
As previously described, \acro{ENS} is a state-of-the-art, nonmyopic active search policy in the single-fidelity setting.
In our experiments, this policy simply ignores the low-fidelity oracle,
serving as a single-fidelity baseline to illustrate the benefit of having access to low-fidelity queries.
We also test against the \acro{MF--UCB} policy for multifidelity active search, recently proposed by \citet{klyuchnikov2019figuring}.
Under this policy, each candidate $x$ has a \acro{UCB}--style score of $\alpha \left( x, \mathcal{D} \right) = \pi + \beta \sqrt{\pi (1 - \pi)}$, where $\pi$ is the probability of $x$ having a positive label on the fidelity being queried and $\beta$ is the exploration/exploitation tradeoff parameter.%
\footnote{The authors also considered an odd scenario in which the correlation between the two oracles is \emph{negative.} We assume that this correlation is always positive, and it is constructed to be so.}
We set $\beta = 0.01$ for $L$ queries and $\beta = 0.001$ for $H$ queries, as suggested in the same work.
For another benchmark, we consider a simple but natural heuristic for multifidelity optimization, that low-fidelity queries should serve to narrow down the most-promising search regions (exploration), so that more informed queries could be made on the higher-fidelity oracle (exploitation).
Inspired by this heuristic,
we design a policy we call \acro{UG} (for \emph{uncertainty} and \emph{greedy} sampling) that always queries the most uncertain points on $L$ and the points most likely to be positive on $H$.
Uncertainty sampling is chosen for the role of exploration due to its popularity as an active learning technique \cite{lewis1994sequential}.
\acro{UG} may also be viewed as the limit of \acro{UCB} when $\beta$ approaches infinity on $L$ for maximum exploration and 0 on $H$ for maximum exploitation.

Datasets for multifidelity active search are not readily available due to the relative novelty of the problem setting, at least not in our motivating area of scientific discovery. %
%
%
However, numerous high-quality datasets are available in the single-fidelity setting, which we will adopt and use to simulate multifidelity search.
Namely, for each dataset, we simulate the noisy labels returned by the low-fidelity oracle using the following procedure.
We first create a duplicate of the true labels.
Given $\theta \in (0, 1)$, we randomly select a fraction $\theta$ of the positives from this duplicate set and ``flip'' their labels to negative.
We also randomly select the same number of negatives and flip their values to positive.
This perturbed set of labels is then used as the $L$ labels.
This construction yields simulated $L$ labels with a false positive rate of $\theta$ and a false negative rate of $\theta \, r / (1 - r)$, where $r \in (0, 1)$ is the prevalence rate of the positive set $\mathcal{R}$ in $\mathcal{X}$.
In a typical active search problem, $\mathcal{R}$ is rare and $r \ll 1$, making the false negative rate much lower than the false positive rate, a common characteristic of many real-world scientific discovery and testing procedures.

We set $\theta \in \{ 0.1, 0.3 \}$.
We set $k$, the number of $L$ queries that are made for each $H$ query, to be either 2 or 5, and set the budget on $H$ to be $300$.
In each experiment, a policy starts with an initial training dataset of one randomly selected target whose $L$ label is also positive.

\subsection{Datasets}

We conducted experiments on three real-world scientific discovery datasets used in previous studies \citep{jiang2017efficient, jiang2018efficient, jiang2019cost}.
The first two come from drug discovery, where the goal is to discover chemical compounds exhibiting binding activity with a given protein.
Each protein defines the target for an active search problem. Here we used the first 50 proteins from the \acro{B}inding\acro{DB} database \cite{liu2007bindingdb} described by \citet{jiang2017efficient}.
A set of 100\,000 compounds sampled from the \acro{ZINC} database \cite{sterling2015zinc} served as a shared negative set.
Features for the compounds are binary vectors encoding chemical characteristics, also known as a chemoinformatic fingerprint.
We considered two fingerprints delivering good performance in previous studies: \acro{ECFP}4 and \acro{G}pi\acro{DAPH}3.
The size of the positive set for these 50 targets ranged from 205 to 1488 (mean 538), having an average prevalence rate of $r \approx 0.5 \%$.
For each of these two datasets and 50 targets, we repeat each experiment five times, for a total of 500 search simulations.

The other dataset is related to a materials science application.
The targets in this case are alloys that can form bulk metallic glasses (\acro{BMG}s), which have higher toughness and better wear resistance than crystalline alloys. 
This dataset comprises 106\,810 alloys from the materials literature \cite{kawazoe1997nonequilibrium,ward2016general}, 4275 of which exhibit glass-forming ability ($r \approx 4 \%$).
We repeated each experiment 50 times for this dataset.

We report the average number of targets found by each policy across the experiments with standard errors in parentheses in Table \ref{tab:sim-results};
each column corresponds to a specific setting of $\theta$ and $k$ under a dataset.
We observe that our policy \acro{MF--ENS} outperforms all baselines by a large margin.
%
In each column, a two-sided paired $t$--test rejects the hypothesis that the average difference in the number of targets found between \acro{MF--ENS} and any baseline is zero with overwhelming confidence, returning a $p$--value of at most $4 \times 10^{-5}$.
We report these $p$--values in the appendix.
Finally, looking across the columns, we notice the expected trends:
performance of all algorithms except for \acro{ENS}, which does not utilize fidelity $L$, improves with higher $k$ (when $L$ is cheaper) or with lower $\theta$ (when $L$ is more accurate).

\subsection{Performance Gain from Multifidelity Search}

To further examine the benefit of having access to more than just the exact oracle,
we visualize the difference in the cumulative number of targets found between \acro{MF--ENS} and the single-fidelity policy \acro{ENS}, averaged across all experiments, in Figure \ref{fig:vs_single}.
We observe that \acro{MF--ENS} completely dominates \acro{ENS}, finding roughly linearly more targets throughout the search.
This large difference in empirical performance illustrates the usefulness of the simulated low-fidelity oracle in our experiments, and suggests that our approach is likely to benefit search with any budget.

\subsection{Nonmyopic Behavior}

We have claimed that \acro{MF--ENS} is nonmyopic and aware of its remaining budget at any given time during a search.
We demonstrate this nonmyopia by first comparing the progressive probabilities of the points queried on $H$ (at the time of the queries being made) by the policy against the myopic baselines \acro{MF--UCB} and \acro{UG}, averaged across all experiments, in the left panel of Figure \ref{fig:nonmyopia}.
Initially, \acro{MF--ENS} chooses points with lower probabilities, exploring the space.
As the search progresses, \acro{MF--ENS} queries more promising points, smoothly transitioning to exploitation.
The opposite trend can be observed for \acro{UCB} and \acro{UG}, whose probabilities decrease over time due to greedy behavior.
This difference translates to distinct patterns in the cumulative reward achieved by these policies.
The right panel of Figure \ref{fig:nonmyopia} shows the difference in the cumulative number of targets found between \acro{MF--ENS} and \acro{MF--UCB}, also averaged across all experiments.
During the first half of the search, \acro{MF--ENS} appears to perform \emph{worse} than \acro{MF--UCB}, but quickly recovers and outperforms the latter in the end.
The corresponding plot comparing \acro{MF--ENS} and \acro{UG} shows a similar trend and is deferred to the appendix.

Overall, this phenomenon perfectly highlights the automatic tradeoff between exploration and exploitation exhibited by \acro{MF--ENS}:
the policy makes its initial queries to explore the search space, often requesting labels that are not the most likely to be positive and failing to collect substantial immediate reward;
however, as the budget decreases, its queries grow more exploitative and are ultimately more successful than those from myopic policies by leveraging what it has learned.

%% file: media/utility_diff_ENS.tex
\begin{tikzpicture}

\definecolor{color0}{rgb}{0.12156862745098,0.466666666666667,0.705882352941177}

\begin{axis}[
height={150},
width={230},
legend cell align={left},
legend style={fill opacity=0.8, draw opacity=1, text opacity=1, at={(0.05,0.95)}, anchor=north west, draw=none, font=\small},
tick align=outside,
tick pos=left,
x grid style={white!69.0196078431373!black},
xlabel={number of $H$ queries},
tick label style={font=\scriptsize},
label style={font=\small},
xmin=0, xmax=300,
xtick style={color=black},
y grid style={white!69.0196078431373!black},
ylabel={difference in utility},
ymin=-1.31087663520689, ymax=26.3309644725483,
ytick style={color=black},
axis x line=bottom,
axis y line=left
]
\addplot [semithick, color0]
table {%
1 -0.0366666666666666
2 -0.0149999999999999
3 0.01
4 0.0883333333333334
5 0.145
6 0.253333333333333
7 0.308333333333334
8 0.403333333333333
9 0.45
10 0.55
11 0.608333333333333
12 0.688333333333333
13 0.728333333333333
14 0.788333333333334
15 0.846666666666668
16 0.916666666666666
17 0.969999999999999
18 1.03166666666667
19 1.11833333333333
20 1.17666666666667
21 1.255
22 1.37
23 1.44833333333333
24 1.475
25 1.52166666666667
26 1.61
27 1.68333333333333
28 1.74833333333333
29 1.82166666666667
30 1.88
31 1.94
32 1.99833333333333
33 2.085
34 2.11166666666667
35 2.16333333333333
36 2.20666666666667
37 2.23666666666667
38 2.28833333333333
39 2.38666666666667
40 2.48333333333333
41 2.53
42 2.58666666666667
43 2.66166666666667
44 2.725
45 2.79833333333333
46 2.855
47 2.93
48 3.00833333333333
49 3.07666666666667
50 3.13833333333334
51 3.21
52 3.3
53 3.355
54 3.435
55 3.495
56 3.57333333333334
57 3.66
58 3.69833333333333
59 3.77833333333334
60 3.84999999999999
61 3.925
62 3.95166666666667
63 4.00333333333334
64 4.06
65 4.14833333333333
66 4.235
67 4.34333333333333
68 4.39666666666666
69 4.46333333333334
70 4.54833333333333
71 4.60833333333333
72 4.705
73 4.78166666666667
74 4.82333333333333
75 4.885
76 4.96666666666666
77 5.015
78 5.07166666666667
79 5.105
80 5.20333333333333
81 5.26333333333333
82 5.31166666666667
83 5.355
84 5.46666666666666
85 5.53833333333333
86 5.62166666666666
87 5.69333333333333
88 5.76833333333333
89 5.81
90 5.87333333333333
91 5.93833333333333
92 5.97666666666666
93 6.05333333333333
94 6.09833333333333
95 6.13000000000001
96 6.19333333333333
97 6.255
98 6.28999999999999
99 6.37833333333333
100 6.41499999999999
101 6.48833333333333
102 6.565
103 6.60833333333333
104 6.65666666666667
105 6.70833333333333
106 6.77499999999999
107 6.855
108 6.91833333333334
109 6.97666666666667
110 7.035
111 7.11333333333333
112 7.17333333333333
113 7.27833333333334
114 7.36666666666667
115 7.425
116 7.51166666666667
117 7.57333333333332
118 7.63499999999999
119 7.72833333333334
120 7.74666666666667
121 7.80499999999999
122 7.89
123 7.94666666666667
124 8.02333333333334
125 8.08166666666668
126 8.14333333333333
127 8.22499999999999
128 8.29666666666667
129 8.405
130 8.52000000000001
131 8.61499999999999
132 8.68499999999999
133 8.80500000000001
134 8.87166666666666
135 8.96333333333332
136 9.02666666666667
137 9.105
138 9.18666666666667
139 9.27
140 9.36
141 9.45
142 9.51666666666667
143 9.60833333333333
144 9.68333333333334
145 9.77499999999999
146 9.86666666666667
147 9.92833333333334
148 10
149 10.0766666666667
150 10.17
151 10.2116666666667
152 10.2733333333333
153 10.3516666666667
154 10.4183333333333
155 10.4633333333333
156 10.565
157 10.635
158 10.6633333333333
159 10.7166666666667
160 10.7983333333333
161 10.88
162 10.9433333333333
163 11.035
164 11.1083333333333
165 11.1883333333333
166 11.2766666666667
167 11.3333333333333
168 11.4483333333333
169 11.5066666666667
170 11.565
171 11.6383333333333
172 11.7116666666667
173 11.8166666666667
174 11.8683333333333
175 11.9683333333333
176 12.06
177 12.135
178 12.2566666666667
179 12.385
180 12.5
181 12.575
182 12.64
183 12.7533333333333
184 12.8316666666667
185 12.91
186 12.9766666666667
187 13.055
188 13.1866666666667
189 13.3066666666667
190 13.41
191 13.495
192 13.5833333333333
193 13.6766666666666
194 13.7633333333333
195 13.895
196 13.9933333333333
197 14.08
198 14.2066666666666
199 14.285
200 14.3633333333333
201 14.47
202 14.5916666666667
203 14.6783333333333
204 14.7583333333333
205 14.8466666666667
206 14.9133333333333
207 15.0083333333334
208 15.085
209 15.17
210 15.2933333333334
211 15.3883333333333
212 15.4616666666667
213 15.555
214 15.69
215 15.7866666666667
216 15.8683333333333
217 15.9683333333333
218 16.0216666666667
219 16.1116666666667
220 16.205
221 16.2883333333333
222 16.365
223 16.425
224 16.52
225 16.6166666666667
226 16.7283333333334
227 16.81
228 16.9216666666667
229 16.975
230 17.04
231 17.1166666666666
232 17.2016666666667
233 17.2733333333333
234 17.325
235 17.4116666666667
236 17.5116666666667
237 17.63
238 17.735
239 17.7933333333333
240 17.885
241 17.9733333333334
242 18.08
243 18.1833333333333
244 18.24
245 18.2883333333333
246 18.3633333333333
247 18.4316666666667
248 18.5083333333333
249 18.6
250 18.6733333333333
251 18.7916666666667
252 18.8733333333333
253 18.9166666666667
254 18.9866666666667
255 19.0416666666667
256 19.1216666666667
257 19.1933333333333
258 19.2433333333333
259 19.2966666666667
260 19.325
261 19.4083333333333
262 19.4666666666667
263 19.5
264 19.58
265 19.6316666666667
266 19.6733333333333
267 19.7166666666667
268 19.75
269 19.7733333333333
270 19.8066666666666
271 19.875
272 19.9066666666667
273 19.9733333333333
274 20.0333333333333
275 20.1133333333333
276 20.17
277 20.22
278 20.325
279 20.3883333333333
280 20.425
281 20.5033333333333
282 20.575
283 20.625
284 20.7066666666667
285 20.8066666666667
286 20.855
287 20.915
288 21.0316666666667
289 21.1033333333333
290 21.19
291 21.2766666666666
292 21.3666666666667
293 21.4366666666667
294 21.5266666666667
295 21.6283333333333
296 21.7133333333333
297 21.8016666666667
298 21.93
299 22.0766666666667
300 22.25
};
\addlegendentry{mean difference}
\path [fill=color0, fill opacity=0.2]
(axis cs:1,-0.0189664204232145)
--(axis cs:1,-0.0543669129101188)
--(axis cs:2,-0.0544293121271102)
--(axis cs:3,-0.0450067280735686)
--(axis cs:4,0.0162387029317626)
--(axis cs:5,0.0581789409034187)
--(axis cs:6,0.149454500118324)
--(axis cs:7,0.19076553863986)
--(axis cs:8,0.272607132305213)
--(axis cs:9,0.305988485898063)
--(axis cs:10,0.394176929057615)
--(axis cs:11,0.442421400617599)
--(axis cs:12,0.51050261706687)
--(axis cs:13,0.540523334997309)
--(axis cs:14,0.585803255525412)
--(axis cs:15,0.632551384655884)
--(axis cs:16,0.690131994221675)
--(axis cs:17,0.733462235464513)
--(axis cs:18,0.784068909098568)
--(axis cs:19,0.856941995943415)
--(axis cs:20,0.904441732025519)
--(axis cs:21,0.969785475386352)
--(axis cs:22,1.07328016049323)
--(axis cs:23,1.1412943809238)
--(axis cs:24,1.15742484354338)
--(axis cs:25,1.19123830273422)
--(axis cs:26,1.26831433110705)
--(axis cs:27,1.33010989678317)
--(axis cs:28,1.3829068768084)
--(axis cs:29,1.44327060801254)
--(axis cs:30,1.48974410341599)
--(axis cs:31,1.53678053571728)
--(axis cs:32,1.58482313920498)
--(axis cs:33,1.65753332719534)
--(axis cs:34,1.67067055873845)
--(axis cs:35,1.7071402548011)
--(axis cs:36,1.73702301595648)
--(axis cs:37,1.75754385301905)
--(axis cs:38,1.79795590807496)
--(axis cs:39,1.88295470304322)
--(axis cs:40,1.97014914182942)
--(axis cs:41,2.00538257449301)
--(axis cs:42,2.0529429463383)
--(axis cs:43,2.11720472429792)
--(axis cs:44,2.17126122464452)
--(axis cs:45,2.23419775769166)
--(axis cs:46,2.28596066845193)
--(axis cs:47,2.3489911814671)
--(axis cs:48,2.41703580679791)
--(axis cs:49,2.4785163743027)
--(axis cs:50,2.53102176799396)
--(axis cs:51,2.59143008502532)
--(axis cs:52,2.67161477193857)
--(axis cs:53,2.71724547822108)
--(axis cs:54,2.78679634742727)
--(axis cs:55,2.83603695842048)
--(axis cs:56,2.90336501954936)
--(axis cs:57,2.98123233293225)
--(axis cs:58,3.01016204754348)
--(axis cs:59,3.08087472916498)
--(axis cs:60,3.1421632227498)
--(axis cs:61,3.20759186188841)
--(axis cs:62,3.22347932214647)
--(axis cs:63,3.26318580208911)
--(axis cs:64,3.30867182251897)
--(axis cs:65,3.38629744889481)
--(axis cs:66,3.46229292715928)
--(axis cs:67,3.56065323402144)
--(axis cs:68,3.60437663626251)
--(axis cs:69,3.66128570750841)
--(axis cs:70,3.73481095515054)
--(axis cs:71,3.78338552639357)
--(axis cs:72,3.8715372536648)
--(axis cs:73,3.93726896558289)
--(axis cs:74,3.96882304008009)
--(axis cs:75,4.0188030778429)
--(axis cs:76,4.08905806769675)
--(axis cs:77,4.12985714853869)
--(axis cs:78,4.17751841032252)
--(axis cs:79,4.20201092873159)
--(axis cs:80,4.29021102887403)
--(axis cs:81,4.34052393887484)
--(axis cs:82,4.37837850451)
--(axis cs:83,4.41180249852126)
--(axis cs:84,4.51403859219618)
--(axis cs:85,4.57408373972079)
--(axis cs:86,4.65033119747559)
--(axis cs:87,4.71252869594483)
--(axis cs:88,4.77958843754069)
--(axis cs:89,4.81354870658674)
--(axis cs:90,4.86628658485498)
--(axis cs:91,4.92138005641027)
--(axis cs:92,4.95131125220943)
--(axis cs:93,5.01957314097106)
--(axis cs:94,5.05642006614203)
--(axis cs:95,5.08263849265109)
--(axis cs:96,5.13609098078365)
--(axis cs:97,5.18857847702076)
--(axis cs:98,5.21463160361336)
--(axis cs:99,5.29762121694402)
--(axis cs:100,5.32617274423504)
--(axis cs:101,5.39081663603355)
--(axis cs:102,5.46004984206003)
--(axis cs:103,5.49784395473976)
--(axis cs:104,5.53799146821084)
--(axis cs:105,5.58028304324101)
--(axis cs:106,5.64115808919673)
--(axis cs:107,5.71223097578872)
--(axis cs:108,5.76783987784427)
--(axis cs:109,5.81554784115721)
--(axis cs:110,5.86377395969213)
--(axis cs:111,5.93237079065511)
--(axis cs:112,5.98265238334077)
--(axis cs:113,6.07562396796734)
--(axis cs:114,6.15534695589596)
--(axis cs:115,6.20417885717056)
--(axis cs:116,6.281234287463)
--(axis cs:117,6.33143755611065)
--(axis cs:118,6.38453422691546)
--(axis cs:119,6.46688900037679)
--(axis cs:120,6.47570938477342)
--(axis cs:121,6.52431585014468)
--(axis cs:122,6.6019187769827)
--(axis cs:123,6.64833019360536)
--(axis cs:124,6.71592216646363)
--(axis cs:125,6.76297919641417)
--(axis cs:126,6.81534940742779)
--(axis cs:127,6.88430458608562)
--(axis cs:128,6.94657239204943)
--(axis cs:129,7.04246176271401)
--(axis cs:130,7.14836665181336)
--(axis cs:131,7.23407057917587)
--(axis cs:132,7.29253383248342)
--(axis cs:133,7.40475322323311)
--(axis cs:134,7.46052942594136)
--(axis cs:135,7.54557966010489)
--(axis cs:136,7.60439212966484)
--(axis cs:137,7.67320160827396)
--(axis cs:138,7.74698719066846)
--(axis cs:139,7.82650461135946)
--(axis cs:140,7.90861579280886)
--(axis cs:141,7.99183537872084)
--(axis cs:142,8.04987484812694)
--(axis cs:143,8.13272549858693)
--(axis cs:144,8.20174887366603)
--(axis cs:145,8.28279554539427)
--(axis cs:146,8.36456758037847)
--(axis cs:147,8.41856417804533)
--(axis cs:148,8.48137826936223)
--(axis cs:149,8.55159785592023)
--(axis cs:150,8.63658640939078)
--(axis cs:151,8.67219907392769)
--(axis cs:152,8.72767020735635)
--(axis cs:153,8.80345184879481)
--(axis cs:154,8.86008196816306)
--(axis cs:155,8.89800021550991)
--(axis cs:156,8.99179453882071)
--(axis cs:157,9.05456790800671)
--(axis cs:158,9.07734646577603)
--(axis cs:159,9.12616662275724)
--(axis cs:160,9.19951259500698)
--(axis cs:161,9.27346268983283)
--(axis cs:162,9.3273357275287)
--(axis cs:163,9.41315496517896)
--(axis cs:164,9.47431328740782)
--(axis cs:165,9.54437443582857)
--(axis cs:166,9.62123765915417)
--(axis cs:167,9.66899090385203)
--(axis cs:168,9.77621631539168)
--(axis cs:169,9.82521303268987)
--(axis cs:170,9.87257399599286)
--(axis cs:171,9.93942943217342)
--(axis cs:172,10.0049395716601)
--(axis cs:173,10.1018888389162)
--(axis cs:174,10.1450564180496)
--(axis cs:175,10.2377488912631)
--(axis cs:176,10.3232579589276)
--(axis cs:177,10.3913070968377)
--(axis cs:178,10.5055350658002)
--(axis cs:179,10.623116450316)
--(axis cs:180,10.729088362456)
--(axis cs:181,10.7914839468041)
--(axis cs:182,10.8478831270888)
--(axis cs:183,10.9532717743387)
--(axis cs:184,11.0192807298282)
--(axis cs:185,11.088172879928)
--(axis cs:186,11.1445875295639)
--(axis cs:187,11.2124201546787)
--(axis cs:188,11.333777679261)
--(axis cs:189,11.4463773589976)
--(axis cs:190,11.5409292979308)
--(axis cs:191,11.6177340432067)
--(axis cs:192,11.700456630435)
--(axis cs:193,11.7843593516763)
--(axis cs:194,11.8626930241409)
--(axis cs:195,11.9869209656184)
--(axis cs:196,12.0727986260576)
--(axis cs:197,12.1510176009549)
--(axis cs:198,12.2651564500681)
--(axis cs:199,12.3325321699963)
--(axis cs:200,12.4041039960463)
--(axis cs:201,12.5016370707749)
--(axis cs:202,12.6118994796414)
--(axis cs:203,12.6891402950301)
--(axis cs:204,12.7613460768065)
--(axis cs:205,12.8394254018772)
--(axis cs:206,12.8952581428306)
--(axis cs:207,12.981673513214)
--(axis cs:208,13.0482783060381)
--(axis cs:209,13.1224054027461)
--(axis cs:210,13.233178161261)
--(axis cs:211,13.3165885515279)
--(axis cs:212,13.3800756260841)
--(axis cs:213,13.4634847079381)
--(axis cs:214,13.5879866881546)
--(axis cs:215,13.6760163510562)
--(axis cs:216,13.7489152439519)
--(axis cs:217,13.8388025747109)
--(axis cs:218,13.8804752906575)
--(axis cs:219,13.9621493444161)
--(axis cs:220,14.0445657734295)
--(axis cs:221,14.117415067448)
--(axis cs:222,14.1833120521236)
--(axis cs:223,14.2340162348412)
--(axis cs:224,14.3194241999247)
--(axis cs:225,14.4039030749662)
--(axis cs:226,14.507777697798)
--(axis cs:227,14.5772561701696)
--(axis cs:228,14.6793662335159)
--(axis cs:229,14.7224965584507)
--(axis cs:230,14.7802367735777)
--(axis cs:231,14.8502862400546)
--(axis cs:232,14.9266260127731)
--(axis cs:233,14.9887072644227)
--(axis cs:234,15.0311890786624)
--(axis cs:235,15.1075026587822)
--(axis cs:236,15.1983718402758)
--(axis cs:237,15.3056532782103)
--(axis cs:238,15.3983143631493)
--(axis cs:239,15.4460726433055)
--(axis cs:240,15.5257301689493)
--(axis cs:241,15.604012700813)
--(axis cs:242,15.7014196991969)
--(axis cs:243,15.7977211261026)
--(axis cs:244,15.847076199967)
--(axis cs:245,15.8876014485768)
--(axis cs:246,15.9529346176495)
--(axis cs:247,16.010418907559)
--(axis cs:248,16.076504787285)
--(axis cs:249,16.1587324461922)
--(axis cs:250,16.2245952442767)
--(axis cs:251,16.3311872664625)
--(axis cs:252,16.4010068376868)
--(axis cs:253,16.4356662904946)
--(axis cs:254,16.4992897438374)
--(axis cs:255,16.5450348994986)
--(axis cs:256,16.61812776783)
--(axis cs:257,16.6827752948887)
--(axis cs:258,16.724102181719)
--(axis cs:259,16.7710776037256)
--(axis cs:260,16.7946171666112)
--(axis cs:261,16.870671760581)
--(axis cs:262,16.9222341916285)
--(axis cs:263,16.9490663886801)
--(axis cs:264,17.0206196959807)
--(axis cs:265,17.0652694444995)
--(axis cs:266,17.096319087835)
--(axis cs:267,17.1322215462086)
--(axis cs:268,17.1555818691962)
--(axis cs:269,17.1768298427246)
--(axis cs:270,17.206222270962)
--(axis cs:271,17.2658909939015)
--(axis cs:272,17.2928499009012)
--(axis cs:273,17.3516763579301)
--(axis cs:274,17.4041391954842)
--(axis cs:275,17.4739809198342)
--(axis cs:276,17.5226122700161)
--(axis cs:277,17.565009628594)
--(axis cs:278,17.6624964815811)
--(axis cs:279,17.7176380612998)
--(axis cs:280,17.7461115569788)
--(axis cs:281,17.8158748209217)
--(axis cs:282,17.8818836666284)
--(axis cs:283,17.9248355925565)
--(axis cs:284,17.9982721469069)
--(axis cs:285,18.0875093475978)
--(axis cs:286,18.1304945336286)
--(axis cs:287,18.1819165019746)
--(axis cs:288,18.2905693713957)
--(axis cs:289,18.3555039603494)
--(axis cs:290,18.4342717292283)
--(axis cs:291,18.5139107280336)
--(axis cs:292,18.5963516631717)
--(axis cs:293,18.6586605692052)
--(axis cs:294,18.7438204324039)
--(axis cs:295,18.8397728257998)
--(axis cs:296,18.9134952403322)
--(axis cs:297,18.9965464592221)
--(axis cs:298,19.1173918453038)
--(axis cs:299,19.2580259660336)
--(axis cs:300,19.4254828505314)
--(axis cs:300,25.0745171494686)
--(axis cs:300,25.0745171494686)
--(axis cs:299,24.8953073672998)
--(axis cs:298,24.7426081546962)
--(axis cs:297,24.6067868741112)
--(axis cs:296,24.5131714263345)
--(axis cs:295,24.4168938408669)
--(axis cs:294,24.3095129009294)
--(axis cs:293,24.2146727641282)
--(axis cs:292,24.1369816701616)
--(axis cs:291,24.0394226052997)
--(axis cs:290,23.9457282707717)
--(axis cs:289,23.8511627063172)
--(axis cs:288,23.7727639619376)
--(axis cs:287,23.6480834980254)
--(axis cs:286,23.5795054663714)
--(axis cs:285,23.5258239857355)
--(axis cs:284,23.4150611864265)
--(axis cs:283,23.3251644074435)
--(axis cs:282,23.2681163333716)
--(axis cs:281,23.190791845745)
--(axis cs:280,23.1038884430212)
--(axis cs:279,23.0590286053668)
--(axis cs:278,22.9875035184189)
--(axis cs:277,22.874990371406)
--(axis cs:276,22.8173877299839)
--(axis cs:275,22.7526857468324)
--(axis cs:274,22.6625274711825)
--(axis cs:273,22.5949903087365)
--(axis cs:272,22.5204834324321)
--(axis cs:271,22.4841090060985)
--(axis cs:270,22.4071110623713)
--(axis cs:269,22.3698368239421)
--(axis cs:268,22.3444181308038)
--(axis cs:267,22.3011117871247)
--(axis cs:266,22.2503475788317)
--(axis cs:265,22.1980638888338)
--(axis cs:264,22.1393803040193)
--(axis cs:263,22.0509336113199)
--(axis cs:262,22.0110991417048)
--(axis cs:261,21.9459949060856)
--(axis cs:260,21.8553828333888)
--(axis cs:259,21.8222557296077)
--(axis cs:258,21.7625644849477)
--(axis cs:257,21.703891371778)
--(axis cs:256,21.6252055655033)
--(axis cs:255,21.5382984338348)
--(axis cs:254,21.474043589496)
--(axis cs:253,21.3976670428388)
--(axis cs:252,21.3456598289799)
--(axis cs:251,21.2521460668708)
--(axis cs:250,21.1220714223899)
--(axis cs:249,21.0412675538078)
--(axis cs:248,20.9401618793817)
--(axis cs:247,20.8529144257743)
--(axis cs:246,20.7737320490172)
--(axis cs:245,20.6890652180898)
--(axis cs:244,20.632923800033)
--(axis cs:243,20.568945540564)
--(axis cs:242,20.4585803008031)
--(axis cs:241,20.3426539658537)
--(axis cs:240,20.2442698310507)
--(axis cs:239,20.1405940233612)
--(axis cs:238,20.0716856368507)
--(axis cs:237,19.9543467217897)
--(axis cs:236,19.8249614930575)
--(axis cs:235,19.7158306745511)
--(axis cs:234,19.6188109213376)
--(axis cs:233,19.557959402244)
--(axis cs:232,19.4767073205603)
--(axis cs:231,19.3830470932788)
--(axis cs:230,19.2997632264223)
--(axis cs:229,19.2275034415493)
--(axis cs:228,19.1639670998174)
--(axis cs:227,19.0427438298304)
--(axis cs:226,18.9488889688687)
--(axis cs:225,18.8294302583671)
--(axis cs:224,18.7205758000753)
--(axis cs:223,18.6159837651588)
--(axis cs:222,18.5466879478764)
--(axis cs:221,18.4592515992187)
--(axis cs:220,18.3654342265705)
--(axis cs:219,18.2611839889172)
--(axis cs:218,18.1628580426758)
--(axis cs:217,18.0978640919558)
--(axis cs:216,17.9877514227147)
--(axis cs:215,17.8973169822772)
--(axis cs:214,17.7920133118454)
--(axis cs:213,17.6465152920619)
--(axis cs:212,17.5432577072492)
--(axis cs:211,17.4600781151387)
--(axis cs:210,17.3534885054057)
--(axis cs:209,17.2175945972539)
--(axis cs:208,17.1217216939619)
--(axis cs:207,17.0349931534527)
--(axis cs:206,16.9314085238361)
--(axis cs:205,16.8539079314561)
--(axis cs:204,16.7553205898602)
--(axis cs:203,16.6675263716366)
--(axis cs:202,16.571433853692)
--(axis cs:201,16.4383629292251)
--(axis cs:200,16.3225626706204)
--(axis cs:199,16.2374678300037)
--(axis cs:198,16.1481768832652)
--(axis cs:197,16.0089823990451)
--(axis cs:196,15.9138680406091)
--(axis cs:195,15.8030790343816)
--(axis cs:194,15.6639736425258)
--(axis cs:193,15.568973981657)
--(axis cs:192,15.4662100362316)
--(axis cs:191,15.3722659567933)
--(axis cs:190,15.2790707020692)
--(axis cs:189,15.1669559743357)
--(axis cs:188,15.0395556540723)
--(axis cs:187,14.8975798453213)
--(axis cs:186,14.8087458037694)
--(axis cs:185,14.731827120072)
--(axis cs:184,14.6440526035051)
--(axis cs:183,14.553394892328)
--(axis cs:182,14.4321168729112)
--(axis cs:181,14.3585160531959)
--(axis cs:180,14.270911637544)
--(axis cs:179,14.146883549684)
--(axis cs:178,14.0077982675332)
--(axis cs:177,13.8786929031623)
--(axis cs:176,13.7967420410724)
--(axis cs:175,13.6989177754035)
--(axis cs:174,13.5916102486171)
--(axis cs:173,13.5314444944172)
--(axis cs:172,13.4183937616732)
--(axis cs:171,13.3372372344932)
--(axis cs:170,13.2574260040071)
--(axis cs:169,13.1881203006435)
--(axis cs:168,13.120450351275)
--(axis cs:167,12.9976757628146)
--(axis cs:166,12.9320956741792)
--(axis cs:165,12.8322922308381)
--(axis cs:164,12.7423533792588)
--(axis cs:163,12.656845034821)
--(axis cs:162,12.559330939138)
--(axis cs:161,12.4865373101672)
--(axis cs:160,12.3971540716597)
--(axis cs:159,12.3071667105761)
--(axis cs:158,12.2493202008906)
--(axis cs:157,12.2154320919933)
--(axis cs:156,12.1382054611793)
--(axis cs:155,12.0286664511568)
--(axis cs:154,11.9765846985036)
--(axis cs:153,11.8998814845385)
--(axis cs:152,11.8189964593103)
--(axis cs:151,11.7511342594056)
--(axis cs:150,11.7034135906092)
--(axis cs:149,11.6017354774131)
--(axis cs:148,11.5186217306378)
--(axis cs:147,11.4381024886213)
--(axis cs:146,11.3687657529549)
--(axis cs:145,11.2672044546057)
--(axis cs:144,11.1649177930006)
--(axis cs:143,11.0839411680797)
--(axis cs:142,10.9834584852064)
--(axis cs:141,10.9081646212792)
--(axis cs:140,10.8113842071911)
--(axis cs:139,10.7134953886405)
--(axis cs:138,10.6263461426649)
--(axis cs:137,10.536798391726)
--(axis cs:136,10.4489412036685)
--(axis cs:135,10.3810870065618)
--(axis cs:134,10.282803907392)
--(axis cs:133,10.2052467767669)
--(axis cs:132,10.0774661675166)
--(axis cs:131,9.99592942082413)
--(axis cs:130,9.89163334818664)
--(axis cs:129,9.76753823728598)
--(axis cs:128,9.6467609412839)
--(axis cs:127,9.56569541391438)
--(axis cs:126,9.47131725923888)
--(axis cs:125,9.40035413691916)
--(axis cs:124,9.33074450020304)
--(axis cs:123,9.24500313972798)
--(axis cs:122,9.1780812230173)
--(axis cs:121,9.08568414985532)
--(axis cs:120,9.01762394855991)
--(axis cs:119,8.98977766628988)
--(axis cs:118,8.88546577308454)
--(axis cs:117,8.81522911055601)
--(axis cs:116,8.74209904587034)
--(axis cs:115,8.64582114282944)
--(axis cs:114,8.57798637743737)
--(axis cs:113,8.48104269869933)
--(axis cs:112,8.36401428332589)
--(axis cs:111,8.29429587601155)
--(axis cs:110,8.20622604030788)
--(axis cs:109,8.13778549217613)
--(axis cs:108,8.06882678882239)
--(axis cs:107,7.99776902421128)
--(axis cs:106,7.90884191080327)
--(axis cs:105,7.83638362342565)
--(axis cs:104,7.7753418651225)
--(axis cs:103,7.71882271192691)
--(axis cs:102,7.66995015793997)
--(axis cs:101,7.58585003063311)
--(axis cs:100,7.50382725576496)
--(axis cs:99,7.45904544972265)
--(axis cs:98,7.36536839638664)
--(axis cs:97,7.32142152297924)
--(axis cs:96,7.25057568588302)
--(axis cs:95,7.17736150734891)
--(axis cs:94,7.14024660052464)
--(axis cs:93,7.0870935256956)
--(axis cs:92,7.00202208112391)
--(axis cs:91,6.9552866102564)
--(axis cs:90,6.88038008181169)
--(axis cs:89,6.80645129341326)
--(axis cs:88,6.75707822912598)
--(axis cs:87,6.67413797072184)
--(axis cs:86,6.59300213585775)
--(axis cs:85,6.50258292694588)
--(axis cs:84,6.41929474113715)
--(axis cs:83,6.29819750147874)
--(axis cs:82,6.24495482882333)
--(axis cs:81,6.18614272779183)
--(axis cs:80,6.11645563779263)
--(axis cs:79,6.00798907126841)
--(axis cs:78,5.96581492301082)
--(axis cs:77,5.90014285146131)
--(axis cs:76,5.84427526563659)
--(axis cs:75,5.7511969221571)
--(axis cs:74,5.67784362658658)
--(axis cs:73,5.62606436775044)
--(axis cs:72,5.5384627463352)
--(axis cs:71,5.4332811402731)
--(axis cs:70,5.36185571151613)
--(axis cs:69,5.26538095915826)
--(axis cs:68,5.18895669707083)
--(axis cs:67,5.12601343264523)
--(axis cs:66,5.00770707284072)
--(axis cs:65,4.91036921777186)
--(axis cs:64,4.81132817748103)
--(axis cs:63,4.74348086457755)
--(axis cs:62,4.67985401118686)
--(axis cs:61,4.64240813811159)
--(axis cs:60,4.5578367772502)
--(axis cs:59,4.47579193750169)
--(axis cs:58,4.38650461912319)
--(axis cs:57,4.33876766706775)
--(axis cs:56,4.24330164711731)
--(axis cs:55,4.15396304157952)
--(axis cs:54,4.08320365257273)
--(axis cs:53,3.99275452177892)
--(axis cs:52,3.92838522806143)
--(axis cs:51,3.82856991497468)
--(axis cs:50,3.74564489867271)
--(axis cs:49,3.67481695903063)
--(axis cs:48,3.59963085986875)
--(axis cs:47,3.5110088185329)
--(axis cs:46,3.42403933154807)
--(axis cs:45,3.362468908975)
--(axis cs:44,3.27873877535548)
--(axis cs:43,3.20612860903541)
--(axis cs:42,3.12039038699504)
--(axis cs:41,3.05461742550699)
--(axis cs:40,2.99651752483724)
--(axis cs:39,2.89037863029011)
--(axis cs:38,2.77871075859171)
--(axis cs:37,2.71578948031428)
--(axis cs:36,2.67631031737685)
--(axis cs:35,2.61952641186556)
--(axis cs:34,2.55266277459488)
--(axis cs:33,2.51246667280466)
--(axis cs:32,2.41184352746169)
--(axis cs:31,2.34321946428272)
--(axis cs:30,2.27025589658401)
--(axis cs:29,2.2000627253208)
--(axis cs:28,2.11375978985827)
--(axis cs:27,2.0365567698835)
--(axis cs:26,1.95168566889295)
--(axis cs:25,1.85209503059912)
--(axis cs:24,1.79257515645662)
--(axis cs:23,1.75537228574286)
--(axis cs:22,1.66671983950677)
--(axis cs:21,1.54021452461365)
--(axis cs:20,1.44889160130781)
--(axis cs:19,1.37972467072325)
--(axis cs:18,1.27926442423477)
--(axis cs:17,1.20653776453549)
--(axis cs:16,1.14320133911166)
--(axis cs:15,1.06078194867745)
--(axis cs:14,0.990863411141254)
--(axis cs:13,0.916143331669358)
--(axis cs:12,0.866164049599796)
--(axis cs:11,0.774245266049068)
--(axis cs:10,0.705823070942385)
--(axis cs:9,0.594011514101937)
--(axis cs:8,0.534059534361453)
--(axis cs:7,0.425901128026807)
--(axis cs:6,0.357212166548343)
--(axis cs:5,0.231821059096581)
--(axis cs:4,0.160427963734904)
--(axis cs:3,0.0650067280735686)
--(axis cs:2,0.0244293121271102)
--(axis cs:1,-0.0189664204232145)
--cycle;
\addlegendimage{area legend, fill=color0, fill opacity=0.2}
\addlegendentry{95\% \acro{CI}}

\addplot [semithick, white!50.1960784313725!black, dashed, forget plot]
table {%
1 0
2 0
3 0
4 0
5 0
6 0
7 0
8 0
9 0
10 0
11 0
12 0
13 0
14 0
15 0
16 0
17 0
18 0
19 0
20 0
21 0
22 0
23 0
24 0
25 0
26 0
27 0
28 0
29 0
30 0
31 0
32 0
33 0
34 0
35 0
36 0
37 0
38 0
39 0
40 0
41 0
42 0
43 0
44 0
45 0
46 0
47 0
48 0
49 0
50 0
51 0
52 0
53 0
54 0
55 0
56 0
57 0
58 0
59 0
60 0
61 0
62 0
63 0
64 0
65 0
66 0
67 0
68 0
69 0
70 0
71 0
72 0
73 0
74 0
75 0
76 0
77 0
78 0
79 0
80 0
81 0
82 0
83 0
84 0
85 0
86 0
87 0
88 0
89 0
90 0
91 0
92 0
93 0
94 0
95 0
96 0
97 0
98 0
99 0
100 0
101 0
102 0
103 0
104 0
105 0
106 0
107 0
108 0
109 0
110 0
111 0
112 0
113 0
114 0
115 0
116 0
117 0
118 0
119 0
120 0
121 0
122 0
123 0
124 0
125 0
126 0
127 0
128 0
129 0
130 0
131 0
132 0
133 0
134 0
135 0
136 0
137 0
138 0
139 0
140 0
141 0
142 0
143 0
144 0
145 0
146 0
147 0
148 0
149 0
150 0
151 0
152 0
153 0
154 0
155 0
156 0
157 0
158 0
159 0
160 0
161 0
162 0
163 0
164 0
165 0
166 0
167 0
168 0
169 0
170 0
171 0
172 0
173 0
174 0
175 0
176 0
177 0
178 0
179 0
180 0
181 0
182 0
183 0
184 0
185 0
186 0
187 0
188 0
189 0
190 0
191 0
192 0
193 0
194 0
195 0
196 0
197 0
198 0
199 0
200 0
201 0
202 0
203 0
204 0
205 0
206 0
207 0
208 0
209 0
210 0
211 0
212 0
213 0
214 0
215 0
216 0
217 0
218 0
219 0
220 0
221 0
222 0
223 0
224 0
225 0
226 0
227 0
228 0
229 0
230 0
231 0
232 0
233 0
234 0
235 0
236 0
237 0
238 0
239 0
240 0
241 0
242 0
243 0
244 0
245 0
246 0
247 0
248 0
249 0
250 0
251 0
252 0
253 0
254 0
255 0
256 0
257 0
258 0
259 0
260 0
261 0
262 0
263 0
264 0
265 0
266 0
267 0
268 0
269 0
270 0
271 0
272 0
273 0
274 0
275 0
276 0
277 0
278 0
279 0
280 0
281 0
282 0
283 0
284 0
285 0
286 0
287 0
288 0
289 0
290 0
291 0
292 0
293 0
294 0
295 0
296 0
297 0
298 0
299 0
300 0
};
\end{axis}

\end{tikzpicture}

%% file: media/queried_probs.tex
\begin{tikzpicture}

\begin{axis}[
height={150},
width={230},
legend cell align={left},
legend style={fill opacity=0.8, draw opacity=1, text opacity=1, at={(0.97,0.03)}, anchor=south east, draw=none, font=\small},
tick align=outside,
tick pos=left,
x grid style={white!69.0196078431373!black},
xlabel={number of $H$ queries},
xmin=0, xmax=300,
xtick style={color=black},
y grid style={white!69.0196078431373!black},
ylabel={queried probability},
ymin=0.449421463236954, ymax=0.912803469261272,
ytick style={color=black},
tick label style={font=\scriptsize},
label style={font=\small},
axis x line=bottom,
axis y line=left
]
\addplot [semithick, cyan]
table {%
1 0.497180048860589
2 0.470484281692605
3 0.511916818192494
4 0.55158277735688
5 0.576779438176984
6 0.588627271931334
7 0.608218243232119
8 0.627415240375353
9 0.629850582544636
10 0.655325927495655
11 0.660608253345991
12 0.669442862655414
13 0.67720476467931
14 0.683506694096266
15 0.685626949612675
16 0.692318813146733
17 0.693739742822051
18 0.700819611108498
19 0.706608492886861
20 0.710736899727307
21 0.716423448622354
22 0.719295047597515
23 0.724027858892554
24 0.728725064092114
25 0.73535249372054
26 0.739731979155394
27 0.73702764739456
28 0.738618351230375
29 0.742938261939482
30 0.752449847952149
31 0.752308536342901
32 0.754532994140002
33 0.755297683632527
34 0.75742717242356
35 0.758439981545302
36 0.763076734842773
37 0.759844259225924
38 0.76013843672151
39 0.767175562781661
40 0.766309190768168
41 0.76867156538466
42 0.767221076617044
43 0.770535990380568
44 0.776806486593675
45 0.77775646900849
46 0.777033542111336
47 0.778470035351024
48 0.782315394999886
49 0.782119907429613
50 0.784004783745322
51 0.785800822157389
52 0.78671131235655
53 0.789036451416737
54 0.790036622645826
55 0.792467603307674
56 0.792287908000117
57 0.795665694801599
58 0.79621736387455
59 0.798725924134495
60 0.797961301218867
61 0.795694191141637
62 0.796296812157382
63 0.797281057167212
64 0.796352783444936
65 0.802789403521747
66 0.79984796371317
67 0.801877813856158
68 0.799746411425285
69 0.800226392850354
70 0.803132908450687
71 0.802922880834555
72 0.804882101293609
73 0.800076655897031
74 0.802390843445107
75 0.803450363854738
76 0.801419345791745
77 0.806995808316419
78 0.806046684259693
79 0.800802802294262
80 0.800643469568309
81 0.804365828010498
82 0.807523233481029
83 0.806103759813407
84 0.811943533777326
85 0.812026196447862
86 0.814827279637798
87 0.812575530207507
88 0.814236479461921
89 0.815591081283403
90 0.816058059020949
91 0.81379203796827
92 0.812993815734056
93 0.813773284733218
94 0.813965643592865
95 0.815834902492174
96 0.82037759092173
97 0.816721375065253
98 0.820607013911342
99 0.819119305632405
100 0.819038048366138
101 0.818628535182205
102 0.825178185682767
103 0.817771768660637
104 0.82222136824655
105 0.820459156911789
106 0.821868532159602
107 0.820125459328838
108 0.826117426620598
109 0.82231957319527
110 0.82476623096984
111 0.823118625337194
112 0.826081649865813
113 0.828401659840299
114 0.824502627494912
115 0.820776410083392
116 0.825800693828742
117 0.822140226601747
118 0.824158609796331
119 0.830069012092957
120 0.823533071205678
121 0.826944040697527
122 0.825531412354664
123 0.82971676825062
124 0.831814390244786
125 0.83319631652177
126 0.833842507754989
127 0.830542997043995
128 0.830190851725812
129 0.83006905918816
130 0.833440872119039
131 0.835620234925065
132 0.831642913385784
133 0.835836978349658
134 0.833948607925803
135 0.836164736725251
136 0.833663269554992
137 0.837692659536473
138 0.836821173472596
139 0.838888785416112
140 0.839249374113812
141 0.840558837911889
142 0.837603791784381
143 0.841557024758122
144 0.84295415865783
145 0.839741088357012
146 0.839109462903301
147 0.837097272386269
148 0.838882443405004
149 0.84350997638078
150 0.839810523536544
151 0.848348616430528
152 0.844636497286929
153 0.849358206756799
154 0.84426166622858
155 0.84801716472442
156 0.851028564877175
157 0.844966233141936
158 0.840747527237694
159 0.846987586473645
160 0.845367172363117
161 0.849724113080226
162 0.841446632258318
163 0.852049019961464
164 0.852862960345107
165 0.850029783240978
166 0.850697050088165
167 0.84984396651064
168 0.855577498904531
169 0.846715336758685
170 0.847739557307906
171 0.853140914040877
172 0.850279800857543
173 0.85312245314709
174 0.852148964607242
175 0.8489520222595
176 0.846560649871338
177 0.853605496892924
178 0.854580387293005
179 0.856996835997186
180 0.854695119835919
181 0.849339614171737
182 0.852700304679693
183 0.846875900780158
184 0.854745252783046
185 0.852475518249547
186 0.847117964158311
187 0.851402614909965
188 0.858821488239639
189 0.854597291604295
190 0.854465749759175
191 0.855221909342434
192 0.855315821383744
193 0.853432570392405
194 0.850106517042591
195 0.858163995542731
196 0.853556052894108
197 0.851578868348028
198 0.852570102559591
199 0.855861528788187
200 0.853677932340663
201 0.859487596776748
202 0.855204674549005
203 0.859381852845158
204 0.861411320512508
205 0.857335323496996
206 0.857088639385027
207 0.854090775717664
208 0.857847842507941
209 0.862517476564861
210 0.861284673934873
211 0.858078389417087
212 0.860398991585464
213 0.860913640896403
214 0.862918365284818
215 0.862904035441157
216 0.861095263160845
217 0.866353957524107
218 0.863115386749318
219 0.865980484297547
220 0.865961462142901
221 0.865690769320982
222 0.86627787095194
223 0.863355819546629
224 0.867612214823949
225 0.868818711965152
226 0.870080977966355
227 0.866136570992285
228 0.867930866833828
229 0.866487909606825
230 0.867257386150933
231 0.869204141160649
232 0.86779820619813
233 0.86760902998783
234 0.870173397882358
235 0.869524440290954
236 0.874820473592066
237 0.875220535225315
238 0.86795868046608
239 0.874255658392919
240 0.871993307368473
241 0.875348925190717
242 0.872639438985345
243 0.873696592758261
244 0.876672159577301
245 0.877139185421679
246 0.87089022738226
247 0.875698518208521
248 0.877908748326276
249 0.874836375791539
250 0.882667719408326
251 0.884828258142765
252 0.882508640716584
253 0.878730050302219
254 0.882085484880668
255 0.88178304604958
256 0.882727208660314
257 0.883138171023963
258 0.880692559456283
259 0.880629278765526
260 0.878319042486477
261 0.883264795098041
262 0.878955299051484
263 0.88102571603022
264 0.88371499564569
265 0.882371180734319
266 0.88690483060892
267 0.889094904405602
268 0.885659636188577
269 0.891740650805621
270 0.886956420354922
271 0.884793606955716
272 0.88576317419988
273 0.889659046721816
274 0.881348494594535
275 0.88642671045495
276 0.884542807015226
277 0.883281014095125
278 0.887357769362807
279 0.880771174977661
280 0.878356776799462
281 0.881506353278847
282 0.883302354676296
283 0.880019673761083
284 0.878783761651122
285 0.881342699730696
286 0.878516724465974
287 0.881698387325159
288 0.876313563100119
289 0.874135897699361
290 0.881343690163019
291 0.877535449572073
292 0.873548220312855
293 0.87020547660378
294 0.868804132354683
295 0.870405606715261
296 0.865006965863972
297 0.85738824930764
298 0.857905486461073
299 0.852122261831169
300 0.840624508105779
};
\addlegendentry{\acro{MF-ENS}}
\addplot [semithick, orange]
table {%
1 0.623251804770436
2 0.736195591749159
3 0.77122386033899
4 0.783416579334733
5 0.796614400199568
6 0.80358045602563
7 0.809613782238236
8 0.814802673424875
9 0.817754278459613
10 0.819395730208419
11 0.823347998323504
12 0.825423408592813
13 0.826621528808402
14 0.827168209793896
15 0.82795247636693
16 0.828425792463881
17 0.829339566908492
18 0.829691578882078
19 0.829174656897269
20 0.829916708253305
21 0.829611374130797
22 0.829958484560227
23 0.8302088656056
24 0.828954297132147
25 0.82676137875276
26 0.825951539377831
27 0.8283116171428
28 0.827299437495953
29 0.828145149009966
30 0.827668367496185
31 0.827240028944426
32 0.827011069353083
33 0.826253199014695
34 0.826703165068386
35 0.826197349849915
36 0.82573882325726
37 0.825941557383927
38 0.82456603664311
39 0.822786756104453
40 0.823674212963363
41 0.823729987368547
42 0.823472200996757
43 0.82335820144651
44 0.822912560337072
45 0.822821828453504
46 0.823158172679579
47 0.824467533789673
48 0.825469152727392
49 0.825555843958906
50 0.826157449704339
51 0.826625848015934
52 0.827074885181784
53 0.827017587036051
54 0.826548799063308
55 0.82563510486292
56 0.825423344672767
57 0.824776758603601
58 0.825045065228648
59 0.823496322124425
60 0.823790835797588
61 0.823772186175361
62 0.82383885096443
63 0.824427512760133
64 0.824667086726457
65 0.824546512474842
66 0.823385156868834
67 0.822489666759867
68 0.821915606559169
69 0.822056591680322
70 0.821405649275202
71 0.819902603697027
72 0.819811911704934
73 0.820120556077103
74 0.820550014255084
75 0.820196464436141
76 0.819773288999546
77 0.819120231855941
78 0.817927255664225
79 0.817984348685397
80 0.817903170215872
81 0.816804168012975
82 0.815264089511233
83 0.814192226391269
84 0.814153722077329
85 0.814157756742771
86 0.813256966415544
87 0.81242946813948
88 0.812742788690838
89 0.811471859444016
90 0.811406559471055
91 0.810570747297613
92 0.811424154618754
93 0.811196294551763
94 0.810508936556229
95 0.810865611514618
96 0.811733627268486
97 0.813291734302756
98 0.813544394262312
99 0.81351896275483
100 0.814559678394171
101 0.814608204339799
102 0.814073376043937
103 0.813879473146124
104 0.8163819356707
105 0.814998986311422
106 0.814137602996859
107 0.814731284295663
108 0.814312382632989
109 0.81407499182611
110 0.814467714549562
111 0.814105854597811
112 0.814948770572176
113 0.814638221866731
114 0.813962048011064
115 0.813766153288974
116 0.813621666019375
117 0.813722706710245
118 0.813392794536422
119 0.813784330438744
120 0.812988832826377
121 0.812213483424463
122 0.812609134859263
123 0.812086720880826
124 0.812112958296808
125 0.810921708208269
126 0.810418272859378
127 0.809371511336518
128 0.808528277657467
129 0.808073231535733
130 0.806845309429806
131 0.805359909947409
132 0.803250096456119
133 0.801836184923006
134 0.800078189259446
135 0.799017571249995
136 0.798470652278683
137 0.797930645086339
138 0.798321767504112
139 0.798546835459638
140 0.798744017090488
141 0.798227879277808
142 0.797785113108416
143 0.797624183532975
144 0.796518667867933
145 0.795650572867384
146 0.795500125093768
147 0.794910060341091
148 0.794854403858286
149 0.794841910364477
150 0.794315162857748
151 0.794051518708935
152 0.795014397563192
153 0.795664793710426
154 0.795356553045628
155 0.793792649333035
156 0.794298853003533
157 0.794486816179538
158 0.795333270286065
159 0.794394285513993
160 0.793621688492882
161 0.79277113876399
162 0.792916218776352
163 0.792270165051014
164 0.79144835453436
165 0.791846026512016
166 0.792292056749767
167 0.791591163451318
168 0.791058946488085
169 0.790683126729001
170 0.790350103068456
171 0.789140526975136
172 0.788162464576539
173 0.787127077951373
174 0.786349148189791
175 0.785568708400338
176 0.784710679463051
177 0.784380627152726
178 0.783796012600127
179 0.783278349976632
180 0.783900618956823
181 0.782589074242137
182 0.782237767305434
183 0.781650164695588
184 0.781613236908542
185 0.781511201899152
186 0.780515670375555
187 0.779806156308865
188 0.780048889012351
189 0.780710688438168
190 0.780792838705811
191 0.781344419031734
192 0.780861413717521
193 0.780410363711466
194 0.780932743477837
195 0.780187022570379
196 0.780507052293096
197 0.779691395298145
198 0.778756396882641
199 0.778564485091891
200 0.778540231826185
201 0.777424499076074
202 0.776844377661194
203 0.775848544947704
204 0.77677055788466
205 0.776341555415236
206 0.776242701155022
207 0.775975671634482
208 0.775686048559129
209 0.77513998755876
210 0.775506331022561
211 0.775017475146005
212 0.775301950579304
213 0.775534483544463
214 0.77516684038719
215 0.775423378944501
216 0.775056909777227
217 0.775226592101658
218 0.775019462002615
219 0.775384136659732
220 0.774566554258976
221 0.774443653051397
222 0.774484308667189
223 0.773530652106802
224 0.773360325851081
225 0.773646729209348
226 0.772484609040873
227 0.771888461924878
228 0.771438007316839
229 0.770827528600756
230 0.769920009877903
231 0.769149509127888
232 0.768454743307282
233 0.768938545019299
234 0.769350615826863
235 0.76944195437152
236 0.768825074141398
237 0.767893232115054
238 0.766241716371623
239 0.765922536182869
240 0.765224829432832
241 0.763493573592251
242 0.762866710692449
243 0.762272565133424
244 0.762026338053681
245 0.761400872876518
246 0.761248931821576
247 0.761086630800808
248 0.76010615130924
249 0.759526565765248
250 0.759059046365805
251 0.757459473678294
252 0.755759802437373
253 0.755575379148154
254 0.75501132601102
255 0.754617489554509
256 0.75389294698109
257 0.754136674775669
258 0.754190879284177
259 0.75466850223012
260 0.754617182376002
261 0.755202277301131
262 0.754871282016437
263 0.754405367141401
264 0.753524492470537
265 0.75309037207602
266 0.752605730533363
267 0.751901369371783
268 0.751459282889252
269 0.751853387335309
270 0.751182314737558
271 0.751331993063609
272 0.750082797146336
273 0.750097100894414
274 0.750727965347808
275 0.749108497536617
276 0.747854056542891
277 0.747351886299796
278 0.746371940384713
279 0.745883330276943
280 0.744414605237313
281 0.743945875393655
282 0.743381589683406
283 0.7434807076142
284 0.742263797029792
285 0.741267950488317
286 0.740216880195708
287 0.738955972542722
288 0.737723841952331
289 0.736994340525698
290 0.736517321944298
291 0.736448592369496
292 0.736047657650734
293 0.735569211869653
294 0.735483623314096
295 0.735439964875642
296 0.736054738104496
297 0.735343119752435
298 0.734000514029461
299 0.733963628432123
300 0.734443058787893
};
\addlegendentry{\acro{MF-UCB}}
\addplot [semithick, orange, dashed]
table {%
1 0.623251804770436
2 0.714210852509547
3 0.738497896317546
4 0.754528107638415
5 0.770843395504049
6 0.780986163317454
7 0.790764969179748
8 0.799745749413492
9 0.803480521168059
10 0.808191188721346
11 0.813054749817441
12 0.815527897093469
13 0.818016691617041
14 0.81919693478685
15 0.819778185866447
16 0.822621457195524
17 0.825743373105323
18 0.826622409116083
19 0.824781333479262
20 0.825010275151949
21 0.825762004448431
22 0.825900343397589
23 0.826334897812328
24 0.827454900194219
25 0.827822536098315
26 0.827247125366495
27 0.827722575106463
28 0.825815449840148
29 0.82559662593877
30 0.826140177816737
31 0.824847013257834
32 0.823366908745487
33 0.821769228151153
34 0.821789500241435
35 0.820482336184756
36 0.819636521889502
37 0.81920802081862
38 0.817965655292688
39 0.815905596304258
40 0.816022605282831
41 0.816957408756326
42 0.81732794407752
43 0.817630071803761
44 0.818658374920209
45 0.819617084109031
46 0.819858923446268
47 0.820167595187374
48 0.818826178782002
49 0.818742843277726
50 0.818645820097483
51 0.82001201100352
52 0.820365753630146
53 0.819684264964867
54 0.818908878219206
55 0.818592668731276
56 0.816783624633019
57 0.816113672560646
58 0.816878879919573
59 0.815762294149887
60 0.815300136354365
61 0.815584471437676
62 0.816516400201673
63 0.816471020135063
64 0.816231784649604
65 0.815468845030474
66 0.815242361088917
67 0.81499493615998
68 0.814506104871131
69 0.815278167746557
70 0.813668999200155
71 0.813329255534763
72 0.812060406073241
73 0.811682637155162
74 0.810882100623311
75 0.8097245419425
76 0.808516412423752
77 0.807451159473456
78 0.806954161307133
79 0.805722811323581
80 0.806846171518311
81 0.805505110929205
82 0.805571921545786
83 0.804991915394924
84 0.804509929956501
85 0.805373201208842
86 0.80613177640722
87 0.806711908836749
88 0.807383689886188
89 0.806932487362994
90 0.806403371515183
91 0.805296718171539
92 0.805103627628315
93 0.804801839036295
94 0.80412128168908
95 0.805033889551588
96 0.804886268551548
97 0.805065649675352
98 0.805236683635128
99 0.806189622982962
100 0.806160826540534
101 0.807148832601666
102 0.807403520295062
103 0.809376287926194
104 0.811208013149845
105 0.810413854354561
106 0.810212691973919
107 0.810790192385163
108 0.810319736551848
109 0.811037872797431
110 0.810802973205483
111 0.81153246155627
112 0.810278491066148
113 0.810877971849036
114 0.810236063177936
115 0.809851627073178
116 0.809696683958324
117 0.808174624315704
118 0.807808594985913
119 0.808290233649943
120 0.807474475102685
121 0.807145971138263
122 0.806394738692693
123 0.806161149233131
124 0.804962312742892
125 0.804169992515623
126 0.803668235836808
127 0.803747508761688
128 0.802164953422322
129 0.801092351069194
130 0.800610819113392
131 0.799230389647535
132 0.797581066818914
133 0.797474947675974
134 0.797077656450018
135 0.79684187723626
136 0.796742104176359
137 0.796947378105569
138 0.796087757835579
139 0.79555617943875
140 0.796079812429126
141 0.794988577292302
142 0.794516363173666
143 0.794093673819371
144 0.79326634611928
145 0.792134727432748
146 0.792230236219044
147 0.792131386856696
148 0.791307415214957
149 0.790894668162546
150 0.791450705435349
151 0.791194182722513
152 0.791563604068854
153 0.790886653542404
154 0.791262122885754
155 0.792152824044925
156 0.790510244721268
157 0.79017661205033
158 0.789614555958536
159 0.788953266652293
160 0.789333377457613
161 0.789210564021662
162 0.789650119414993
163 0.790179545801259
164 0.790668909316246
165 0.790163509132568
166 0.790217090774236
167 0.790322031501665
168 0.789473763928546
169 0.788427295455909
170 0.788251639403671
171 0.788711150381388
172 0.788883446228098
173 0.78833003619639
174 0.787749113479551
175 0.788044298140028
176 0.787624156776742
177 0.787243859717177
178 0.787526227873817
179 0.787452280392168
180 0.786462677779178
181 0.786323743940696
182 0.787082052941065
183 0.787481496490076
184 0.786131885482141
185 0.785431719774614
186 0.785381117096372
187 0.783952094033726
188 0.782896405831908
189 0.782091172153717
190 0.781583598824927
191 0.781140536335247
192 0.781039039473284
193 0.78071893399314
194 0.780582483217525
195 0.779928583896059
196 0.778388523211616
197 0.77708753809011
198 0.776474848969467
199 0.776338399076868
200 0.77670717051163
201 0.775964298776278
202 0.77611879736896
203 0.776514704812801
204 0.775963133912997
205 0.776721779093265
206 0.776384184037124
207 0.775533302389014
208 0.774436773866801
209 0.773553368932987
210 0.772418134735429
211 0.771805748455163
212 0.771426437663504
213 0.770414622286309
214 0.770252010200769
215 0.769965597299838
216 0.770688208255911
217 0.769198178929797
218 0.768923989573924
219 0.768708222820257
220 0.768671836142694
221 0.768422515859494
222 0.768484575735557
223 0.767923333310552
224 0.76761229009305
225 0.767600470386078
226 0.767012250525596
227 0.766539353465286
228 0.765780901414873
229 0.764643837732729
230 0.762858600633958
231 0.762172248969778
232 0.763179328563475
233 0.762573936739724
234 0.7607499199606
235 0.76028559064646
236 0.760205877612577
237 0.760525279737982
238 0.760280826184173
239 0.76046710474164
240 0.761375424181407
241 0.761259634601601
242 0.761817397078654
243 0.761938732735184
244 0.76091962956708
245 0.760375496911071
246 0.760009542604372
247 0.758441710763904
248 0.75792546569279
249 0.757683679760025
250 0.757631300842263
251 0.757052588983668
252 0.755932555603922
253 0.75540739108787
254 0.75521505450042
255 0.754490529491161
256 0.754609546516626
257 0.75429382201394
258 0.753395024561594
259 0.753076512571902
260 0.752333341116324
261 0.752080492154282
262 0.750069621403857
263 0.749042775683852
264 0.747248516189327
265 0.746319642960187
266 0.744638835962205
267 0.74300159748953
268 0.742249946831401
269 0.741248330365364
270 0.740955430030263
271 0.740744510695713
272 0.740524222225033
273 0.740243354078307
274 0.739906346445716
275 0.739012850984357
276 0.739377336136416
277 0.738843669441802
278 0.737415073363234
279 0.736662163029765
280 0.735494339744975
281 0.734778612693415
282 0.733831514358212
283 0.732713195277347
284 0.73165681724618
285 0.7314720138941
286 0.730073214693292
287 0.728925001768765
288 0.727406705722327
289 0.727058304576222
290 0.727182596262309
291 0.726150328987585
292 0.726617956255864
293 0.726401889818022
294 0.725851804016142
295 0.724813792139869
296 0.724324550752585
297 0.72403063059788
298 0.72279588692682
299 0.721946327091299
300 0.720652843016745
};
\addlegendentry{\acro{UG}}
\end{axis}

\end{tikzpicture}

%% file: media/utility_diff_UCB.tex
\begin{tikzpicture}

\definecolor{color0}{rgb}{0.12156862745098,0.466666666666667,0.705882352941177}

\begin{axis}[
height={150},
width={230},
legend cell align={left},
legend style={fill opacity=0.8, draw opacity=1, text opacity=1, at={(0.05,0.95)}, anchor=north west, draw=none, font=\small},
tick align=outside,
tick pos=left,
x grid style={white!69.0196078431373!black},
xlabel={number of $H$ queries},
xmin=0, xmax=300,
xtick style={color=black},
tick label style={font=\scriptsize},
label style={font=\small},
y grid style={white!69.0196078431373!black},
ylabel={difference in utility},
ymin=-6.94900920325835, ymax=22.1088930458939,
ytick style={color=black},
axis x line=bottom,
axis y line=left
]
\addplot [semithick, color0]
table {%
1 -0.22
2 -0.55
3 -0.856666666666667
4 -1.11
5 -1.35166666666667
6 -1.56833333333333
7 -1.77666666666667
8 -1.96333333333333
9 -2.15333333333333
10 -2.31166666666667
11 -2.44666666666667
12 -2.59666666666667
13 -2.74666666666667
14 -2.875
15 -3.00333333333333
16 -3.14333333333333
17 -3.26
18 -3.35833333333333
19 -3.45
20 -3.55333333333333
21 -3.65
22 -3.73166666666667
23 -3.82166666666667
24 -3.92666666666667
25 -4.01666666666667
26 -4.065
27 -4.11666666666667
28 -4.16333333333333
29 -4.21333333333333
30 -4.23833333333333
31 -4.26333333333333
32 -4.27166666666666
33 -4.28333333333333
34 -4.32166666666667
35 -4.36166666666667
36 -4.39666666666666
37 -4.41166666666667
38 -4.46666666666667
39 -4.49
40 -4.515
41 -4.54
42 -4.58166666666667
43 -4.59666666666666
44 -4.61166666666666
45 -4.63166666666667
46 -4.64
47 -4.66
48 -4.675
49 -4.70333333333333
50 -4.72
51 -4.73333333333333
52 -4.73333333333333
53 -4.74666666666667
54 -4.74833333333333
55 -4.755
56 -4.755
57 -4.75
58 -4.745
59 -4.70833333333333
60 -4.67333333333334
61 -4.66
62 -4.66166666666667
63 -4.65666666666667
64 -4.66666666666666
65 -4.62333333333333
66 -4.60666666666666
67 -4.56833333333333
68 -4.55333333333333
69 -4.52333333333333
70 -4.49166666666667
71 -4.47499999999999
72 -4.47000000000001
73 -4.455
74 -4.45333333333333
75 -4.42833333333333
76 -4.41166666666666
77 -4.38666666666666
78 -4.35833333333333
79 -4.35333333333334
80 -4.305
81 -4.265
82 -4.24833333333333
83 -4.22166666666666
84 -4.19000000000001
85 -4.14333333333333
86 -4.10833333333333
87 -4.06166666666667
88 -4.01666666666667
89 -3.97833333333332
90 -3.93833333333333
91 -3.89500000000001
92 -3.83833333333334
93 -3.77166666666668
94 -3.735
95 -3.72833333333332
96 -3.67833333333334
97 -3.66666666666667
98 -3.63500000000001
99 -3.595
100 -3.57166666666667
101 -3.56166666666667
102 -3.515
103 -3.52499999999999
104 -3.52
105 -3.48500000000001
106 -3.46166666666667
107 -3.43166666666666
108 -3.41833333333334
109 -3.40833333333333
110 -3.39833333333333
111 -3.34
112 -3.29833333333335
113 -3.26666666666667
114 -3.21499999999999
115 -3.18666666666667
116 -3.14
117 -3.12833333333334
118 -3.10000000000001
119 -3.05333333333333
120 -3.02500000000001
121 -2.99833333333333
122 -2.95833333333334
123 -2.89166666666667
124 -2.84333333333333
125 -2.79499999999999
126 -2.75166666666667
127 -2.71166666666667
128 -2.66166666666666
129 -2.58333333333333
130 -2.5
131 -2.41499999999999
132 -2.35000000000001
133 -2.25166666666667
134 -2.17166666666667
135 -2.095
136 -2.01499999999999
137 -1.94166666666666
138 -1.86
139 -1.81
140 -1.73166666666667
141 -1.64
142 -1.58666666666666
143 -1.50166666666667
144 -1.43166666666666
145 -1.34833333333333
146 -1.255
147 -1.17666666666666
148 -1.08333333333333
149 -0.984999999999999
150 -0.888333333333335
151 -0.803333333333342
152 -0.718333333333334
153 -0.625
154 -0.530000000000001
155 -0.471666666666664
156 -0.366666666666646
157 -0.291666666666686
158 -0.211666666666645
159 -0.144999999999982
160 -0.0600000000000023
161 0.039999999999992
162 0.14166666666668
163 0.241666666666674
164 0.353333333333325
165 0.444999999999993
166 0.531666666666666
167 0.620000000000005
168 0.743333333333311
169 0.841666666666669
170 0.933333333333337
171 1.04000000000002
172 1.14666666666668
173 1.27166666666668
174 1.38166666666666
175 1.495
176 1.625
177 1.75666666666666
178 1.89833333333334
179 1.99666666666667
180 2.11000000000001
181 2.20333333333332
182 2.31166666666667
183 2.41500000000002
184 2.51666666666665
185 2.62333333333333
186 2.69833333333332
187 2.78999999999999
188 2.91166666666669
189 3.03166666666669
190 3.16333333333333
191 3.26166666666666
192 3.36833333333334
193 3.47333333333333
194 3.58500000000001
195 3.69999999999999
196 3.80666666666667
197 3.91833333333335
198 4.01499999999999
199 4.11666666666667
200 4.22166666666666
201 4.35666666666665
202 4.49333333333334
203 4.60833333333335
204 4.71499999999997
205 4.80166666666665
206 4.90666666666667
207 4.99666666666667
208 5.12666666666667
209 5.25166666666667
210 5.39166666666668
211 5.53166666666667
212 5.65333333333334
213 5.76999999999998
214 5.91499999999999
215 6.03333333333333
216 6.13499999999999
217 6.25333333333333
218 6.345
219 6.44333333333336
220 6.56333333333333
221 6.66499999999999
222 6.79666666666665
223 6.93000000000001
224 7.05666666666667
225 7.185
226 7.31833333333336
227 7.44166666666666
228 7.55833333333334
229 7.67166666666668
230 7.78833333333333
231 7.90333333333334
232 8.01333333333335
233 8.125
234 8.24833333333333
235 8.39000000000001
236 8.55333333333334
237 8.72
238 8.84333333333333
239 8.97833333333332
240 9.14166666666665
241 9.31666666666669
242 9.48500000000001
243 9.67666666666668
244 9.80499999999998
245 9.96666666666667
246 10.13
247 10.27
248 10.435
249 10.6016666666667
250 10.7683333333333
251 10.9683333333333
252 11.145
253 11.2933333333333
254 11.4616666666667
255 11.6333333333333
256 11.8066666666667
257 11.965
258 12.1233333333333
259 12.2783333333333
260 12.4033333333333
261 12.5716666666667
262 12.7183333333333
263 12.875
264 13.0516666666666
265 13.1916666666667
266 13.3416666666667
267 13.5166666666667
268 13.6533333333333
269 13.8116666666667
270 13.9416666666667
271 14.11
272 14.265
273 14.44
274 14.565
275 14.7383333333333
276 14.9116666666667
277 15.0716666666667
278 15.2433333333333
279 15.3816666666667
280 15.4966666666667
281 15.6433333333333
282 15.7716666666667
283 15.9166666666667
284 16.0766666666667
285 16.2433333333333
286 16.3983333333333
287 16.5233333333333
288 16.6716666666667
289 16.8116666666667
290 16.9483333333333
291 17.07
292 17.17
293 17.265
294 17.3716666666667
295 17.4566666666667
296 17.535
297 17.5983333333333
298 17.6983333333333
299 17.7583333333334
300 17.795
};
\addlegendentry{mean difference}
\path [fill=color0, fill opacity=0.2]
(axis cs:1,-0.170214158674823)
--(axis cs:1,-0.269785841325177)
--(axis cs:2,-0.625595731048725)
--(axis cs:3,-0.957727160310786)
--(axis cs:4,-1.2352913342248)
--(axis cs:5,-1.50231764103572)
--(axis cs:6,-1.74109741079022)
--(axis cs:7,-1.97191099497912)
--(axis cs:8,-2.17863073436006)
--(axis cs:9,-2.38860717475591)
--(axis cs:10,-2.56412158091201)
--(axis cs:11,-2.71788485356326)
--(axis cs:12,-2.88599816522047)
--(axis cs:13,-3.0534800035104)
--(axis cs:14,-3.19939571234992)
--(axis cs:15,-3.34513289166024)
--(axis cs:16,-3.50159234537382)
--(axis cs:17,-3.63505624771984)
--(axis cs:18,-3.74884486901035)
--(axis cs:19,-3.85378677226002)
--(axis cs:20,-3.97314486874221)
--(axis cs:21,-4.08697435745296)
--(axis cs:22,-4.18552250748173)
--(axis cs:23,-4.29123655976775)
--(axis cs:24,-4.41226803631379)
--(axis cs:25,-4.51596545498558)
--(axis cs:26,-4.57785609472893)
--(axis cs:27,-4.64094003669733)
--(axis cs:28,-4.69992972453293)
--(axis cs:29,-4.76340737752289)
--(axis cs:30,-4.80001256782035)
--(axis cs:31,-4.8370869765941)
--(axis cs:32,-4.85740018599837)
--(axis cs:33,-4.88047229181326)
--(axis cs:34,-4.93189930692822)
--(axis cs:35,-4.98315290419761)
--(axis cs:36,-5.03018515335754)
--(axis cs:37,-5.05699850241144)
--(axis cs:38,-5.1245812650068)
--(axis cs:39,-5.16023390054666)
--(axis cs:40,-5.1973569125158)
--(axis cs:41,-5.23404602855116)
--(axis cs:42,-5.28896878693092)
--(axis cs:43,-5.31414038544648)
--(axis cs:44,-5.34123010302819)
--(axis cs:45,-5.3703744715371)
--(axis cs:46,-5.39171659777426)
--(axis cs:47,-5.42424964856411)
--(axis cs:48,-5.44906530021817)
--(axis cs:49,-5.48884607584784)
--(axis cs:50,-5.51482820418044)
--(axis cs:51,-5.53970554313458)
--(axis cs:52,-5.54883692327919)
--(axis cs:53,-5.57352807709985)
--(axis cs:54,-5.5851385707187)
--(axis cs:55,-5.60470999222509)
--(axis cs:56,-5.61667479744528)
--(axis cs:57,-5.6223539643132)
--(axis cs:58,-5.62819546466052)
--(axis cs:59,-5.59811432012775)
--(axis cs:60,-5.57141005256738)
--(axis cs:61,-5.56748374534032)
--(axis cs:62,-5.57875943867105)
--(axis cs:63,-5.5852546018769)
--(axis cs:64,-5.60597590485475)
--(axis cs:65,-5.57117018414855)
--(axis cs:66,-5.56257771714575)
--(axis cs:67,-5.53185966704179)
--(axis cs:68,-5.5265737778806)
--(axis cs:69,-5.50709929997144)
--(axis cs:70,-5.48328998366696)
--(axis cs:71,-5.47766395107989)
--(axis cs:72,-5.48111520378554)
--(axis cs:73,-5.47411468678013)
--(axis cs:74,-5.48244467168429)
--(axis cs:75,-5.46616862053453)
--(axis cs:76,-5.46064311121336)
--(axis cs:77,-5.44458241201711)
--(axis cs:78,-5.42166076055648)
--(axis cs:79,-5.42348743346072)
--(axis cs:80,-5.38427158540217)
--(axis cs:81,-5.35434557620224)
--(axis cs:82,-5.34632945423512)
--(axis cs:83,-5.32963356137486)
--(axis cs:84,-5.30802777882523)
--(axis cs:85,-5.26998289377931)
--(axis cs:86,-5.24040073817809)
--(axis cs:87,-5.20157879052856)
--(axis cs:88,-5.16278082726555)
--(axis cs:89,-5.13184381459955)
--(axis cs:90,-5.09952168487705)
--(axis cs:91,-5.06646480184026)
--(axis cs:92,-5.01974572935712)
--(axis cs:93,-4.96089445379319)
--(axis cs:94,-4.93288277011413)
--(axis cs:95,-4.93732686013637)
--(axis cs:96,-4.89795084780778)
--(axis cs:97,-4.8923862995322)
--(axis cs:98,-4.86849499804183)
--(axis cs:99,-4.83558075730244)
--(axis cs:100,-4.81949426479253)
--(axis cs:101,-4.81927248758444)
--(axis cs:102,-4.77814108104431)
--(axis cs:103,-4.79571294069243)
--(axis cs:104,-4.79713279595987)
--(axis cs:105,-4.76969881352731)
--(axis cs:106,-4.7538504421445)
--(axis cs:107,-4.73132048232577)
--(axis cs:108,-4.72534355708626)
--(axis cs:109,-4.72475625234659)
--(axis cs:110,-4.72373731829066)
--(axis cs:111,-4.67253788605527)
--(axis cs:112,-4.63967286376435)
--(axis cs:113,-4.61636399087787)
--(axis cs:114,-4.57082948286073)
--(axis cs:115,-4.54921478220033)
--(axis cs:116,-4.50884700022659)
--(axis cs:117,-4.50791395179248)
--(axis cs:118,-4.48743411590592)
--(axis cs:119,-4.44781278720663)
--(axis cs:120,-4.42676516355505)
--(axis cs:121,-4.40708008747936)
--(axis cs:122,-4.37439329509314)
--(axis cs:123,-4.31278771771221)
--(axis cs:124,-4.26997645433164)
--(axis cs:125,-4.22804732514719)
--(axis cs:126,-4.19036520887315)
--(axis cs:127,-4.15930040442675)
--(axis cs:128,-4.11605011902645)
--(axis cs:129,-4.04730556941996)
--(axis cs:130,-3.97221129778227)
--(axis cs:131,-3.89426411275219)
--(axis cs:132,-3.83529813584047)
--(axis cs:133,-3.74174212810581)
--(axis cs:134,-3.66834748453028)
--(axis cs:135,-3.59725001378868)
--(axis cs:136,-3.52238225177419)
--(axis cs:137,-3.45320133881812)
--(axis cs:138,-3.3755241431687)
--(axis cs:139,-3.33097436771135)
--(axis cs:140,-3.2572256113395)
--(axis cs:141,-3.17066618031677)
--(axis cs:142,-3.12135640062172)
--(axis cs:143,-3.04344652617702)
--(axis cs:144,-2.97817671859982)
--(axis cs:145,-2.90198489157744)
--(axis cs:146,-2.81512349026831)
--(axis cs:147,-2.74156121596995)
--(axis cs:148,-2.65537994337973)
--(axis cs:149,-2.5641950654064)
--(axis cs:150,-2.47420355603405)
--(axis cs:151,-2.39781264805387)
--(axis cs:152,-2.3200605660699)
--(axis cs:153,-2.23351762964615)
--(axis cs:154,-2.14605533729806)
--(axis cs:155,-2.09682100576327)
--(axis cs:156,-1.99946259751098)
--(axis cs:157,-1.93205815745506)
--(axis cs:158,-1.86203797234079)
--(axis cs:159,-1.80408230050705)
--(axis cs:160,-1.72792535399309)
--(axis cs:161,-1.63563251480723)
--(axis cs:162,-1.54425192440924)
--(axis cs:163,-1.45106524708142)
--(axis cs:164,-1.34862072342827)
--(axis cs:165,-1.26637721571)
--(axis cs:166,-1.18798446922045)
--(axis cs:167,-1.10629702062374)
--(axis cs:168,-0.987759772420511)
--(axis cs:169,-0.896125904094288)
--(axis cs:170,-0.814617307852039)
--(axis cs:171,-0.71693356969041)
--(axis cs:172,-0.618250784506738)
--(axis cs:173,-0.499101699845756)
--(axis cs:174,-0.397611136593714)
--(axis cs:175,-0.291388059791946)
--(axis cs:176,-0.169697155457158)
--(axis cs:177,-0.0439343074217369)
--(axis cs:178,0.0909020891545871)
--(axis cs:179,0.184636063339939)
--(axis cs:180,0.291216207212856)
--(axis cs:181,0.379027752435003)
--(axis cs:182,0.481125702381692)
--(axis cs:183,0.576155962209911)
--(axis cs:184,0.67002981695938)
--(axis cs:185,0.769999055654114)
--(axis cs:186,0.83743600885329)
--(axis cs:187,0.922134319180193)
--(axis cs:188,1.03549197332261)
--(axis cs:189,1.14826405337678)
--(axis cs:190,1.27419121389805)
--(axis cs:191,1.36706203729147)
--(axis cs:192,1.46633496088199)
--(axis cs:193,1.56399712321206)
--(axis cs:194,1.66747483111328)
--(axis cs:195,1.77380736770783)
--(axis cs:196,1.87021746808959)
--(axis cs:197,1.97348559805622)
--(axis cs:198,2.05994884641591)
--(axis cs:199,2.15075226900309)
--(axis cs:200,2.24705983740992)
--(axis cs:201,2.37386423843191)
--(axis cs:202,2.50108079083929)
--(axis cs:203,2.60657695081407)
--(axis cs:204,2.70394925133531)
--(axis cs:205,2.7801593043404)
--(axis cs:206,2.87681146839832)
--(axis cs:207,2.95848770005701)
--(axis cs:208,3.07906757945401)
--(axis cs:209,3.19633928670298)
--(axis cs:210,3.32732598758187)
--(axis cs:211,3.45779685185388)
--(axis cs:212,3.57165265162291)
--(axis cs:213,3.67985497884052)
--(axis cs:214,3.81627638372287)
--(axis cs:215,3.9250928770539)
--(axis cs:216,4.01745255011775)
--(axis cs:217,4.12633621207552)
--(axis cs:218,4.20805591049883)
--(axis cs:219,4.29646237656333)
--(axis cs:220,4.4078677832759)
--(axis cs:221,4.49810809716613)
--(axis cs:222,4.62151579208435)
--(axis cs:223,4.74619297782115)
--(axis cs:224,4.86598796171021)
--(axis cs:225,4.9851394927195)
--(axis cs:226,5.11230603184376)
--(axis cs:227,5.22594141630314)
--(axis cs:228,5.33577475892028)
--(axis cs:229,5.4405817747621)
--(axis cs:230,5.54736670237813)
--(axis cs:231,5.65429684052413)
--(axis cs:232,5.75486495792871)
--(axis cs:233,5.85663686778445)
--(axis cs:234,5.97109221886611)
--(axis cs:235,6.10242217574249)
--(axis cs:236,6.25690232051822)
--(axis cs:237,6.41233232282678)
--(axis cs:238,6.52462773477576)
--(axis cs:239,6.65331524017412)
--(axis cs:240,6.80707146063246)
--(axis cs:241,6.97002909637916)
--(axis cs:242,7.12804482659769)
--(axis cs:243,7.30898696393892)
--(axis cs:244,7.42744857294872)
--(axis cs:245,7.57815576349175)
--(axis cs:246,7.73047372624138)
--(axis cs:247,7.86078707950318)
--(axis cs:248,8.01462941744041)
--(axis cs:249,8.17132613748689)
--(axis cs:250,8.32879227070714)
--(axis cs:251,8.51869166250428)
--(axis cs:252,8.68470504821099)
--(axis cs:253,8.82265254608599)
--(axis cs:254,8.9796039624671)
--(axis cs:255,9.14052723791852)
--(axis cs:256,9.30384920134252)
--(axis cs:257,9.45098890807839)
--(axis cs:258,9.5979871691798)
--(axis cs:259,9.74024136594872)
--(axis cs:260,9.8546253852817)
--(axis cs:261,10.0124399237983)
--(axis cs:262,10.150226790435)
--(axis cs:263,10.2946029985954)
--(axis cs:264,10.4619375906265)
--(axis cs:265,10.5908108792917)
--(axis cs:266,10.7294060897357)
--(axis cs:267,10.8962538905743)
--(axis cs:268,11.0236809303333)
--(axis cs:269,11.1724333416189)
--(axis cs:270,11.293775714342)
--(axis cs:271,11.4518913678166)
--(axis cs:272,11.5984260469858)
--(axis cs:273,11.7635419201179)
--(axis cs:274,11.8764126724724)
--(axis cs:275,12.0387506728559)
--(axis cs:276,12.2014224267357)
--(axis cs:277,12.3517075929281)
--(axis cs:278,12.513534864007)
--(axis cs:279,12.6406570725774)
--(axis cs:280,12.7444150526752)
--(axis cs:281,12.8794130742271)
--(axis cs:282,12.994634012479)
--(axis cs:283,13.1285108325444)
--(axis cs:284,13.2747714149054)
--(axis cs:285,13.4312719854318)
--(axis cs:286,13.5757597491275)
--(axis cs:287,13.687578281841)
--(axis cs:288,13.8226955116146)
--(axis cs:289,13.9507053842907)
--(axis cs:290,14.0739040088914)
--(axis cs:291,14.182972012621)
--(axis cs:292,14.2700396059505)
--(axis cs:293,14.3523403548341)
--(axis cs:294,14.4463389648588)
--(axis cs:295,14.5205315040263)
--(axis cs:296,14.5876570552288)
--(axis cs:297,14.6399836946919)
--(axis cs:298,14.7277760820253)
--(axis cs:299,14.7760693696341)
--(axis cs:300,14.8019206927039)
--(axis cs:300,20.7880793072961)
--(axis cs:300,20.7880793072961)
--(axis cs:299,20.7405972970326)
--(axis cs:298,20.6688905846413)
--(axis cs:297,20.5566829719748)
--(axis cs:296,20.4823429447712)
--(axis cs:295,20.3928018293071)
--(axis cs:294,20.2969943684746)
--(axis cs:293,20.1776596451659)
--(axis cs:292,20.0699603940495)
--(axis cs:291,19.957027987379)
--(axis cs:290,19.8227626577752)
--(axis cs:289,19.6726279490426)
--(axis cs:288,19.5206378217187)
--(axis cs:287,19.3590883848257)
--(axis cs:286,19.2209069175392)
--(axis cs:285,19.0553946812349)
--(axis cs:284,18.8785619184279)
--(axis cs:283,18.7048225007889)
--(axis cs:282,18.5486993208543)
--(axis cs:281,18.4072535924396)
--(axis cs:280,18.2489182806581)
--(axis cs:279,18.1226762607559)
--(axis cs:278,17.9731318026596)
--(axis cs:277,17.7916257404052)
--(axis cs:276,17.6219109065976)
--(axis cs:275,17.4379159938107)
--(axis cs:274,17.2535873275276)
--(axis cs:273,17.1164580798821)
--(axis cs:272,16.9315739530142)
--(axis cs:271,16.7681086321834)
--(axis cs:270,16.5895576189913)
--(axis cs:269,16.4508999917144)
--(axis cs:268,16.2829857363333)
--(axis cs:267,16.137079442759)
--(axis cs:266,15.9539272435976)
--(axis cs:265,15.7925224540417)
--(axis cs:264,15.6413957427068)
--(axis cs:263,15.4553970014046)
--(axis cs:262,15.2864398762317)
--(axis cs:261,15.130893409535)
--(axis cs:260,14.952041281385)
--(axis cs:259,14.8164253007179)
--(axis cs:258,14.6486794974869)
--(axis cs:257,14.4790110919216)
--(axis cs:256,14.3094841319908)
--(axis cs:255,14.1261394287481)
--(axis cs:254,13.9437293708662)
--(axis cs:253,13.7640141205807)
--(axis cs:252,13.605294951789)
--(axis cs:251,13.4179750041624)
--(axis cs:250,13.2078743959595)
--(axis cs:249,13.0320071958464)
--(axis cs:248,12.8553705825596)
--(axis cs:247,12.6792129204968)
--(axis cs:246,12.5295262737586)
--(axis cs:245,12.3551775698416)
--(axis cs:244,12.1825514270513)
--(axis cs:243,12.0443463693944)
--(axis cs:242,11.8419551734023)
--(axis cs:241,11.6633042369542)
--(axis cs:240,11.4762618727009)
--(axis cs:239,11.3033514264925)
--(axis cs:238,11.1620389318909)
--(axis cs:237,11.0276676771732)
--(axis cs:236,10.8497643461484)
--(axis cs:235,10.6775778242575)
--(axis cs:234,10.5255744478006)
--(axis cs:233,10.3933631322155)
--(axis cs:232,10.271801708738)
--(axis cs:231,10.1523698261425)
--(axis cs:230,10.0292999642885)
--(axis cs:229,9.90275155857123)
--(axis cs:228,9.78089190774639)
--(axis cs:227,9.65739191703019)
--(axis cs:226,9.52436063482291)
--(axis cs:225,9.3848605072805)
--(axis cs:224,9.24734537162313)
--(axis cs:223,9.11380702217885)
--(axis cs:222,8.97181754124898)
--(axis cs:221,8.83189190283387)
--(axis cs:220,8.71879888339076)
--(axis cs:219,8.59020429010333)
--(axis cs:218,8.48194408950117)
--(axis cs:217,8.38033045459114)
--(axis cs:216,8.25254744988225)
--(axis cs:215,8.14157378961276)
--(axis cs:214,8.01372361627713)
--(axis cs:213,7.86014502115948)
--(axis cs:212,7.73501401504375)
--(axis cs:211,7.60553648147946)
--(axis cs:210,7.45600734575147)
--(axis cs:209,7.30699404663035)
--(axis cs:208,7.17426575387933)
--(axis cs:207,7.03484563327632)
--(axis cs:206,6.93652186493502)
--(axis cs:205,6.82317402899293)
--(axis cs:204,6.72605074866469)
--(axis cs:203,6.6100897158526)
--(axis cs:202,6.48558587582738)
--(axis cs:201,6.33946909490142)
--(axis cs:200,6.19627349592341)
--(axis cs:199,6.08258106433024)
--(axis cs:198,5.97005115358409)
--(axis cs:197,5.86318106861045)
--(axis cs:196,5.74311586524374)
--(axis cs:195,5.62619263229217)
--(axis cs:194,5.50252516888672)
--(axis cs:193,5.3826695434546)
--(axis cs:192,5.27033170578468)
--(axis cs:191,5.15627129604186)
--(axis cs:190,5.05247545276861)
--(axis cs:189,4.91506927995655)
--(axis cs:188,4.78784136001072)
--(axis cs:187,4.65786568081981)
--(axis cs:186,4.55923065781338)
--(axis cs:185,4.47666761101255)
--(axis cs:184,4.36330351637395)
--(axis cs:183,4.25384403779009)
--(axis cs:182,4.14220763095164)
--(axis cs:181,4.02763891423166)
--(axis cs:180,3.92878379278714)
--(axis cs:179,3.80869726999339)
--(axis cs:178,3.70576457751208)
--(axis cs:177,3.55726764075507)
--(axis cs:176,3.41969715545716)
--(axis cs:175,3.28138805979195)
--(axis cs:174,3.16094446992705)
--(axis cs:173,3.04243503317909)
--(axis cs:172,2.91158411784007)
--(axis cs:171,2.79693356969041)
--(axis cs:170,2.68128397451871)
--(axis cs:169,2.57945923742762)
--(axis cs:168,2.47442643908718)
--(axis cs:167,2.34629702062374)
--(axis cs:166,2.25131780255379)
--(axis cs:165,2.15637721571)
--(axis cs:164,2.05528739009494)
--(axis cs:163,1.93439858041475)
--(axis cs:162,1.82758525774257)
--(axis cs:161,1.71563251480723)
--(axis cs:160,1.60792535399309)
--(axis cs:159,1.51408230050705)
--(axis cs:158,1.43870463900746)
--(axis cs:157,1.34872482412173)
--(axis cs:156,1.26612926417765)
--(axis cs:155,1.15348767242993)
--(axis cs:154,1.08605533729806)
--(axis cs:153,0.983517629646149)
--(axis cs:152,0.883393899403235)
--(axis cs:151,0.791145981387204)
--(axis cs:150,0.697536889367384)
--(axis cs:149,0.594195065406402)
--(axis cs:148,0.488713276713061)
--(axis cs:147,0.38822788263662)
--(axis cs:146,0.305123490268314)
--(axis cs:145,0.20531822491077)
--(axis cs:144,0.114843385266485)
--(axis cs:143,0.0401131928436862)
--(axis cs:142,-0.0519769327116144)
--(axis cs:141,-0.109333819683228)
--(axis cs:140,-0.206107721993833)
--(axis cs:139,-0.289025632288646)
--(axis cs:138,-0.344475856831296)
--(axis cs:137,-0.430131994515218)
--(axis cs:136,-0.507617748225806)
--(axis cs:135,-0.592749986211322)
--(axis cs:134,-0.674985848803051)
--(axis cs:133,-0.76159120522752)
--(axis cs:132,-0.864701864159533)
--(axis cs:131,-0.935735887247813)
--(axis cs:130,-1.02778870221773)
--(axis cs:129,-1.11936109724671)
--(axis cs:128,-1.20728321430688)
--(axis cs:127,-1.26403292890658)
--(axis cs:126,-1.31296812446018)
--(axis cs:125,-1.36195267485281)
--(axis cs:124,-1.41669021233503)
--(axis cs:123,-1.47054561562112)
--(axis cs:122,-1.54227337157352)
--(axis cs:121,-1.58958657918731)
--(axis cs:120,-1.62323483644495)
--(axis cs:119,-1.65885387946003)
--(axis cs:118,-1.71256588409408)
--(axis cs:117,-1.74875271487419)
--(axis cs:116,-1.77115299977341)
--(axis cs:115,-1.824118551133)
--(axis cs:114,-1.85917051713927)
--(axis cs:113,-1.91696934245546)
--(axis cs:112,-1.95699380290232)
--(axis cs:111,-2.00746211394473)
--(axis cs:110,-2.07292934837601)
--(axis cs:109,-2.09191041432007)
--(axis cs:108,-2.1113231095804)
--(axis cs:107,-2.13201285100757)
--(axis cs:106,-2.16948289118883)
--(axis cs:105,-2.20030118647269)
--(axis cs:104,-2.24286720404013)
--(axis cs:103,-2.25428705930757)
--(axis cs:102,-2.25185891895569)
--(axis cs:101,-2.30406084574889)
--(axis cs:100,-2.3238390685408)
--(axis cs:99,-2.35441924269757)
--(axis cs:98,-2.40150500195817)
--(axis cs:97,-2.44094703380114)
--(axis cs:96,-2.45871581885889)
--(axis cs:95,-2.51933980653029)
--(axis cs:94,-2.53711722988587)
--(axis cs:93,-2.58243887954014)
--(axis cs:92,-2.65692093730955)
--(axis cs:91,-2.72353519815974)
--(axis cs:90,-2.77714498178961)
--(axis cs:89,-2.82482285206712)
--(axis cs:88,-2.87055250606778)
--(axis cs:87,-2.92175454280477)
--(axis cs:86,-2.97626592848858)
--(axis cs:85,-3.01668377288736)
--(axis cs:84,-3.07197222117477)
--(axis cs:83,-3.11369977195847)
--(axis cs:82,-3.15033721243155)
--(axis cs:81,-3.17565442379776)
--(axis cs:80,-3.22572841459783)
--(axis cs:79,-3.28317923320595)
--(axis cs:78,-3.29500590611019)
--(axis cs:77,-3.32875092131623)
--(axis cs:76,-3.36269022211997)
--(axis cs:75,-3.39049804613213)
--(axis cs:74,-3.42422199498237)
--(axis cs:73,-3.43588531321987)
--(axis cs:72,-3.45888479621446)
--(axis cs:71,-3.47233604892011)
--(axis cs:70,-3.50004334966637)
--(axis cs:69,-3.53956736669523)
--(axis cs:68,-3.58009288878606)
--(axis cs:67,-3.60480699962488)
--(axis cs:66,-3.65075561618758)
--(axis cs:65,-3.67549648251811)
--(axis cs:64,-3.72735742847858)
--(axis cs:63,-3.72807873145643)
--(axis cs:62,-3.74457389466229)
--(axis cs:61,-3.75251625465968)
--(axis cs:60,-3.77525661409928)
--(axis cs:59,-3.81855234653891)
--(axis cs:58,-3.86180453533948)
--(axis cs:57,-3.8776460356868)
--(axis cs:56,-3.89332520255471)
--(axis cs:55,-3.90529000777491)
--(axis cs:54,-3.91152809594797)
--(axis cs:53,-3.91980525623348)
--(axis cs:52,-3.91782974338747)
--(axis cs:51,-3.92696112353209)
--(axis cs:50,-3.92517179581956)
--(axis cs:49,-3.91782059081883)
--(axis cs:48,-3.90093469978183)
--(axis cs:47,-3.89575035143589)
--(axis cs:46,-3.88828340222574)
--(axis cs:45,-3.89295886179624)
--(axis cs:44,-3.88210323030514)
--(axis cs:43,-3.87919294788686)
--(axis cs:42,-3.87436454640241)
--(axis cs:41,-3.84595397144884)
--(axis cs:40,-3.8326430874842)
--(axis cs:39,-3.81976609945334)
--(axis cs:38,-3.80875206832653)
--(axis cs:37,-3.7663348309219)
--(axis cs:36,-3.7631481799758)
--(axis cs:35,-3.74018042913572)
--(axis cs:34,-3.71143402640511)
--(axis cs:33,-3.68619437485341)
--(axis cs:32,-3.68593314733497)
--(axis cs:31,-3.68957969007257)
--(axis cs:30,-3.67665409884632)
--(axis cs:29,-3.66325928914378)
--(axis cs:28,-3.62673694213374)
--(axis cs:27,-3.592393296636)
--(axis cs:26,-3.55214390527107)
--(axis cs:25,-3.51736787834775)
--(axis cs:24,-3.44106529701955)
--(axis cs:23,-3.35209677356558)
--(axis cs:22,-3.2778108258516)
--(axis cs:21,-3.21302564254704)
--(axis cs:20,-3.13352179792446)
--(axis cs:19,-3.04621322773998)
--(axis cs:18,-2.96782179765632)
--(axis cs:17,-2.88494375228016)
--(axis cs:16,-2.78507432129285)
--(axis cs:15,-2.66153377500643)
--(axis cs:14,-2.55060428765008)
--(axis cs:13,-2.43985332982294)
--(axis cs:12,-2.30733516811286)
--(axis cs:11,-2.17544847977008)
--(axis cs:10,-2.05921175242132)
--(axis cs:9,-1.91805949191076)
--(axis cs:8,-1.74803593230661)
--(axis cs:7,-1.58142233835421)
--(axis cs:6,-1.39556925587645)
--(axis cs:5,-1.20101569229762)
--(axis cs:4,-0.9847086657752)
--(axis cs:3,-0.755606173022548)
--(axis cs:2,-0.474404268951275)
--(axis cs:1,-0.170214158674823)
--cycle;
\addlegendimage{area legend, fill=color0, fill opacity=0.2}
\addlegendentry{95\% \acro{CI}}

\addplot [semithick, white!50.1960784313725!black, dashed, forget plot]
table {%
1 0
2 0
3 0
4 0
5 0
6 0
7 0
8 0
9 0
10 0
11 0
12 0
13 0
14 0
15 0
16 0
17 0
18 0
19 0
20 0
21 0
22 0
23 0
24 0
25 0
26 0
27 0
28 0
29 0
30 0
31 0
32 0
33 0
34 0
35 0
36 0
37 0
38 0
39 0
40 0
41 0
42 0
43 0
44 0
45 0
46 0
47 0
48 0
49 0
50 0
51 0
52 0
53 0
54 0
55 0
56 0
57 0
58 0
59 0
60 0
61 0
62 0
63 0
64 0
65 0
66 0
67 0
68 0
69 0
70 0
71 0
72 0
73 0
74 0
75 0
76 0
77 0
78 0
79 0
80 0
81 0
82 0
83 0
84 0
85 0
86 0
87 0
88 0
89 0
90 0
91 0
92 0
93 0
94 0
95 0
96 0
97 0
98 0
99 0
100 0
101 0
102 0
103 0
104 0
105 0
106 0
107 0
108 0
109 0
110 0
111 0
112 0
113 0
114 0
115 0
116 0
117 0
118 0
119 0
120 0
121 0
122 0
123 0
124 0
125 0
126 0
127 0
128 0
129 0
130 0
131 0
132 0
133 0
134 0
135 0
136 0
137 0
138 0
139 0
140 0
141 0
142 0
143 0
144 0
145 0
146 0
147 0
148 0
149 0
150 0
151 0
152 0
153 0
154 0
155 0
156 0
157 0
158 0
159 0
160 0
161 0
162 0
163 0
164 0
165 0
166 0
167 0
168 0
169 0
170 0
171 0
172 0
173 0
174 0
175 0
176 0
177 0
178 0
179 0
180 0
181 0
182 0
183 0
184 0
185 0
186 0
187 0
188 0
189 0
190 0
191 0
192 0
193 0
194 0
195 0
196 0
197 0
198 0
199 0
200 0
201 0
202 0
203 0
204 0
205 0
206 0
207 0
208 0
209 0
210 0
211 0
212 0
213 0
214 0
215 0
216 0
217 0
218 0
219 0
220 0
221 0
222 0
223 0
224 0
225 0
226 0
227 0
228 0
229 0
230 0
231 0
232 0
233 0
234 0
235 0
236 0
237 0
238 0
239 0
240 0
241 0
242 0
243 0
244 0
245 0
246 0
247 0
248 0
249 0
250 0
251 0
252 0
253 0
254 0
255 0
256 0
257 0
258 0
259 0
260 0
261 0
262 0
263 0
264 0
265 0
266 0
267 0
268 0
269 0
270 0
271 0
272 0
273 0
274 0
275 0
276 0
277 0
278 0
279 0
280 0
281 0
282 0
283 0
284 0
285 0
286 0
287 0
288 0
289 0
290 0
291 0
292 0
293 0
294 0
295 0
296 0
297 0
298 0
299 0
300 0
};
\end{axis}

\end{tikzpicture}

%% file: main_06_conclusion.tex
\section{Conclusion}

We have proposed a multifidelity active search model in which an exact oracle and a cheaper surrogate are queried in parallel.
We presented a novel nonmyopic policy (based on two-stage rollout) for this setting that reasons about the remaining queries on both fidelities seeking to maximize the total number of discoveries.
Our policy is aware of the remaining budget and dynamically balances exploration and exploitation.
Experiments on real-world data demonstrate that the policy significantly outperforms myopic benchmarks.

%% file: media/utility_diff_UG.tex
\begin{tikzpicture}

\definecolor{color0}{rgb}{0.12156862745098,0.466666666666667,0.705882352941177}

\begin{axis}[
height={150},
width={230},
legend cell align={left},
legend style={fill opacity=0.8, draw opacity=1, text opacity=1, at={(0.05,0.95)}, anchor=north west, draw=none, font=\small},
tick align=outside,
tick pos=left,
x grid style={white!69.0196078431373!black},
xlabel={number of $H$ queries},
xmin=-13.95, xmax=314.95,
xtick style={color=black},
tick label style={font=\scriptsize},
label style={font=\small},
y grid style={white!69.0196078431373!black},
ylabel={difference in utility},
ymin=-6.64789702353792, ymax=20.2273885056756,
ytick style={color=black},
axis x line=bottom,
axis y line=left
]
\addplot [semithick, color0]
table {%
1 -0.22
2 -0.535
3 -0.828333333333333
4 -1.05666666666667
5 -1.285
6 -1.47666666666667
7 -1.68
8 -1.865
9 -2.045
10 -2.19833333333333
11 -2.31666666666667
12 -2.45666666666667
13 -2.59666666666667
14 -2.71833333333333
15 -2.85333333333333
16 -2.995
17 -3.10833333333333
18 -3.21166666666667
19 -3.28666666666667
20 -3.37833333333333
21 -3.45666666666667
22 -3.53
23 -3.61333333333333
24 -3.72333333333333
25 -3.815
26 -3.88333333333333
27 -3.94166666666667
28 -3.99666666666667
29 -4.05833333333333
30 -4.09166666666667
31 -4.12
32 -4.125
33 -4.13333333333333
34 -4.18166666666666
35 -4.20333333333333
36 -4.22166666666666
37 -4.24166666666666
38 -4.28666666666667
39 -4.31833333333333
40 -4.34166666666667
41 -4.37
42 -4.42333333333334
43 -4.44333333333333
44 -4.47833333333333
45 -4.465
46 -4.485
47 -4.50166666666667
48 -4.53333333333334
49 -4.55166666666667
50 -4.555
51 -4.56
52 -4.54666666666667
53 -4.55166666666667
54 -4.55
55 -4.54666666666667
56 -4.535
57 -4.53666666666666
58 -4.52666666666667
59 -4.505
60 -4.47166666666667
61 -4.44833333333333
62 -4.44
63 -4.425
64 -4.43166666666666
65 -4.39666666666667
66 -4.37833333333333
67 -4.36666666666667
68 -4.34666666666667
69 -4.33333333333333
70 -4.30333333333333
71 -4.27333333333333
72 -4.24
73 -4.21166666666667
74 -4.18333333333334
75 -4.14166666666667
76 -4.10666666666667
77 -4.07333333333333
78 -4.00666666666667
79 -3.98333333333333
80 -3.93833333333334
81 -3.90166666666667
82 -3.88
83 -3.83333333333333
84 -3.78666666666668
85 -3.75999999999999
86 -3.72666666666667
87 -3.68166666666667
88 -3.63666666666666
89 -3.60833333333333
90 -3.55833333333332
91 -3.53166666666667
92 -3.48166666666667
93 -3.41500000000001
94 -3.38166666666667
95 -3.36333333333333
96 -3.33833333333334
97 -3.31166666666667
98 -3.26833333333333
99 -3.23
100 -3.21833333333333
101 -3.2
102 -3.15666666666667
103 -3.15166666666666
104 -3.13499999999999
105 -3.11333333333334
106 -3.09833333333334
107 -3.06999999999999
108 -3.07166666666667
109 -3.06666666666666
110 -3.065
111 -3.01333333333334
112 -2.98
113 -2.96000000000001
114 -2.93333333333332
115 -2.91000000000001
116 -2.86333333333333
117 -2.83000000000001
118 -2.80166666666668
119 -2.73666666666666
120 -2.71333333333334
121 -2.68000000000001
122 -2.63500000000001
123 -2.58499999999999
124 -2.54666666666667
125 -2.51499999999999
126 -2.44666666666666
127 -2.38666666666667
128 -2.355
129 -2.26833333333333
130 -2.18333333333332
131 -2.10666666666667
132 -2.06
133 -1.98166666666667
134 -1.92333333333335
135 -1.86333333333334
136 -1.80166666666666
137 -1.75666666666666
138 -1.7
139 -1.65000000000001
140 -1.59333333333333
141 -1.53666666666666
142 -1.48999999999999
143 -1.42833333333334
144 -1.35166666666666
145 -1.285
146 -1.205
147 -1.14
148 -1.06333333333333
149 -0.971666666666664
150 -0.875
151 -0.796666666666667
152 -0.728333333333339
153 -0.63333333333334
154 -0.564999999999998
155 -0.498333333333335
156 -0.399999999999977
157 -0.350000000000023
158 -0.286666666666662
159 -0.218333333333334
160 -0.153333333333336
161 -0.056666666666672
162 0.028333333333336
163 0.0983333333333292
164 0.175000000000011
165 0.238333333333316
166 0.33833333333331
167 0.439999999999998
168 0.536666666666662
169 0.616666666666674
170 0.688333333333333
171 0.761666666666684
172 0.823333333333323
173 0.908333333333331
174 0.981666666666655
175 1.06833333333333
176 1.15666666666667
177 1.25999999999999
178 1.37
179 1.45500000000001
180 1.54166666666669
181 1.58333333333331
182 1.63500000000002
183 1.69
184 1.76499999999999
185 1.84999999999999
186 1.91
187 1.97333333333333
188 2.08666666666667
189 2.18833333333336
190 2.30999999999997
191 2.42833333333334
192 2.535
193 2.60666666666665
194 2.71000000000001
195 2.83500000000001
196 2.94499999999999
197 3.02500000000001
198 3.13666666666666
199 3.21333333333334
200 3.28833333333333
201 3.40666666666667
202 3.50833333333335
203 3.62166666666667
204 3.70999999999998
205 3.75999999999999
206 3.875
207 3.95833333333334
208 4.065
209 4.17833333333331
210 4.32166666666669
211 4.43833333333333
212 4.55833333333334
213 4.70166666666665
214 4.83166666666665
215 4.94499999999999
216 5.04666666666665
217 5.17333333333332
218 5.26166666666666
219 5.36500000000001
220 5.46333333333334
221 5.57499999999999
222 5.70666666666665
223 5.83833333333334
224 5.95833333333334
225 6.07666666666668
226 6.22333333333336
227 6.34999999999999
228 6.46333333333334
229 6.58833333333334
230 6.70500000000001
231 6.81833333333333
232 6.94
233 7.04833333333332
234 7.185
235 7.32833333333335
236 7.44499999999999
237 7.60166666666666
238 7.71666666666667
239 7.80499999999998
240 7.92999999999998
241 8.07166666666669
242 8.19500000000002
243 8.35166666666666
244 8.45833333333331
245 8.57666666666668
246 8.715
247 8.81833333333336
248 8.93833333333333
249 9.07333333333332
250 9.215
251 9.36500000000001
252 9.51166666666666
253 9.65166666666667
254 9.79499999999999
255 9.93333333333334
256 10.09
257 10.23
258 10.36
259 10.4716666666667
260 10.5966666666667
261 10.7516666666667
262 10.9016666666667
263 11.0433333333333
264 11.2333333333333
265 11.3833333333333
266 11.5616666666667
267 11.7333333333333
268 11.8833333333333
269 12.0516666666666
270 12.2016666666667
271 12.3566666666667
272 12.5033333333333
273 12.6683333333334
274 12.8116666666667
275 12.97
276 13.155
277 13.2916666666667
278 13.4766666666667
279 13.6116666666667
280 13.735
281 13.8816666666667
282 14.0366666666667
283 14.195
284 14.3483333333334
285 14.4933333333333
286 14.6316666666667
287 14.745
288 14.89
289 14.995
290 15.1133333333333
291 15.2316666666667
292 15.3133333333333
293 15.4033333333333
294 15.515
295 15.62
296 15.6866666666667
297 15.7516666666667
298 15.83
299 15.9133333333334
300 15.9666666666667
};
\addlegendentry{mean difference}
\addplot [semithick, white!50.1960784313725!black, dashed, forget plot]
table {%
1 0
2 0
3 0
4 0
5 0
6 0
7 0
8 0
9 0
10 0
11 0
12 0
13 0
14 0
15 0
16 0
17 0
18 0
19 0
20 0
21 0
22 0
23 0
24 0
25 0
26 0
27 0
28 0
29 0
30 0
31 0
32 0
33 0
34 0
35 0
36 0
37 0
38 0
39 0
40 0
41 0
42 0
43 0
44 0
45 0
46 0
47 0
48 0
49 0
50 0
51 0
52 0
53 0
54 0
55 0
56 0
57 0
58 0
59 0
60 0
61 0
62 0
63 0
64 0
65 0
66 0
67 0
68 0
69 0
70 0
71 0
72 0
73 0
74 0
75 0
76 0
77 0
78 0
79 0
80 0
81 0
82 0
83 0
84 0
85 0
86 0
87 0
88 0
89 0
90 0
91 0
92 0
93 0
94 0
95 0
96 0
97 0
98 0
99 0
100 0
101 0
102 0
103 0
104 0
105 0
106 0
107 0
108 0
109 0
110 0
111 0
112 0
113 0
114 0
115 0
116 0
117 0
118 0
119 0
120 0
121 0
122 0
123 0
124 0
125 0
126 0
127 0
128 0
129 0
130 0
131 0
132 0
133 0
134 0
135 0
136 0
137 0
138 0
139 0
140 0
141 0
142 0
143 0
144 0
145 0
146 0
147 0
148 0
149 0
150 0
151 0
152 0
153 0
154 0
155 0
156 0
157 0
158 0
159 0
160 0
161 0
162 0
163 0
164 0
165 0
166 0
167 0
168 0
169 0
170 0
171 0
172 0
173 0
174 0
175 0
176 0
177 0
178 0
179 0
180 0
181 0
182 0
183 0
184 0
185 0
186 0
187 0
188 0
189 0
190 0
191 0
192 0
193 0
194 0
195 0
196 0
197 0
198 0
199 0
200 0
201 0
202 0
203 0
204 0
205 0
206 0
207 0
208 0
209 0
210 0
211 0
212 0
213 0
214 0
215 0
216 0
217 0
218 0
219 0
220 0
221 0
222 0
223 0
224 0
225 0
226 0
227 0
228 0
229 0
230 0
231 0
232 0
233 0
234 0
235 0
236 0
237 0
238 0
239 0
240 0
241 0
242 0
243 0
244 0
245 0
246 0
247 0
248 0
249 0
250 0
251 0
252 0
253 0
254 0
255 0
256 0
257 0
258 0
259 0
260 0
261 0
262 0
263 0
264 0
265 0
266 0
267 0
268 0
269 0
270 0
271 0
272 0
273 0
274 0
275 0
276 0
277 0
278 0
279 0
280 0
281 0
282 0
283 0
284 0
285 0
286 0
287 0
288 0
289 0
290 0
291 0
292 0
293 0
294 0
295 0
296 0
297 0
298 0
299 0
300 0
};

\path [fill=color0, fill opacity=0.2]
(axis cs:1,-0.170214158674823)
--(axis cs:1,-0.269785841325177)
--(axis cs:2,-0.610650012719115)
--(axis cs:3,-0.929798935546162)
--(axis cs:4,-1.18413752801155)
--(axis cs:5,-1.4370502112895)
--(axis cs:6,-1.65185116666825)
--(axis cs:7,-1.87890610747314)
--(axis cs:8,-2.0847590953493)
--(axis cs:9,-2.28496842369215)
--(axis cs:10,-2.45660213571807)
--(axis cs:11,-2.59382520104084)
--(axis cs:12,-2.75271687776389)
--(axis cs:13,-2.90979679738609)
--(axis cs:14,-3.04940544606562)
--(axis cs:15,-3.20189662690477)
--(axis cs:16,-3.35986784103419)
--(axis cs:17,-3.49048340960067)
--(axis cs:18,-3.60853191151558)
--(axis cs:19,-3.69743876550629)
--(axis cs:20,-3.80555456911345)
--(axis cs:21,-3.90191744913216)
--(axis cs:22,-3.99126351227005)
--(axis cs:23,-4.08962438833686)
--(axis cs:24,-4.21566506997826)
--(axis cs:25,-4.32153949393068)
--(axis cs:26,-4.40368225150934)
--(axis cs:27,-4.47357223497766)
--(axis cs:28,-4.54047455203171)
--(axis cs:29,-4.61502303631373)
--(axis cs:30,-4.66098480064091)
--(axis cs:31,-4.70101158918776)
--(axis cs:32,-4.71797097853866)
--(axis cs:33,-4.73844807546001)
--(axis cs:34,-4.79907010363418)
--(axis cs:35,-4.83275260895801)
--(axis cs:36,-4.8645342226501)
--(axis cs:37,-4.89647978501477)
--(axis cs:38,-4.95443620306941)
--(axis cs:39,-4.99888179298314)
--(axis cs:40,-5.03433451097045)
--(axis cs:41,-5.07500622943731)
--(axis cs:42,-5.14233173026108)
--(axis cs:43,-5.17352575442094)
--(axis cs:44,-5.22275839797612)
--(axis cs:45,-5.22110823844618)
--(axis cs:46,-5.25433867769245)
--(axis cs:47,-5.28440658089276)
--(axis cs:48,-5.32532354806379)
--(axis cs:49,-5.35332084053577)
--(axis cs:50,-5.36657812786249)
--(axis cs:51,-5.38331277033766)
--(axis cs:52,-5.3797166812773)
--(axis cs:53,-5.39636934027122)
--(axis cs:54,-5.40399035684845)
--(axis cs:55,-5.41255148947203)
--(axis cs:56,-5.41297496042951)
--(axis cs:57,-5.42395493378363)
--(axis cs:58,-5.4262931358464)
--(axis cs:59,-5.41108332957148)
--(axis cs:60,-5.38685705389871)
--(axis cs:61,-5.37381368806005)
--(axis cs:62,-5.37658898021618)
--(axis cs:63,-5.37260896806254)
--(axis cs:64,-5.39041403956108)
--(axis cs:65,-5.36420769243105)
--(axis cs:66,-5.35555092342736)
--(axis cs:67,-5.35261015977853)
--(axis cs:68,-5.34302939020089)
--(axis cs:69,-5.33952163338346)
--(axis cs:70,-5.31853370536504)
--(axis cs:71,-5.29894776036884)
--(axis cs:72,-5.27353773003757)
--(axis cs:73,-5.25380095153985)
--(axis cs:74,-5.23351894868103)
--(axis cs:75,-5.19997119853739)
--(axis cs:76,-5.17469304679148)
--(axis cs:77,-5.1497552367485)
--(axis cs:78,-5.08922849194167)
--(axis cs:79,-5.07329430824877)
--(axis cs:80,-5.03801281741665)
--(axis cs:81,-5.01104528021692)
--(axis cs:82,-4.9988195053198)
--(axis cs:83,-4.96351919672383)
--(axis cs:84,-4.92836605266572)
--(axis cs:85,-4.91245273721727)
--(axis cs:86,-4.88534980003384)
--(axis cs:87,-4.8497269133152)
--(axis cs:88,-4.811481496507)
--(axis cs:89,-4.79033440253362)
--(axis cs:90,-4.74917644247771)
--(axis cs:91,-4.73388016411339)
--(axis cs:92,-4.69359074215432)
--(axis cs:93,-4.63518382737931)
--(axis cs:94,-4.61127750601174)
--(axis cs:95,-4.60334391631048)
--(axis cs:96,-4.58617555211054)
--(axis cs:97,-4.56462591214026)
--(axis cs:98,-4.52831810954881)
--(axis cs:99,-4.49659597440317)
--(axis cs:100,-4.49187564594728)
--(axis cs:101,-4.48227454644339)
--(axis cs:102,-4.44319949153134)
--(axis cs:103,-4.4438756897379)
--(axis cs:104,-4.43419489429794)
--(axis cs:105,-4.4213215110979)
--(axis cs:106,-4.41624011067993)
--(axis cs:107,-4.39667162597343)
--(axis cs:108,-4.40548157282585)
--(axis cs:109,-4.40990000186789)
--(axis cs:110,-4.41839411854133)
--(axis cs:111,-4.37555375475291)
--(axis cs:112,-4.35150033127046)
--(axis cs:113,-4.34040464941577)
--(axis cs:114,-4.32033164020906)
--(axis cs:115,-4.3051213287961)
--(axis cs:116,-4.2658339205284)
--(axis cs:117,-4.24211621789382)
--(axis cs:118,-4.22180354342264)
--(axis cs:119,-4.16444943571253)
--(axis cs:120,-4.14954086451889)
--(axis cs:121,-4.1252648619343)
--(axis cs:122,-4.09080227819984)
--(axis cs:123,-4.05092759502072)
--(axis cs:124,-4.02067992443625)
--(axis cs:125,-3.99737390562656)
--(axis cs:126,-3.93913600720966)
--(axis cs:127,-3.8913247601103)
--(axis cs:128,-3.86928138150678)
--(axis cs:129,-3.79163246371876)
--(axis cs:130,-3.71567986178822)
--(axis cs:131,-3.64658402453636)
--(axis cs:132,-3.60666493746121)
--(axis cs:133,-3.53449117624709)
--(axis cs:134,-3.48561009783967)
--(axis cs:135,-3.43346375542229)
--(axis cs:136,-3.37995944810802)
--(axis cs:137,-3.34090162169985)
--(axis cs:138,-3.29160680692383)
--(axis cs:139,-3.24843104400486)
--(axis cs:140,-3.19910696907399)
--(axis cs:141,-3.150381871753)
--(axis cs:142,-3.11132774577583)
--(axis cs:143,-3.05895746892455)
--(axis cs:144,-2.98841937595588)
--(axis cs:145,-2.93058536359412)
--(axis cs:146,-2.85793029909271)
--(axis cs:147,-2.79948930032224)
--(axis cs:148,-2.73121928776831)
--(axis cs:149,-2.64560582477131)
--(axis cs:150,-2.55676523541923)
--(axis cs:151,-2.48504133253361)
--(axis cs:152,-2.4234054850337)
--(axis cs:153,-2.3352818411492)
--(axis cs:154,-2.27549680712508)
--(axis cs:155,-2.21783031656916)
--(axis cs:156,-2.1255963814938)
--(axis cs:157,-2.08258042207646)
--(axis cs:158,-2.02827998411171)
--(axis cs:159,-1.96831829431273)
--(axis cs:160,-1.91425979313615)
--(axis cs:161,-1.82552903518791)
--(axis cs:162,-1.74926967669493)
--(axis cs:163,-1.68524461621239)
--(axis cs:164,-1.61526666435469)
--(axis cs:165,-1.55916701740859)
--(axis cs:166,-1.46544114409281)
--(axis cs:167,-1.36981375611451)
--(axis cs:168,-1.27643915438534)
--(axis cs:169,-1.20166595203187)
--(axis cs:170,-1.13812853017966)
--(axis cs:171,-1.07039268578463)
--(axis cs:172,-1.01774929903029)
--(axis cs:173,-0.938513084205989)
--(axis cs:174,-0.874444785782748)
--(axis cs:175,-0.795421669164018)
--(axis cs:176,-0.715089658187301)
--(axis cs:177,-0.619149980751964)
--(axis cs:178,-0.516009016056586)
--(axis cs:179,-0.437685870041872)
--(axis cs:180,-0.35807724172804)
--(axis cs:181,-0.323900297752205)
--(axis cs:182,-0.280630861616622)
--(axis cs:183,-0.235357667560482)
--(axis cs:184,-0.169181989548391)
--(axis cs:185,-0.0922013914895479)
--(axis cs:186,-0.0407270489239513)
--(axis cs:187,0.0137513256592481)
--(axis cs:188,0.11945984709978)
--(axis cs:189,0.214660990400334)
--(axis cs:190,0.330012249699977)
--(axis cs:191,0.443131263758445)
--(axis cs:192,0.54344977261956)
--(axis cs:193,0.608752028550478)
--(axis cs:194,0.704305434757307)
--(axis cs:195,0.82288651262968)
--(axis cs:196,0.922872164940625)
--(axis cs:197,0.993454055807157)
--(axis cs:198,1.09570209286485)
--(axis cs:199,1.16320347615814)
--(axis cs:200,1.23029226005538)
--(axis cs:201,1.33986895235194)
--(axis cs:202,1.43150419627252)
--(axis cs:203,1.53673747155959)
--(axis cs:204,1.61647079420522)
--(axis cs:205,1.65679973600287)
--(axis cs:206,1.76578675295858)
--(axis cs:207,1.84291073713577)
--(axis cs:208,1.94170708247274)
--(axis cs:209,2.04798753447315)
--(axis cs:210,2.18279163104996)
--(axis cs:211,2.29071856597147)
--(axis cs:212,2.40361882013176)
--(axis cs:213,2.53920216888665)
--(axis cs:214,2.66074128220346)
--(axis cs:215,2.76478305243265)
--(axis cs:216,2.85763308233907)
--(axis cs:217,2.97661309506853)
--(axis cs:218,3.05537249897014)
--(axis cs:219,3.15002293790417)
--(axis cs:220,3.24114361047854)
--(axis cs:221,3.34257274185074)
--(axis cs:222,3.46731867194642)
--(axis cs:223,3.59138563365885)
--(axis cs:224,3.70666540728184)
--(axis cs:225,3.81756500859715)
--(axis cs:226,3.95892305521271)
--(axis cs:227,4.07736190962915)
--(axis cs:228,4.1843699342063)
--(axis cs:229,4.30136880327443)
--(axis cs:230,4.40839493368334)
--(axis cs:231,4.51530633814867)
--(axis cs:232,4.62796292288862)
--(axis cs:233,4.72651417113997)
--(axis cs:234,4.85523765607031)
--(axis cs:235,4.9894087530121)
--(axis cs:236,5.09774888468445)
--(axis cs:237,5.24497574963785)
--(axis cs:238,5.35053183171353)
--(axis cs:239,5.43313829049075)
--(axis cs:240,5.54797309302629)
--(axis cs:241,5.67937423844531)
--(axis cs:242,5.79372936223389)
--(axis cs:243,5.94231891569655)
--(axis cs:244,6.04199327764068)
--(axis cs:245,6.15137271182604)
--(axis cs:246,6.28088587496535)
--(axis cs:247,6.37462457565616)
--(axis cs:248,6.48391595552239)
--(axis cs:249,6.60899725993487)
--(axis cs:250,6.74191204751497)
--(axis cs:251,6.88175298176573)
--(axis cs:252,7.01673778596224)
--(axis cs:253,7.14706472658493)
--(axis cs:254,7.2795048216305)
--(axis cs:255,7.40844094596702)
--(axis cs:256,7.5557150591195)
--(axis cs:257,7.68623628574902)
--(axis cs:258,7.80569407357991)
--(axis cs:259,7.90584674888141)
--(axis cs:260,8.02115951981935)
--(axis cs:261,8.16565204251968)
--(axis cs:262,8.30668872194282)
--(axis cs:263,8.4368761390916)
--(axis cs:264,8.61756588185055)
--(axis cs:265,8.75694795731024)
--(axis cs:266,8.92483353746967)
--(axis cs:267,9.08779745179869)
--(axis cs:268,9.22828864951648)
--(axis cs:269,9.38718927434614)
--(axis cs:270,9.52887446473779)
--(axis cs:271,9.67377303155036)
--(axis cs:272,9.81311685800548)
--(axis cs:273,9.96827549813123)
--(axis cs:274,10.0997926924165)
--(axis cs:275,10.2470211805849)
--(axis cs:276,10.4212021023754)
--(axis cs:277,10.5479326771675)
--(axis cs:278,10.7228680886973)
--(axis cs:279,10.8467858726494)
--(axis cs:280,10.9586917517029)
--(axis cs:281,11.092104797521)
--(axis cs:282,11.2336939530726)
--(axis cs:283,11.3787824336291)
--(axis cs:284,11.5168418180071)
--(axis cs:285,11.6493734777049)
--(axis cs:286,11.7758417758372)
--(axis cs:287,11.8746719566616)
--(axis cs:288,12.0054716939958)
--(axis cs:289,12.0974528594574)
--(axis cs:290,12.2021010143323)
--(axis cs:291,12.3085947765311)
--(axis cs:292,12.3758785752127)
--(axis cs:293,12.4527130785101)
--(axis cs:294,12.5513484643781)
--(axis cs:295,12.6441782979123)
--(axis cs:296,12.6985120358487)
--(axis cs:297,12.7521792740813)
--(axis cs:298,12.8162623080336)
--(axis cs:299,12.8874082014903)
--(axis cs:300,12.9275487153493)
--(axis cs:300,19.005784617984)
--(axis cs:300,19.005784617984)
--(axis cs:299,18.9392584651764)
--(axis cs:298,18.8437376919664)
--(axis cs:297,18.751154059252)
--(axis cs:296,18.6748212974846)
--(axis cs:295,18.5958217020877)
--(axis cs:294,18.4786515356219)
--(axis cs:293,18.3539535881566)
--(axis cs:292,18.250788091454)
--(axis cs:291,18.1547385568022)
--(axis cs:290,18.0245656523344)
--(axis cs:289,17.8925471405426)
--(axis cs:288,17.7745283060042)
--(axis cs:287,17.6153280433384)
--(axis cs:286,17.4874915574961)
--(axis cs:285,17.3372931889618)
--(axis cs:284,17.1798248486596)
--(axis cs:283,17.0112175663709)
--(axis cs:282,16.8396393802608)
--(axis cs:281,16.6712285358123)
--(axis cs:280,16.5113082482971)
--(axis cs:279,16.376547460684)
--(axis cs:278,16.230465244636)
--(axis cs:277,16.0354006561658)
--(axis cs:276,15.8887978976246)
--(axis cs:275,15.6929788194151)
--(axis cs:274,15.5235406409169)
--(axis cs:273,15.3683911685354)
--(axis cs:272,15.1935498086612)
--(axis cs:271,15.039560301783)
--(axis cs:270,14.8744588685955)
--(axis cs:269,14.7161440589872)
--(axis cs:268,14.5383780171502)
--(axis cs:267,14.378869214868)
--(axis cs:266,14.1984997958637)
--(axis cs:265,14.0097187093564)
--(axis cs:264,13.8491007848161)
--(axis cs:263,13.6497905275751)
--(axis cs:262,13.4966446113905)
--(axis cs:261,13.3376812908137)
--(axis cs:260,13.172173813514)
--(axis cs:259,13.0374865844519)
--(axis cs:258,12.9143059264201)
--(axis cs:257,12.773763714251)
--(axis cs:256,12.6242849408805)
--(axis cs:255,12.4582257206996)
--(axis cs:254,12.3104951783695)
--(axis cs:253,12.1562686067484)
--(axis cs:252,12.0065955473711)
--(axis cs:251,11.8482470182343)
--(axis cs:250,11.688087952485)
--(axis cs:249,11.5376694067318)
--(axis cs:248,11.3927507111443)
--(axis cs:247,11.2620420910105)
--(axis cs:246,11.1491141250346)
--(axis cs:245,11.0019606215073)
--(axis cs:244,10.874673389026)
--(axis cs:243,10.7610144176368)
--(axis cs:242,10.5962706377661)
--(axis cs:241,10.463959094888)
--(axis cs:240,10.3120269069737)
--(axis cs:239,10.1768617095093)
--(axis cs:238,10.0828015016198)
--(axis cs:237,9.95835758369548)
--(axis cs:236,9.79225111531555)
--(axis cs:235,9.66725791365457)
--(axis cs:234,9.51476234392969)
--(axis cs:233,9.3701524955267)
--(axis cs:232,9.25203707711138)
--(axis cs:231,9.12136032851799)
--(axis cs:230,9.00160506631666)
--(axis cs:229,8.87529786339224)
--(axis cs:228,8.74229673246037)
--(axis cs:227,8.62263809037085)
--(axis cs:226,8.48774361145396)
--(axis cs:225,8.33576832473619)
--(axis cs:224,8.21000125938483)
--(axis cs:223,8.08528103300782)
--(axis cs:222,7.94601466138691)
--(axis cs:221,7.80742725814926)
--(axis cs:220,7.68552305618813)
--(axis cs:219,7.57997706209583)
--(axis cs:218,7.46796083436319)
--(axis cs:217,7.37005357159814)
--(axis cs:216,7.23570025099426)
--(axis cs:215,7.12521694756735)
--(axis cs:214,7.00259205112987)
--(axis cs:213,6.86413116444669)
--(axis cs:212,6.71304784653491)
--(axis cs:211,6.58594810069519)
--(axis cs:210,6.46054170228338)
--(axis cs:209,6.30867913219352)
--(axis cs:208,6.18829291752726)
--(axis cs:207,6.07375592953089)
--(axis cs:206,5.98421324704142)
--(axis cs:205,5.86320026399713)
--(axis cs:204,5.80352920579478)
--(axis cs:203,5.70659586177374)
--(axis cs:202,5.58516247039415)
--(axis cs:201,5.4734643809814)
--(axis cs:200,5.34637440661129)
--(axis cs:199,5.26346319050852)
--(axis cs:198,5.17763124046849)
--(axis cs:197,5.05654594419284)
--(axis cs:196,4.96712783505938)
--(axis cs:195,4.84711348737032)
--(axis cs:194,4.71569456524269)
--(axis cs:193,4.60458130478285)
--(axis cs:192,4.52655022738044)
--(axis cs:191,4.41353540290822)
--(axis cs:190,4.28998775030002)
--(axis cs:189,4.16200567626633)
--(axis cs:188,4.05387348623355)
--(axis cs:187,3.93291534100742)
--(axis cs:186,3.86072704892395)
--(axis cs:185,3.79220139148955)
--(axis cs:184,3.69918198954839)
--(axis cs:183,3.61535766756048)
--(axis cs:182,3.55063086161662)
--(axis cs:181,3.49056696441887)
--(axis cs:180,3.44141057506137)
--(axis cs:179,3.34768587004187)
--(axis cs:178,3.25600901605659)
--(axis cs:177,3.13914998075196)
--(axis cs:176,3.02842299152063)
--(axis cs:175,2.93208833583069)
--(axis cs:174,2.83777811911608)
--(axis cs:173,2.75517975087266)
--(axis cs:172,2.66441596569696)
--(axis cs:171,2.59372601911796)
--(axis cs:170,2.51479519684633)
--(axis cs:169,2.43499928536521)
--(axis cs:168,2.34977248771867)
--(axis cs:167,2.24981375611451)
--(axis cs:166,2.14210781075948)
--(axis cs:165,2.03583368407526)
--(axis cs:164,1.96526666435469)
--(axis cs:163,1.88191128287906)
--(axis cs:162,1.8059363433616)
--(axis cs:161,1.71219570185457)
--(axis cs:160,1.60759312646948)
--(axis cs:159,1.53165162764606)
--(axis cs:158,1.45494665077838)
--(axis cs:157,1.38258042207646)
--(axis cs:156,1.3255963814938)
--(axis cs:155,1.22116364990249)
--(axis cs:154,1.14549680712508)
--(axis cs:153,1.06861517448253)
--(axis cs:152,0.966738818367037)
--(axis cs:151,0.891707999200281)
--(axis cs:150,0.806765235419226)
--(axis cs:149,0.702272491437977)
--(axis cs:148,0.604552621101647)
--(axis cs:147,0.519489300322243)
--(axis cs:146,0.447930299092709)
--(axis cs:145,0.360585363594124)
--(axis cs:144,0.285086042622544)
--(axis cs:143,0.20229080225788)
--(axis cs:142,0.131327745775835)
--(axis cs:141,0.0770485384196669)
--(axis cs:140,0.0124403024073241)
--(axis cs:139,-0.0515689559951409)
--(axis cs:138,-0.108393193076168)
--(axis cs:137,-0.172431711633481)
--(axis cs:136,-0.223373885225314)
--(axis cs:135,-0.293202911244374)
--(axis cs:134,-0.361056568826999)
--(axis cs:133,-0.428842157086239)
--(axis cs:132,-0.513335062538788)
--(axis cs:131,-0.566749308796969)
--(axis cs:130,-0.650986804878448)
--(axis cs:129,-0.745034202947907)
--(axis cs:128,-0.840718618493216)
--(axis cs:127,-0.882008573223032)
--(axis cs:126,-0.954197326123676)
--(axis cs:125,-1.03262609437344)
--(axis cs:124,-1.07265340889708)
--(axis cs:123,-1.11907240497928)
--(axis cs:122,-1.17919772180016)
--(axis cs:121,-1.2347351380657)
--(axis cs:120,-1.27712580214778)
--(axis cs:119,-1.30888389762081)
--(axis cs:118,-1.38152978991069)
--(axis cs:117,-1.41788378210618)
--(axis cs:116,-1.46083274613827)
--(axis cs:115,-1.5148786712039)
--(axis cs:114,-1.54633502645761)
--(axis cs:113,-1.57959535058423)
--(axis cs:112,-1.60849966872954)
--(axis cs:111,-1.65111291191376)
--(axis cs:110,-1.71160588145867)
--(axis cs:109,-1.72343333146545)
--(axis cs:108,-1.73785176050749)
--(axis cs:107,-1.74332837402657)
--(axis cs:106,-1.78042655598674)
--(axis cs:105,-1.80534515556876)
--(axis cs:104,-1.83580510570206)
--(axis cs:103,-1.85945764359543)
--(axis cs:102,-1.87013384180199)
--(axis cs:101,-1.91772545355661)
--(axis cs:100,-1.94479102071939)
--(axis cs:99,-1.96340402559683)
--(axis cs:98,-2.00834855711786)
--(axis cs:97,-2.05870742119307)
--(axis cs:96,-2.09049111455612)
--(axis cs:95,-2.12332275035619)
--(axis cs:94,-2.15205582732159)
--(axis cs:93,-2.19481617262069)
--(axis cs:92,-2.26974259117901)
--(axis cs:91,-2.32945316921994)
--(axis cs:90,-2.36749022418895)
--(axis cs:89,-2.42633226413304)
--(axis cs:88,-2.46185183682633)
--(axis cs:87,-2.51360642001814)
--(axis cs:86,-2.56798353329949)
--(axis cs:85,-2.60754726278273)
--(axis cs:84,-2.64496728066762)
--(axis cs:83,-2.70314746994284)
--(axis cs:82,-2.7611804946802)
--(axis cs:81,-2.79228805311642)
--(axis cs:80,-2.83865384925002)
--(axis cs:79,-2.8933723584179)
--(axis cs:78,-2.92410484139166)
--(axis cs:77,-2.99691142991816)
--(axis cs:76,-3.03864028654185)
--(axis cs:75,-3.08336213479595)
--(axis cs:74,-3.13314771798564)
--(axis cs:73,-3.16953238179349)
--(axis cs:72,-3.20646226996243)
--(axis cs:71,-3.24771890629782)
--(axis cs:70,-3.28813296130163)
--(axis cs:69,-3.3271450332832)
--(axis cs:68,-3.35030394313245)
--(axis cs:67,-3.3807231735548)
--(axis cs:66,-3.4011157432393)
--(axis cs:65,-3.42912564090228)
--(axis cs:64,-3.47291929377226)
--(axis cs:63,-3.47739103193746)
--(axis cs:62,-3.50341101978382)
--(axis cs:61,-3.52285297860662)
--(axis cs:60,-3.55647627943463)
--(axis cs:59,-3.59891667042852)
--(axis cs:58,-3.62704019748693)
--(axis cs:57,-3.64937839954971)
--(axis cs:56,-3.65702503957049)
--(axis cs:55,-3.68078184386131)
--(axis cs:54,-3.69600964315155)
--(axis cs:53,-3.70696399306211)
--(axis cs:52,-3.71361665205603)
--(axis cs:51,-3.73668722966234)
--(axis cs:50,-3.74342187213751)
--(axis cs:49,-3.75001249279756)
--(axis cs:48,-3.74134311860287)
--(axis cs:47,-3.71892675244057)
--(axis cs:46,-3.71566132230755)
--(axis cs:45,-3.70889176155382)
--(axis cs:44,-3.73390826869055)
--(axis cs:43,-3.71314091224573)
--(axis cs:42,-3.70433493640558)
--(axis cs:41,-3.66499377056269)
--(axis cs:40,-3.64899882236289)
--(axis cs:39,-3.63778487368353)
--(axis cs:38,-3.61889713026392)
--(axis cs:37,-3.58685354831856)
--(axis cs:36,-3.57879911068324)
--(axis cs:35,-3.57391405770866)
--(axis cs:34,-3.56426322969915)
--(axis cs:33,-3.52821859120666)
--(axis cs:32,-3.53202902146134)
--(axis cs:31,-3.53898841081224)
--(axis cs:30,-3.52234853269243)
--(axis cs:29,-3.50164363035294)
--(axis cs:28,-3.45285878130163)
--(axis cs:27,-3.40976109835568)
--(axis cs:26,-3.36298441515733)
--(axis cs:25,-3.30846050606932)
--(axis cs:24,-3.23100159668841)
--(axis cs:23,-3.13704227832981)
--(axis cs:22,-3.06873648772995)
--(axis cs:21,-3.01141588420117)
--(axis cs:20,-2.95111209755322)
--(axis cs:19,-2.87589456782704)
--(axis cs:18,-2.81480142181776)
--(axis cs:17,-2.726183257066)
--(axis cs:16,-2.63013215896581)
--(axis cs:15,-2.5047700397619)
--(axis cs:14,-2.38726122060105)
--(axis cs:13,-2.28353653594724)
--(axis cs:12,-2.16061645556945)
--(axis cs:11,-2.0395081322925)
--(axis cs:10,-1.9400645309486)
--(axis cs:9,-1.80503157630785)
--(axis cs:8,-1.6452409046507)
--(axis cs:7,-1.48109389252686)
--(axis cs:6,-1.30148216666508)
--(axis cs:5,-1.1329497887105)
--(axis cs:4,-0.929195805321786)
--(axis cs:3,-0.726867731120505)
--(axis cs:2,-0.459349987280885)
--(axis cs:1,-0.170214158674823)
--cycle;
\addlegendimage{area legend, fill=color0, fill opacity=0.2}
\addlegendentry{95\% \acro{CI}}
\end{axis}

\end{tikzpicture}